%% file: MM-VDN.tex
\documentclass[10pt,twocolumn,letterpaper]{article}

\usepackage{iccv}
\usepackage{times}
\usepackage{epsfig}
\usepackage{graphicx}
\usepackage{amsmath}
\usepackage{mathrsfs}
\usepackage{amssymb}

\DeclareMathOperator*{\argmax}{arg\,max}

\usepackage[pagebackref=true,breaklinks=true,letterpaper=true,colorlinks,bookmarks=false]{hyperref}

\iccvfinalcopy 
\def\iccvPaperID{****}

\begin{document}

\title{A Multi-scale Multiple Instance Video Description Network}
\author{
Huijuan Xu\\
UMass Lowell\\
Lowell, MA\\
{\tt\small hxu1@cs.uml.edu}
\and
Subhashini Venugopalan\\
UT Austin\\
Austin, TX\\
{\tt\small vsub@cs.utexas.edu}
\and
Vasili Ramanishka\\
UMass Lowell\\
Lowell, MA\\
{\tt\small vramanis@cs.uml.edu}
\and
Marcus Rohrbach\\
UC Berkeley, ICSI\\
Berkeley, CA\\
{\tt\small rohrbach@eecs.berkeley.edu}
\and
Kate Saenko\\
UMass Lowell\\
Lowell, MA\\
{\tt\small saenko@cs.uml.edu}
}
\maketitle

\input{abstract}

\input{introduction}

\input{related}
\input{method}
\input{experiment}

\input{conclusion}


{\small
\bibliographystyle{ieee}
\bibliography{egbib}
}

\input{supplement}

\end{document}

%% file: abstract.tex
\begin{abstract}
   Generating natural language descriptions for in-the-wild videos is a challenging task. Most state-of-the-art methods for solving this problem borrow existing deep convolutional neural network (CNN) architectures (Alexnet, Googlenet) to extract a visual representation of the input video. However, these deep CNN architectures are designed for single-label centered-positioned object classification. While they generate strong semantic features, they have no inherent structure allowing them to detect multiple objects of different sizes and locations in the frame. Our paper tries to solve this problem by integrating the base CNN into several fully convolutional neural networks (FCNs) to form a multi-scale network that handles multiple receptive field sizes in the original image.  FCNs, previously applied to image segmentation, can generate class heat-maps efficiently compared to sliding window mechanisms, and can easily handle multiple scales. To further handle the ambiguity over multiple objects and locations, we incorporate the Multiple Instance Learning mechanism (MIL) to consider objects in different positions and at different scales simultaneously. We integrate our multi-scale multi-instance architecture with a sequence-to-sequence recurrent neural network to generate sentence descriptions based on the visual representation. Ours is the first end-to-end trainable architecture that is capable of multi-scale region processing. Evaluation on a Youtube video dataset shows the advantage of our approach compared to the original single-scale whole frame CNN model. Our flexible and efficient architecture can potentially be extended to support other video processing tasks.

\end{abstract}

%% file: introduction.tex
\newcommand{\marcus}[1]{\textcolor{green}{Marcus: #1}}
\section{Introduction}
The ability to automatically
describe videos in natural language has many real-world
applications. For example, it could be used to create one-sentence summaries of short video clips in a user's collection for an easier browsing experience. Other applications include content-based video retrieval, descriptive video service (DVS) for the visually impaired, and automated video surveillance. 

\begin{figure}
\begin{center}
   \includegraphics[width=0.8\linewidth]{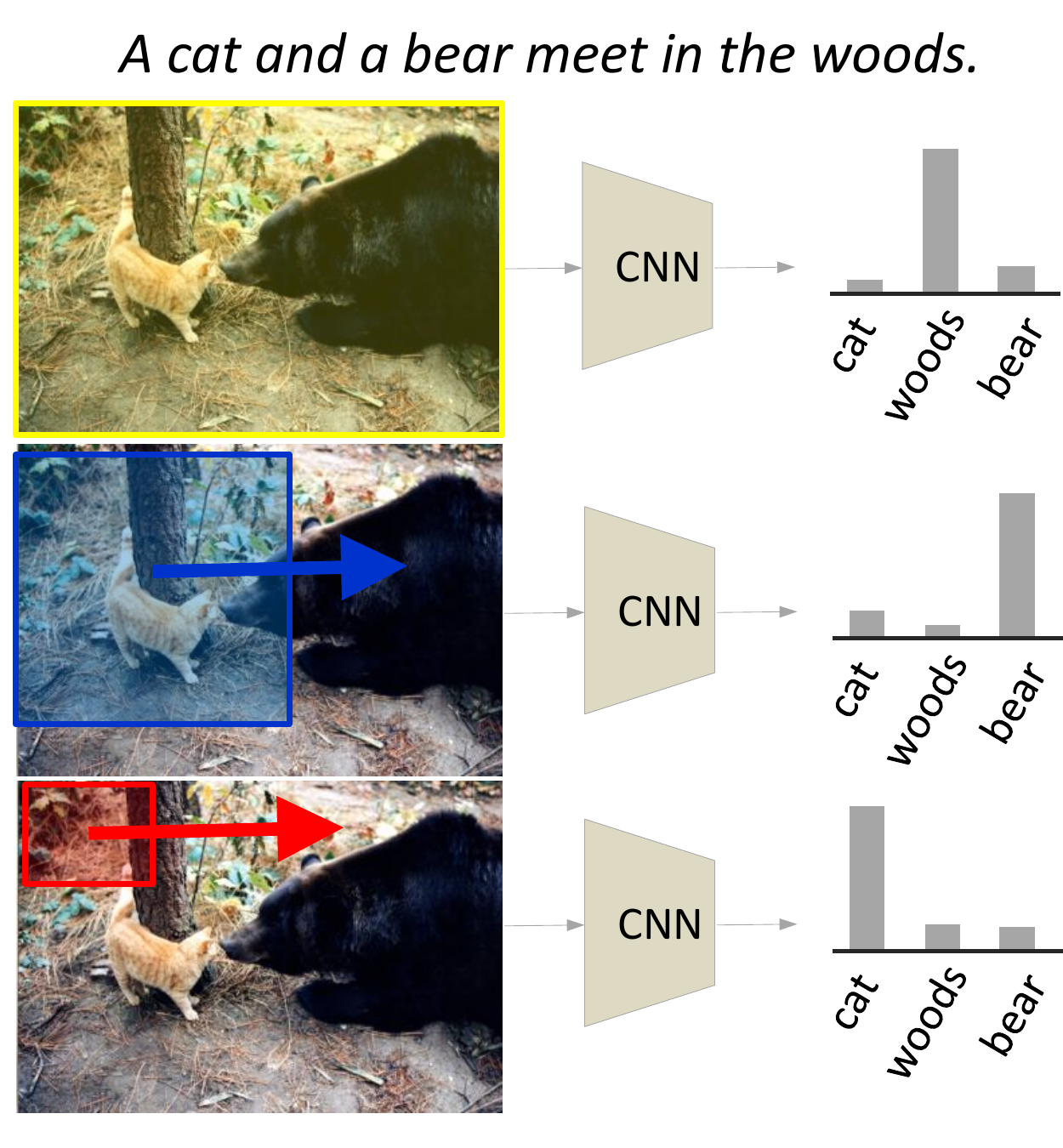}
\end{center}
   \caption{To generate an accurate and detailed description of a video, the visual representation must capture multiple objects in the frame, each of which may have different sizes in the frame. 
   This paper proposes a convolutional neural network integration architecture for video description that simultaneously searches over multiple locations and receptive field sizes.}
\label{fig:scales}
\end{figure}

Most current approaches to this problem make use of pre-trained deep convolutional neural networks (CNNs) as semantic feature extractors for each video frame. These CNN models (e.g. AlexNet~\cite{krizhevsky2012imagenet}, GoogLeNet~\cite{szegedy2014going}, VGG~\cite{vgg16arxiv}) are trained to predict a single object label on images where objects are usually center positioned and occupy most of the image. However, realistic videos are much more complex and contain several objects of different scales in different positions of each video frame, including small objects. As Figure~\ref{fig:scales} illustrates, applying such a CNN to the full frame (top) only detects some of the semantic categories, in this case, only \textit{woods}. To detect smaller objects and actions, receptive fields of different sizes (relative to the original image size) must be used. 

Region processing using a CNN detection model~\cite{girshick2014rich} has been proposed for image-to-text description~\cite{karpathy14arxiv,fang2014captions}, however, applying this model to video would incur considerable computational cost. Furthermore, the region proposal step would prevent end-to-end training of the network. A major advantage of a fully-convolutional network (FCN) is that it can efficiently generate a spatial score map for each class, instead of a single score for each class as in the classification CNN network. Each score in the output score map corresponds to one receptive field in the input image. By up-scaling the input image, FCN can generate larger size score maps whose receptive fields are smaller in the original image. Thus incorporating FCN can capture semantic concepts at different scales and locations in the original input frame. FCN is more efficient than both region proposal methods and the sliding window mechanism. 

In this paper, we propose the first end-to-end trainable video description network that incorporates spatially localized descriptors to capture concepts at multiple scales.
We combine the traditional classification CNN that operates on the scale of the whole image with several smaller receptive fields using an FCN.  
We further incorporate a Multiple Instance Learning (MIL) mechanism to deal with the uncertainty of object scales and positions. The resulting semantic representation of the frames is encoded into a hidden state vector and then decoded into a sentence using a variant of a recurrent neural network proposed in~\cite{s2s:anon}.  
We call our model the \textbf{Multi-scale Multi-instance Video Description Network (MM-VDN)}.

After generating the spatial score maps for the semantic concepts, our network still has to cope with the uncertainty over the number, location, and scale of the detected concepts. Traditional approaches for object detection use training data annotated with the location and size of each object. Such training data are not generally available for videos as they are very time-consuming to collect. Therefore, we propose to use the weak supervision available in the form of sentences to train the network. We incorporate an MIL mechanism, which treats the score maps as bags of examples, each corresponding to some region in the image. An MIL mechanism is applied separately at each scale to select the most likely location, and also applied to select the most likely scale.

Pre-training deep representations on a large classification dataset like ImageNet has proven to be a powerful initialization for many computer vision tasks. We therefore use a pre-trained version of AlexNet, converting it to a fully-convolutional multi-scale network. We note that our approach is general and can be applied to other popular CNNs, such as GoogLeNet or VGG. 
We present a detailed evaluation on a large corpus of Youtube videos~\cite{chen2011collecting}. We evaluate different input scales and training regimes, and show improved results compared to previously proposed architectures.

In the next section, we review related work on multi-scale region processing and video description generation. Then, we describe the design of our MM-VDN network architecture. Finally, we present  experimental results and discussion.

%% file: related.tex
\section{Related work}
Early video description generation methods explicitly constructed a CRF semantic role representation, or Subject-Verb-Object triple, for each video, and used a template model to generate a sentence~\cite{rohrbach:iccv13,thomason:coling14}. Recently, CNNs combined with recurrent neural networks have been applied to the related still-image description task, achieving good results~\cite{cnn_lstm1:archive,cnn_lstm2:archive}. They used the Long Short Term Memory recurrent network (LSTM~\cite{hochreiter1997long}), which incorporates explicitly controllable memory units that allow it to learn long-range temporal dependencies that are very difficult to learn using traditional recurrent networks. 

The hybrid CNN-LSTM recurrent neural network architecture has also recently been applied to video description generation~\cite{video_lstm:naccl,yao15arxiv}. \cite{video_lstm:naccl} first extracts a deep representation of each frame in a video using AlexNet, mean-pools the resulting vectors to produce a vector representation of a complete short video, and then trains an LSTM to decode this vector into a sequence of words in order to produce a descriptive sentence.
However, both of these papers directly use a standard ImageNet-trained CNN model (AlexNet and GoogLeNet) as a feature extractor, without considering the difference between ImageNet and the video domain, such as the object and activity scale, position and the presence of multiple labels in each frame.
We improve on this approach by proposing a multi-scale multi-instance FCN integration model to represent each frame, combined with a sequence-to-sequence variant of LSTM~\cite{s2s:anon} to enable sequential frame processing. 

Before deep learning features gained popularity, researchers proposed spatial pyramid pooling\cite{Lazebnik:2006}, ObjectBank\cite{li2010object}, and other mechanisms to deal with the object scale and position problem for handcrafted features (SIFT, HOG, etc.).
More recently, \cite{sivic} attempted to solve the object scale and position problem in multi-label images using a deep CNN. They make use of AlexNet and add two extra convolutional layers on top of its fc7 layer, and propose a sliding window mechanism using this improved AlexNet model. However the sliding window mechanism is inefficient compared with the recent fully convolutional neural network (FCN)~\cite{long2014fully}. 

The FCN model can be obtained by turning the fully connected layers in the original classification CNN network into convolutional layers. This model can make use of the original classification network's weights as weight initialization but outputs spatial score heat maps for all the classes. This idea was applied to semantic segmentation to produce pixel-wise semantic labels~\cite{long2014fully}\cite{kang2014fully}. But they just use single input scale and get one set of output score maps, then combine the up-scaled output score maps with the previous pooling layers to improve the segmentation result. We utilize FCN in our video description framework to obtain scores for different input scale concepts in the video.

MIL is a well-known weakly supervised learning method that has been recently applied to object classification using deep CNNs. One advantage of deep MIL is that it learns a representation as well as a classifier. To the best of our knowledge, it has not been previously attempted for video understanding tasks.
In~\cite{fcn_MIL_segmenation:CoRR}, the authors introduce the concept of multiple instance learning with FCN to make use of multi-class image labels for training, but their ultimate goal is still image segmentation. We use MIL within each input scale's FCN to deal with the uncertainties about the object positions within each scale, and use MIL between different input scale's FCNs to deal with the uncertainties about different object scales. Our goal is to capture a richer representation of the input frame that can eventually help in generating better visual features for tasks such as video captioning.
We note that our model can also be directly applied to other applications such as multi-label image classification.

%% file: method.tex
\section{Approach}

We present the Multi-scale Multiple Instance Video Description Network (MM-VDN), an end-to-end deep neural network that accepts short videos and produces human-like sentence descriptions. 
We first give an overview of the approach, then describe the components in more detail.

\begin{figure*}
\begin{center}
   \includegraphics[width=\linewidth]{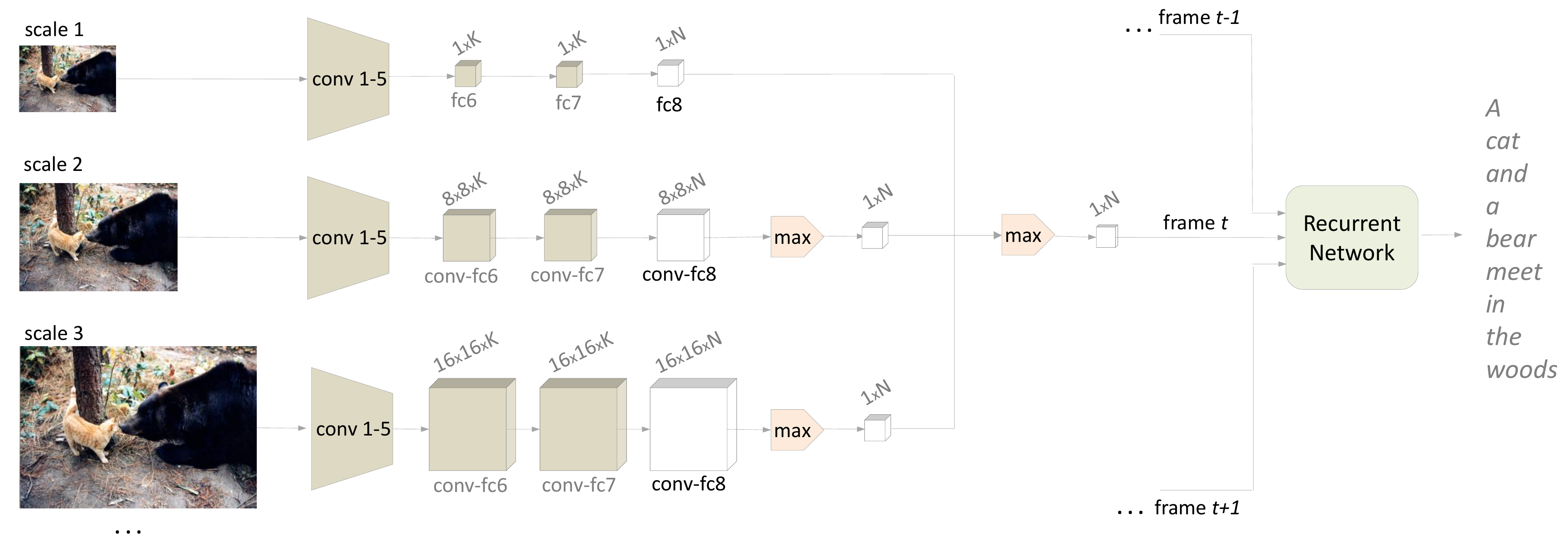}
\end{center}
   \caption{We propose a new end-to-end deep neural network architecture for describing video in natural language. The input to the network is a sequence of video frames, and the output is a sequence of words (a phrase or sentence). Each frame is processed by a multi-scale multi-instance convolutional network and embedded into a $N-$dimensional high-level semantic vector, corresponding to activations of $N$ high-level concepts. A recurrent network accepts such semantic vectors from all frames in the video, and then decodes the resulting state into the output sentence. Unlike previous approaches that used a single-scale single-label architecture (top stream in our network), our network can handle ambiguity in the number, size and location of objects and activities in the scene. Shared parameters transferred from AlexNet are shaded. See text for more details.}
\label{fig:network}
\end{figure*}

\subsection{Overview}
An overview of the architecture is shown in Figure~\ref{fig:network}. The network first reads in and processes frames, then generates a sequence of words. The first part of the network is a visual representation extraction component that processes each input frame to produce an $N$-dimensional hidden semantic state $v_t$, with elements corresponding to scores for the high-level concepts present in the frame. The second part of the network is a recurrent component that takes the sequence of vectors $v_1,...,v_V$ as input and outputs a sequence of words $w_1,...,w_W$. 

Specifically, for the first $t=1,\dots,V$ time steps, the network receives a frame image $I_t$, and applies the visual subnet to produce the hidden state $v_t=f(I_t,\Theta_v)$ using a series of nested convolutional operations whose joint set of parameters is represented by $\Theta_v$:
\begin{equation}
f = f_L \circ f_{L-1} \dots \circ f_1
\end{equation}
Each layer $L$ is defined by its type: a matrix multiplication for convolution or max pooling, an element-wise ReLU non-linearity for an activation function, a normalization layer, etc. Note that, unlike AlexNet, our visual network is fully convolutional at all layers.

The visual hidden state is then passed to the recurrent subnet, which produces a hidden state $z_t$ encoding each additional frame's visual information:
\begin{equation}
z_t=g(v_t; \Theta_r)
\end{equation}
where $\Theta_r$ represents the parameters of all layers in the recurrent subnet, and $g$ is typically a matrix multiplication followed by an element-wise non-linearity. Once the frames have been encoded, for the next time steps $t=V+1,\dots,V+W$, the model starts to decode. Each hidden state $z_{t-1}$ of the previous time stamp is used to obtain the emitted word $w_t$ via a softmax function:
\begin{equation}
p(w_t|z_{t-1}) = \frac{\text{exp}(\Theta_{w}z_{t-1})}{\sum_{w' \in
D} \text{exp}(\Theta_{w'}z_{t-1})} \label{eqn:softmax}
\end{equation}
where $D$ is the word vocabulary. The model learns by maximizing the probability of the correct word sequence given the input frames:
\begin{equation}\label{eqn:condprob}
p(w_1, \ldots, w_W | I_1, \ldots, I_V) 
\end{equation}
More details on how this probability is modeled in our framework are given below.

We first present the details of the visual subnet, which is the main contribution of this paper, followed by a brief overview of the sentence generation recurrent subnet. For full details of the sentence generation recurrent subnet we refer the reader to~\cite{s2s:anon}.

\subsection{Multi-scale Region Processing using an FCN}

The visual subnet (Figure~\ref{fig:network}) is structured as several multi-scale fully convolutional networks connected via the MIL mechanism. We use a base CNN (here AlexNet) that consists of five convolutional and pooling layers followed by three fully-connected layers. The first scale is the base pre-trained CNN classification network applied on the whole frame to capture scene-level semantics. Additional scales consist of the same CNN network but applied in a fully convolutional manner across upsampled versions of the original frame. The MIL mechanism consists of several layers of max-pooling and allows the latent position and scale of concepts to be discovered simultaneously during learning. 

The goal of the visual subnet is to take an input frame $I_t$ and produce a set of probabilities $v_t$ for each high-level concept present in the frame. Typically, this is  accomplished by a CNN trained on image classification, but a classification CNN works best for concepts occupying most of the frame. We therefore cast the CNN as a fully convolutional network to process arbitrary region sizes. 

We start by taking the AlexNet model~\cite{krizhevsky2012imagenet} pre-trained on ImageNet. AlexNet is a well-studied CNN architecture, which consists of five convolutional/pooling layers (conv1-5) and three fully-connected layers (fc6-8). The input image size of AlexNet is designed to be $227\times227$, however, in practice, the input image is resized to $256\times256$, then cropped and mirrored to ten patches of size $227\times227$. Layers fc6 and fc7 have $K$ neurons each, in the case of AlexNet, $K=4096$. Each neuron in the last fc8 layer is trained to predict the probability of a single concept, with a total of $N$ possible concepts. Standard pre-training on ImageNet ILSVRC-1K classification data yields $N=1000$, however, we experiment with other cardinalities of concept spaces. 

Our FCN conversion of AlexNet changes the last three fully-connected layers into convolutional layers, while the first five convolutional layers are kept the same. 
The weights in the last three fully-connected layers of AlexNet are converted to be the filter weights in the last three convolutional layers of the FCN. 
For example, the output of conv5 in AlexNet is $256\times 6\times 6$, which is first  concatenated into a vector of size $9216$, and then connected to fc6. The weight matrix between these two layers is thus of size $4096\times 9216$ in AlexNet. In the FCN, the output of conv5 is no longer concatenated, the size of the filter weights between conv5 and conv-fc6 is of size $4096\times 256\times 6\times 6$, which can be obtained by reshaping the $4096\times 9216$ dimension fc6 weights. Because of this direct conversion relationship, we can initialize the weights of the FCN directly from pre-trained AlexNet weights, and then further fine-tune the concepts on our specific task. FCN can accept arbitrary input image size, so by upsampling the input image, we can get larger output score maps.

The weights of the first seven layers are shared across scales (shown by shading their outputs in Figure~\ref{fig:network}). The fc8 weights are not shared, to allow different concepts to be learned at different scales (e.g., scenes vs objects). We experiment with learning the fc8 weights, either starting from initial pre-trained AlexNet weights, or starting with zero-initialized weights and learning concepts from scratch. A potential advantage of learning new concept weights is that the original pre-trained concepts may not align with those in the video corpus. For example, concepts pre-trained on the ImageNet ILSVRC-1K challenge do not include the concept of the \textit{person} object, while this happens to be the most frequent object mentioned in our Youtube corpus.

In addition to the original whole-image AlexNet, we create several FCNs using the above conversion, one for each scale.
For  the input frame $I_t$ and label set be $\mathscr{C}$,  each FCN produces an output score map $\hat{p}_c(x,y)$ for the $c^{th}$ label at location $(x,y)$. 

\subsection{Multiple Instance Learning over Locations}

Performing supervised learning of semantic concepts in the video frame would require labeled data in the form of frames and object labels at each location and scale. Since such data is very difficult to obtain, we resort to a multiple instance learning approach. In contrast with supervised learning, MIL allows weaker forms of supervision where training examples come in \textit{bags}. Negative bags typically contain exclusively negative instances, while positive bags are only known to contain at least one positive instance. Thus, true positive labels are latent and must be discovered during learning. 

In our case, we consider a positive bag to be all image patches corresponding to all possible receptive field locations at all scales in a given frame. If our task were concept detection, the positive bag label would be the presence/absence for each of $N$ possible concepts. We can treat the words in the sentence as concepts to be predicted. However, unlike in traditional MIL where the bag label is unambiguous, here there is additional ambiguity as we only have access to a sequence of words for the entire sequence of frames. We must therefore solve the alignment problem of assigning words to each frame. In our approach, this is implicitly handled by the recurrent LSTM network, described below. For now, we assume the bag label is itself latent and must be inferred during network training.

We define a MIL max pooling layer on top of each FCN output score map to capture the maximum score $\hat{p}_c(x_c,y_c)$ for each label $c$, which infers the latent object positions.  $(x_c, y_c)$, which denotes the position of the maximum score for class $c$, is obtained by a max-pooling operation:
\begin{equation}
(x_c,y_c) = \argmax_{\forall (x,y)} \hat{p}_c(x,y) \hspace{0.2in} \forall c \in \mathscr{C}
\end{equation}

\subsection{Multiple Instance Learning over Scales}

As mentioned above, the input image size of AlexNet is $227\times 227$, and the output is a score vector of size $N$ of concepts $\mathscr{C}$. After the FCN conversion, if the input image size for the FCN remains unchanged ($227\times 227$), the output will be $1\times 1 \times N$ score maps. When the input image size of the FCN increases, the size of each output score map also increases to $h \times h \times N$, where $h$ is the output score map size. However, for a fixed FCN structure, each score in the output score map corresponds to a fixed size receptive field in the input image ($355\times 355$ \cite{long2014fully}). 
\begin{table}[h]
\centering
\begin{tabular}{|c|c|c|}
\hline
\textbf{Input Image Size}    &  \textbf{Score Map Size} &  \textbf{Height Ratio} \\ \hline
$227 \times 227$  & $1 \times 1$  & 100\%  \\ \hline
$259 \times 259$  & $2 \times 2$   & 100\%   \\ \hline
$323 \times 323$  & $4 \times 4$   & 100\%   \\ \hline
$451 \times 451$  & $8 \times 8$   & 78.7\%   \\ \hline
$707 \times 707$  & $16 \times 16$ & 50.2\%  \\ \hline
\end{tabular}
\hspace{0.1in}
\caption{Height ratio of the receptive field to the original input image for different input sizes in the FCN. The size of the receptive field is $355\times355$ for all input image sizes.}
\label{tab:multi_scale}
\end{table}
The receptive field size in FCN is determined by the FCN structure, and is independent of the input image size. For small input image size, the receptive filed area ($355\times 355$) may contain some padding areas in the margin of the input image.

When the input image of the FCN has been upscaled, each score in the output score map corresponds to a smaller region in the original unscaled image. Thus, by using a FCN coupled with an upscaled input image, we can capture smaller objects. We further combine several FCNs with different input image sizes (scales), and apply MIL across the scales to capture concepts of different scales simultaneously. The ratio of the receptive field height  to the original image height for several different input scales is shown in Table \ref{tab:multi_scale}. 
Note that, for several input scales in Table \ref{tab:multi_scale}, the input image size is smaller than the receptive field size ($355\times 355$). This occurs because the margin of several layers' output (conv2/conv3/conv4/conv5) in the FCN is padded to generate the final score maps.

We define an additional MIL element-wise max layer on top of the multi-scale FCNs to select among different input scales $s$ for each concept.
The output of this layer is the final semantic concept vector for the $t$th frame:
\begin{equation}
v_t = \max_{\forall (s)} \hat{p}_c(x_c,y_c,s) \hspace{0.2in} \forall c \in \mathscr{C}
\end{equation}
The loss on this layer (as on all others) is propagated back from the word output layer.
Next, we describe how this visual representation is aggregated across frames and translated to the output description sentence.

\subsection{Recurrent Subnet for Word Generation}

We use a two-layer encoder-decoder LSTM model proposed in~\cite{s2s:anon} to generate descriptions of video. The encoder part encodes the visual input for each frame, and the decoder part accepts the hidden state from the encoder and outputs words in sequence. Special symbols are used to indicate the beginning and end of the sentence. The encoding-decoding paradigm is used to estimate the conditional probability of an output sequence $(w_1,\cdots, w_W)$ given an input sequence of visual state vectors $(v_1, \cdots, v_V)$, see Eq. \ref{eqn:condprob}.

In the encoding phase, a part of this conditional probability is computed by generating a fixed length representation $z$ based on the entire sequence of inputs ($v_1, \ldots, v_V$). 
\begin{figure*}
\begin{center}
   \includegraphics[width=0.8\linewidth]{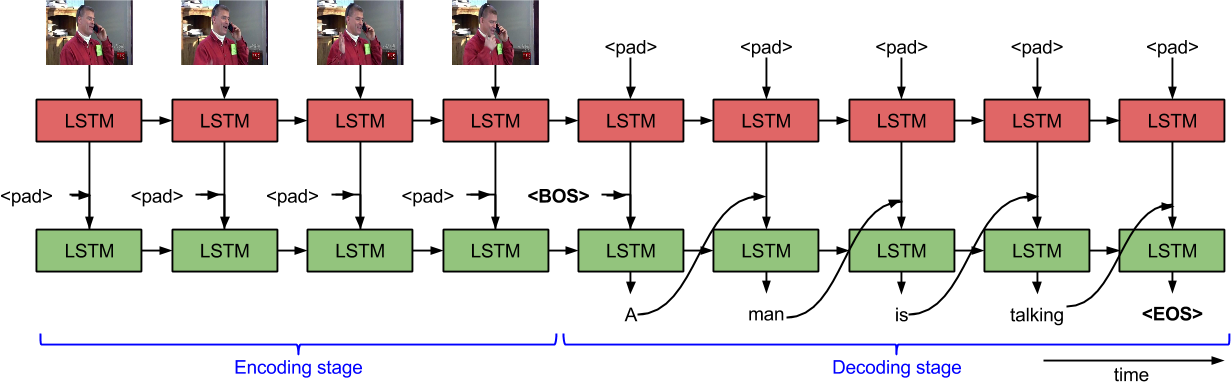}
\end{center}
   \caption{Structure of the LSTM-based recurrent network used in our approach (figure from~\cite{s2s:anon}). Two hidden layers with LSTM cells are used; the first layer encodes the visual input for each frame, and the second layer accepts the hidden state from the first layer and outputs words. Special symbols are used to indicate the beginning and end of the sentence.}
\label{fig:s2s}
\end{figure*}
The decoding step then computes the probabilities of the output sequence of words ($w_1, \ldots, w_W)$ as:
\begin{equation*}
p(w_1, \ldots, w_W | v_1, \ldots, v_V) = \prod_{i=1}^{W} p(w_i | z, w_1, \ldots, w_{i-1})
\end{equation*}
where the distribution of $p(w_t | z, w_1, \ldots, w_{t-1})$ is given by a softmax over all of the words in the vocabulary $D$. The overview of the architecture is shown in Figure~\ref{fig:s2s}, see ~\cite{s2s:anon} for more details.

%% file: experiment.tex
\section{Experiments}
In this section, we evaluate several variants of our approach and compare it to related work.
We use the BLEU\cite{papineni2002bleu} and METEOR\cite{banerjee2005meteor} scores to evaluate the generated sentences against all reference sentences. BLEU is the most commonly used metric in image description literature, but METEOR is also shown to be a good evaluation metric in a recent study \cite{elliott2014comparing}.

\subsection{Dataset and Preprocessing}
We perform our experiments on the Microsoft Research Video Description Corpus (MSVD) \cite{chen2011collecting}. The dataset contains 1,970 short Youtube video clips paired with multiple human-generated natural-language descriptions ($\sim$ 40 English sentence descriptions per video). The video clips are 10 to 25 seconds in duration and typically consist  of  a  single  activity. The 1,970 videos are split into training set (1,200 videos), validation set (100 videos) and testing set (670 videos), as used by the prior work on the same video description task~\cite{guadarrama:iccv13}\cite{thomason:coling14}\cite{xu2015jointly}\cite{video_lstm:naccl}. We perform model selection on the validation set based on the BLEU score.

To pre-train a CNN model which is targeted to our video description dataset, we perform transfer learning by pre-training the classification CNN on ImageNet. For this purpose, we compose a subset of ImageNet consisting of the 566 categories (synsets) which are present in the MSVD dataset. To produce the list of synsets, we manually matched each subject or object that appears in the sentences to the closest synset available in ImageNet. This reduced the more than 900 initial nouns present in MSVD to the 566. We compose the dataset from ImageNet single label images and fine-tune the BVLC reference model (AlexNet) \cite{jia2014caffe} on these categories (all layers are fine-tuned). We use this model as initialization of the AlexNet model in our experiment. We transfer the 566 category AlexNet fc6/fc7/fc8 weights to the FCN conv-fc6/conv-fc7/conv-fc8 layers by reshaping, and initialize the FCN conv1-5 layer weights directly from the 566 category AlexNet.

We also experiment with using the BVLC published 1000-category AlexNet weights as the weight initialization for our model. We find that the experiment results, measured by BLEU and METEOR, are almost the same as using the 566 MSVD category fine-tuned model as weight initialization. In the following experiments, we still choose to use the 566 category fine-tuned model weights as weight initialization of our model.

We also tried several fine-tuning mechanisms with our integrated model, e.g. first fine-tuning each single-scale FCN to get the best set of parameters, and then doing a global fine-tuning pass on all the parameters that need to be fine-tuned. Experimental results did not show any obvious improvements in our  setting, so we choose to directly fine-tune  all the parameters together.

\subsection{Relationship to Related Work}
Several works \cite{guadarrama:iccv13}\cite{thomason:coling14}\cite{xu2015jointly} that explored template based sentence generation method on the same MSVD video dataset only reported SVO accuracy and did not provide a BLEU/METEOR value for generated sentences. 
Recent papers using the CNN and LSTM combination architecture to directly generate sentences for images or videos usually report the BLEU/METEOR scores. \cite{video_lstm:naccl} uses mean pooling of CNN fc7 features as the visual input at each time stamp of a two-layer LSTM to generate descriptions for videos. We use their reported LSTM-YT model result (without fine-tuning on still-image description COCO or FLICKR datasets) as one baseline. We also compare to BLEU and METEOR scores for a template model called FGM proposed in~\cite{thomason:coling14} which uses a factor graph to improve the SVO prediction.

Since our main contribution is to incorporate region processing at multiple scales, we compare our model to others that use the same underlying AlexNet model, but at the standard whole-image scale. Two papers currently in review (\cite{s2s:anon} and \cite{yao15arxiv}) showed that higher overall performance can be obtained by replacing AlexNet with the more powerful GoogLeNet or VGG CNNs. \cite{s2s:anon} also incorporate additional optical flow features, while \cite{yao15arxiv} add 3-D convnet motion features trained on extensive activity corpora. To enable fair comparison, we omit models that use better underlying features or add motion features.

\cite{s2s:anon} extends the two-layer LSTM model used in~\cite{video_lstm:naccl} to encoder-decoder mode by padding the video frame sequence and word sequence. We integrate our multi-scale FCN and MIL model with this two-layer encoder-decoder LSTM model, and show the effectiveness of our multi-scale FCN and MIL mechanism.

\subsection{Ablation Study of Different Scales}
In our  model, the input image for the AlexNet part is resized to $256\times256$ and cropped to $227\times227$ to generate five candidate patches (four corners and one center) without mirroring. The input size for other scales is directly set to be the input size listed in Table~\ref{tab:multi_scale}, without cropping or mirroring. The AlexNet weights are initialized with the 566-category ImageNet model, and conv1 to fc7 weights are kept fixed during training. The FCN weights are initialized with the reshaped 566-category ImageNet model weights, and conv1 to conv-fc7 weights are also kept fixed. The fc8 and conv-fc8 layers directly connect with the LSTM recurrent neural network via max operations, and only fc8, conv-fc8 and the LSTM parameters are fine-tuned on the training videos. We have also tried to fine-tune the weights in conv-fc6 and conv-fc7, but the result was poor compared with keeping these layers fixed.

In this set of experiments, we investigate the effect of using different single input scales in the FCN model, and several combinations of different scales. The BLEU and METEOR scores are shown in Table~\ref{tab:BleuMeteor1}. For a single scale, the original AlexNet whole-frame scale is actually better than the other two scales alone. However, the combination of whole-frame scale and a scale of $451\times451$ gets a boost of 4.7 in BLEU value and a boost of 1 in METEOR value. If we further add the input scale of  $707\times707$, performance is degraded. 
This indicates that the optimal scales are dataset-specific; additional scales could be needed to achieve good performance on other datasets. Our model is flexible enough to integrate different input scales and  combinations of several scales, depending on the dataset.

\begin{table}[t]
\centering
\begin{tabular}{|c|c|c|}
\hline
\textbf{Score Map Size}    &  \textbf{BLEU} &  \textbf{METEOR} \\ \hline
$AlexNet_{bvlc}$  & 32.96  & 28.04  \\ \hline
$AlexNet^*$  & 32.80  & 28.09  \\ \hline
$8 \times 8$  & 31.05   & 27.44   \\ \hline
$16\times 16$  & 27.92 & 25.90  \\ \hline
$AlexNet^* + 8 \times 8$  & \textbf{37.64}  & \textbf{29.00}  \\ \hline
$AlexNet^* + 16 \times 16$  & 26.59  & 25.2  \\ \hline
$AlexNet^* + 8 \times 8 +16 \times 16$  & 31.94  & 26.99  \\ \hline
\end{tabular}
\hspace{0.1in}
\caption{BLEU and METEOR results for the ablation study. $AlexNet^*$ represents the fine-tuned AlexNet model.}
\label{tab:BleuMeteor1}
\end{table}

\subsection{Comparison to Published Results}
There are several variations of BLEU and METEOR calculation methods. We use the same BLEU and METEOR calculation script as was used by the authors of~\cite{video_lstm:naccl} to produce results for all baselines and our method. The comparison of the best MM-VDN model from the ablation study ($AlexNet + 8 \times 8$ score maps) to the other two baselines is listed in Table~\ref{tab:BleuMeteor2}. Our MM-VDN model provides a distinct improvement in both BLEU and METEOR compared to the other two baselines. We note that even better results can be obtained by our model using pre-training on still image data~\cite{video_lstm:naccl} or by replacing AlexNet with deeper convnets such as~\cite{vgg16arxiv,szegedy2014going}. We leave this for future work.

Some examples of the generated sentences are shown in Table~\ref{tab:example_videos}. In several cases, such as in row $1$, $3$, $4$, $5$ of Table~\ref{tab:example_videos}, our model correctly identifies smaller objects that are missed by the whole-frame AlexNet model, like carrot, guitar, piano and skateboard.

\begin{table}[t]
\centering
\begin{tabular}{|c|c|c|}
\hline
\textbf{method}    &  \textbf{BLEU} &  \textbf{METEOR} \\ \hline
FGM~\cite{video_lstm:naccl}  & 13.68  & 23.9  \\ \hline
LSTM-YT~\cite{video_lstm:naccl}  & 31.19  & 26.87 \\ \hline
MM-VDN ($AlexNet^* + 8 \times 8$)  & \textbf{37.64}  & \textbf{29.00}  \\ \hline
\end{tabular}
\hspace{0.1in}
\caption{Comparison to other baselines.}
\label{tab:BleuMeteor2}
\end{table}

\subsection{Visualizations}
In this section, we show some visualizations of the multiscale features learned by our MM-VDN model. In the following table~\ref{tab:visua1}, the first column shows one sampled frame from each video. The second column shows one $8\times 8$ conv-fc8 heat map for this sampled frame, which corresponds to the channel with highest activation value in FCN with input size $451\times 451$. The third column shows the value of the whole frame AlexNet fc8 feature for this sampled frame. The fourth column shows the value of the max pooled conv-fc8 feature in FCN with input size $451\times 451$ for this sampled frame. The last column shows the feature value of the combination of two scales (max of AlexNet fc8 and FCN451 conv-fc8).

The sampled frames are shown in their original size. The heat maps clearly show that the model is able to localize smaller regions and assign them to a high-level semantic feature. The histograms indicate that the highest-scoring semantic conv-fc8 feature is not always the same as the highest-scoring whole-frame fc8 feature, and that the feature values of the two scales can be complementary to each other.

%% file: conclusion.tex
\section{Conclusion}
This paper proposed a Multi-scale Multi-instance Video Description Network (MM-VDN), which combines a convolutional and a recurrent part to simultaneously learn to extract useful high-level concepts and generate language. The MM-VDN model integrates the Fully Convolutional Network (FCN) conversion of a classification CNN to potentially capture medium and small scale concepts in the video frame. It also incorporates a Multi-instance learning mechanism to deal with the uncertainty about the number, position and scale of concepts in the video frame. The  model is shown to be effective on the task of video description generation, compared to the single scale whole-frame classification CNN.

The MM-VDN model is efficient and extensible. Its efficiency makes it especially suitable to process video, where region proposal generation mechanisms would be prohibitively slow. It can integrate several FCNs, each with a different input scale. It can also integrate the FCN conversion of several recent advanced CNN models (e.g. VGG and GoogLeNet). The MM-VDN model can also be applied to other tasks, beyond the video description task. For example, replacing the recurrent network with a multi-label loss layer would allow multi-scale multi-label image classification.

Future work includes investigating the FCN conversion of the recent advanced CNN models in MM-VDN framework, and addition of motion features. We are also interested in investigating the LSTM bidirectional optimization and applying more complex language models to improve the LSTM language generation model.

\section*{Acknowledgement}
This research was partially supported by NSF Awards  IIS-1451244 and IIS-1212798.
Marcus Rohrbach was supported by a fellowship within the FITweltweit-Program of the German Academic Exchange Service (DAAD).

		\begin{table*}[t]
		\begin{center}
			\begin{tabular}{c|l}
				\hline
				\bf MM-VDN predicts correct sentences & \\ \hline \\
				\begin{tabular}{c}
				\includegraphics[width=0.15\textwidth]{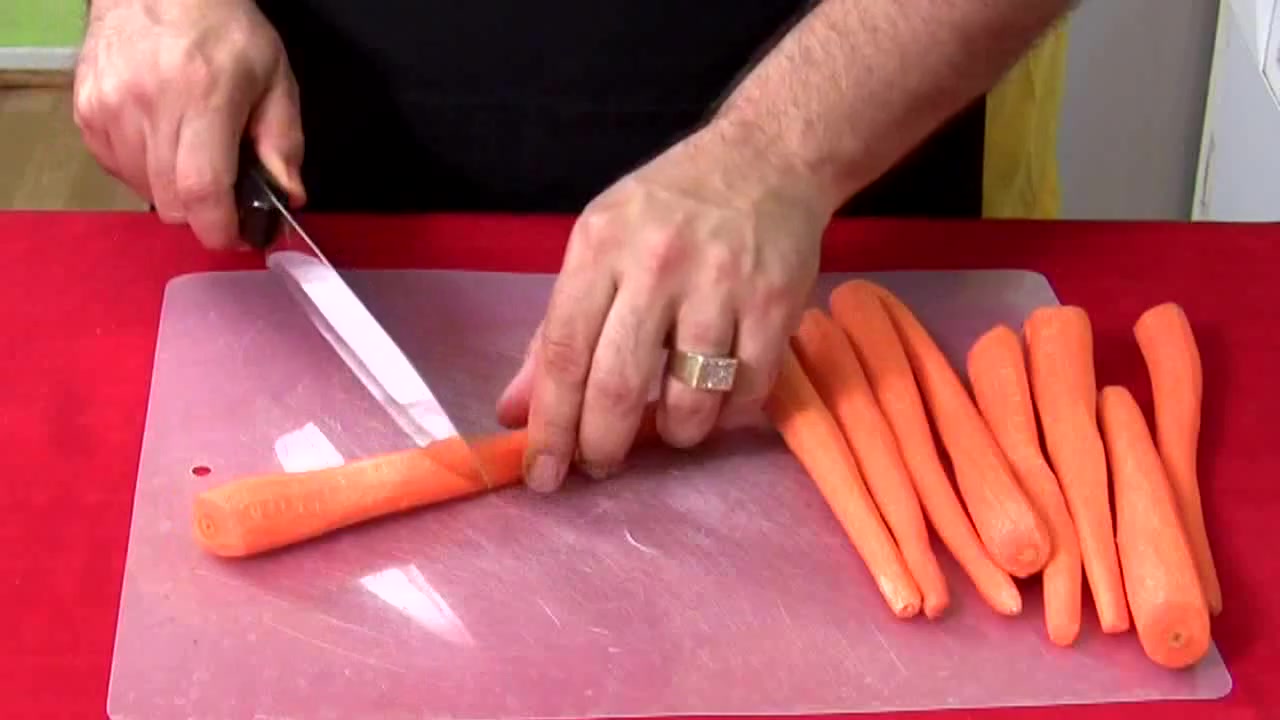}
				\includegraphics[width=0.15\textwidth]{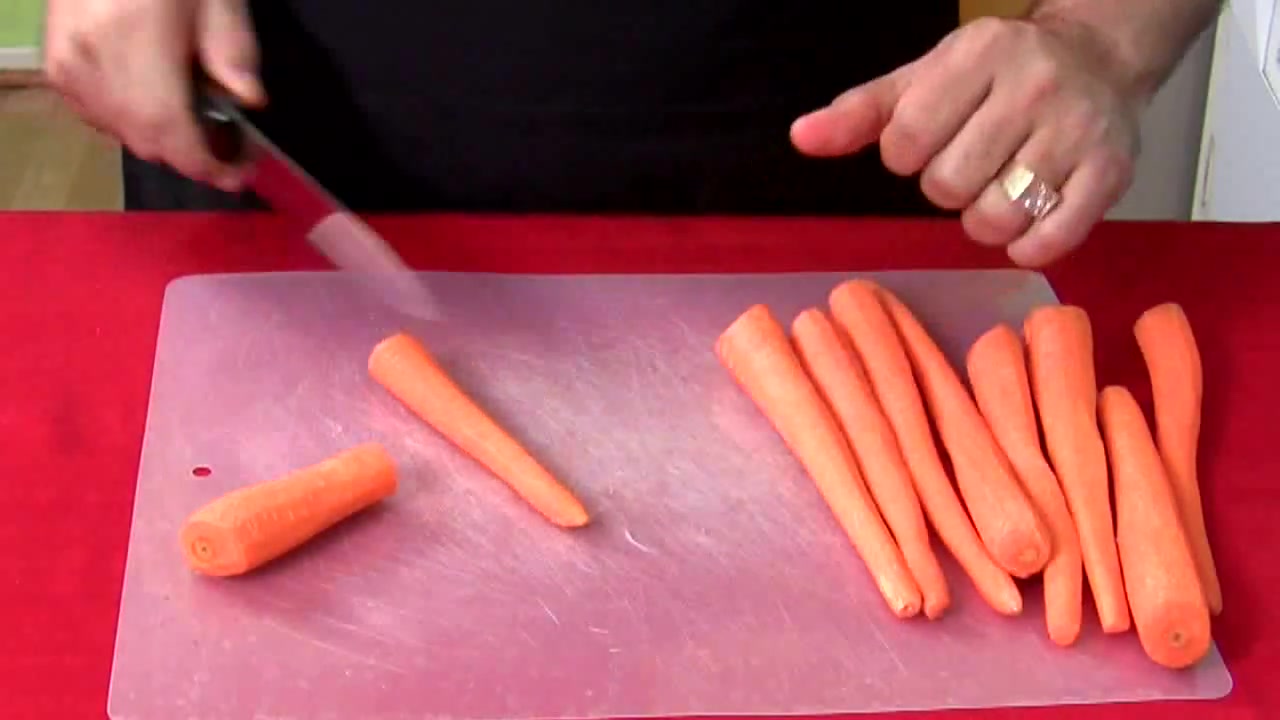}
				\includegraphics[width=0.15\textwidth]{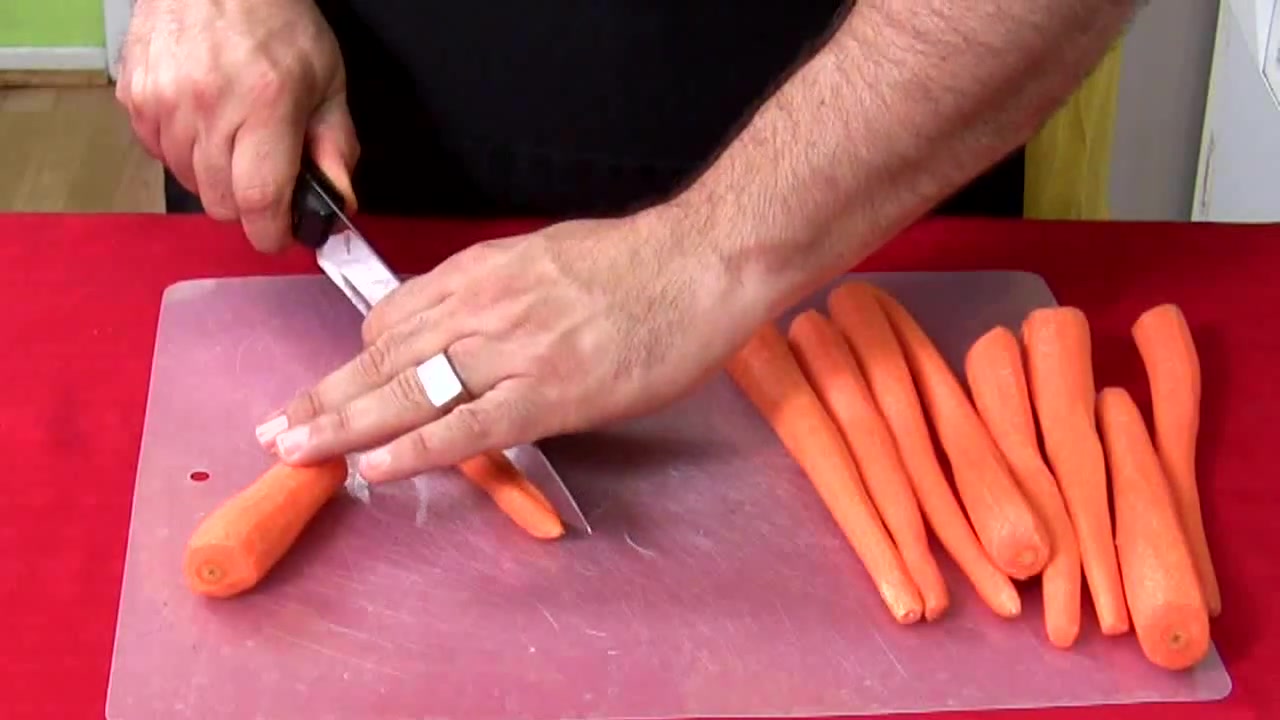}
				\includegraphics[width=0.15\textwidth]{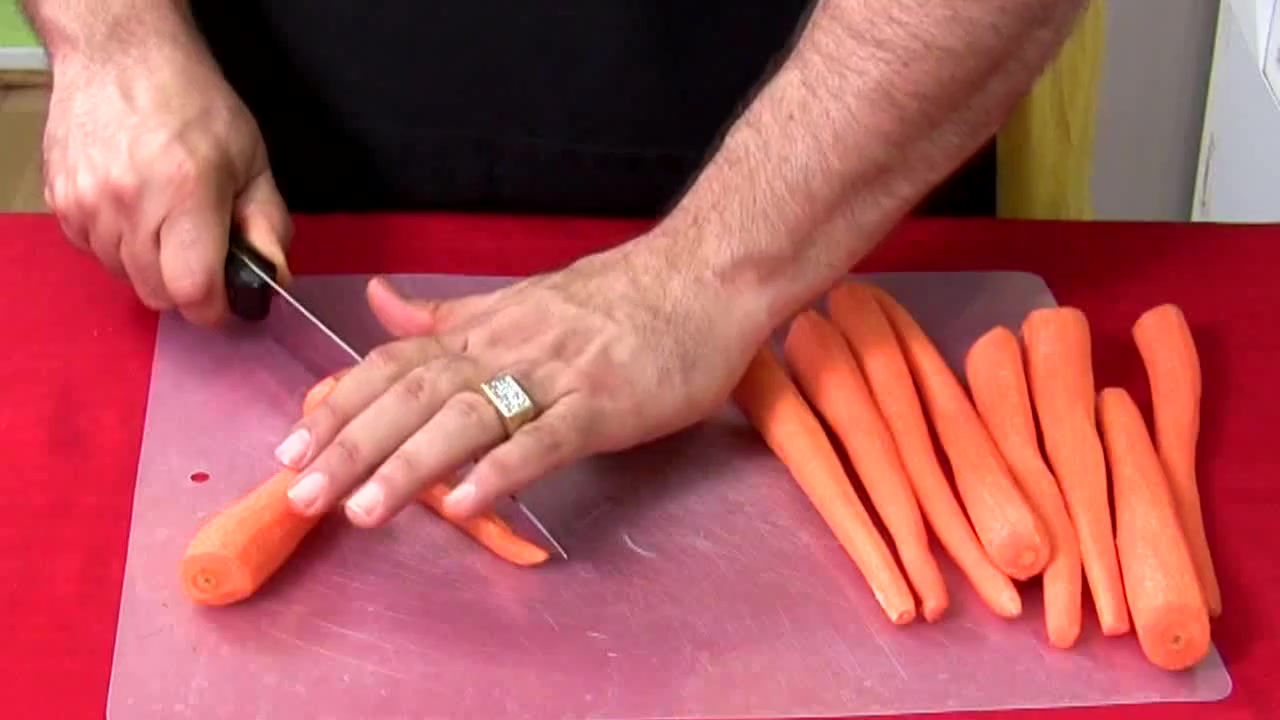}
				\end{tabular}
				&
				\begin{tabular}[c]{@{}l@{}l@{}}
				GT: \textit{A man is cutting carrots.} \\  FGM: A person is cutting a chicken .\\  
				LSTM-YT: A woman is peeling a potato.  \\AlexNet: A man is slicing a tomato. \\
				MM-VDN: \textbf{A man is slicing a carrot.}\end{tabular}
				\\ \cline{2-2}
				\begin{tabular}{c}
				\includegraphics[width=0.15\textwidth]{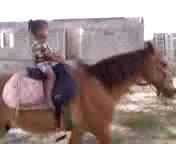}
				\includegraphics[width=0.15\textwidth]{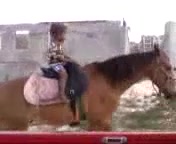}
				\includegraphics[width=0.15\textwidth]{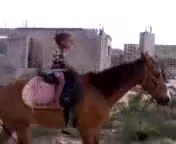}
				\includegraphics[width=0.15\textwidth]{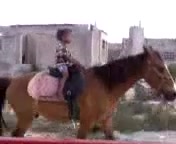}
				\end{tabular}
				&
				\begin{tabular}[c]{@{}l@{}l@{}}
				GT: \textit{A child is riding a horse.} \\ FGM: A person is playing with a person.\\  
				LSTM-YT: A man is running.  \\AlexNet: A man is walking on the ground. \\
				MM-VDN: \textbf{A man is riding a horse.}\end{tabular}
				\\ \cline{2-2}
				\begin{tabular}{c}
				\includegraphics[width=0.15\textwidth]{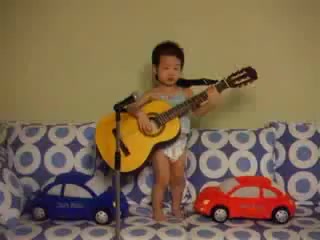}
				\includegraphics[width=0.15\textwidth]{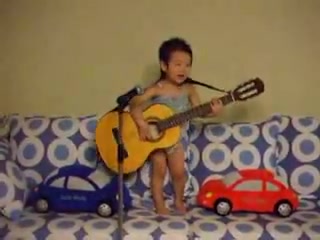}
				\includegraphics[width=0.15\textwidth]{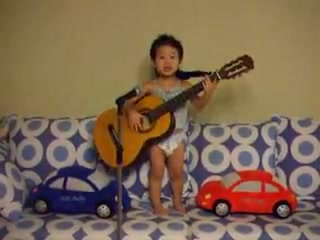}
				\includegraphics[width=0.15\textwidth]{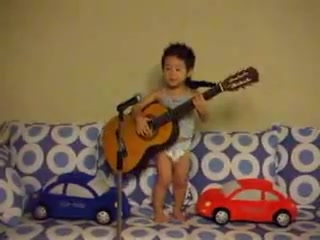}
				\end{tabular}
				&
				\begin{tabular}[c]{@{}l@{}l@{}}
				GT: \textit{A boy is playing a guitar.} \\ FGM: A person is playing with a person.\\  
				LSTM-YT: A women of dancing. \\ AlexNet: A man is playing a horse. \\
				MM-VDN: \textbf{A man is playing a guitar.}\end{tabular}
				\\ \cline{2-2}
				\begin{tabular}{c}
				\includegraphics[width=0.15\textwidth]{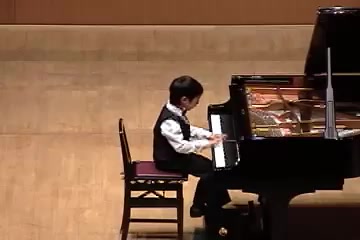}
				\includegraphics[width=0.15\textwidth]{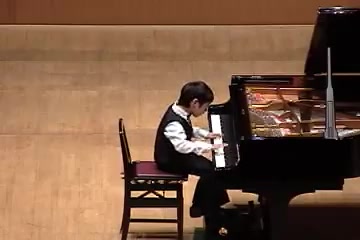}
				\includegraphics[width=0.15\textwidth]{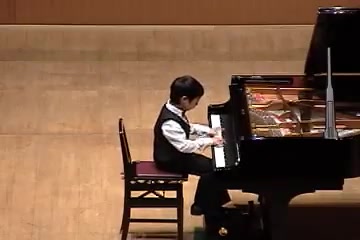}
				\includegraphics[width=0.15\textwidth]{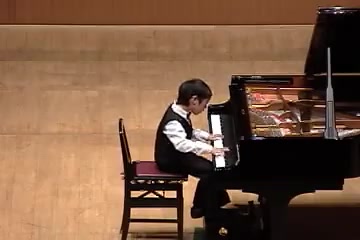}
				\end{tabular}
				&
				\begin{tabular}[c]{@{}l@{}l@{}}
				GT: \textit{A boy is playing a piano.} \\  FGM: A person is playing the guitar.\\  
				LSTM-YT: A man is playing the boxing. \\ AlexNet: A man is playing the guitar. \\
				MM-VDN: \textbf{A man is playing a piano.}\end{tabular}
				\\ \cline{2-2}
				\begin{tabular}{c}
				\includegraphics[width=0.15\textwidth]{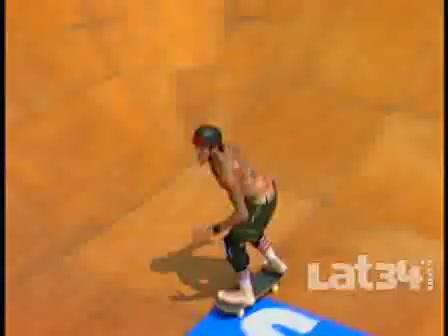}
				\includegraphics[width=0.15\textwidth]{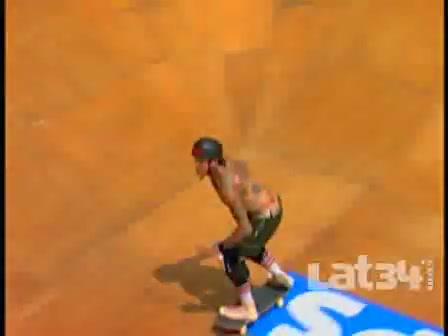}
				\includegraphics[width=0.15\textwidth]{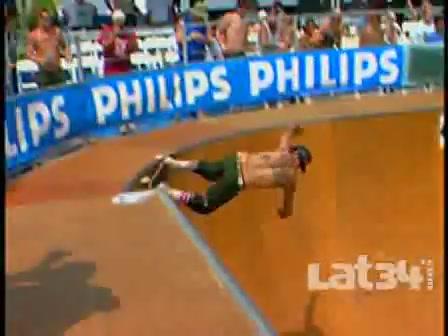}
				\includegraphics[width=0.15\textwidth]{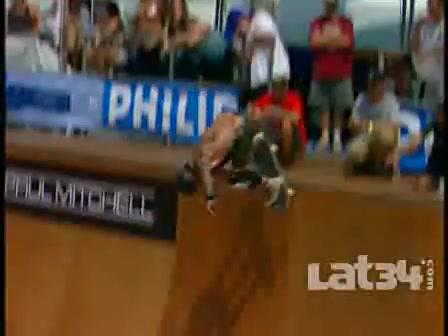}
				\end{tabular}
				&
				\begin{tabular}[c]{@{}l@{}l@{}}
				GT: \textit{A man is skateboarding.} \\ FGM: A person is playing with a person.\\  
				LSTM-YT: A cat is running on a toy.  \\ AlexNet: A man is pushing a wall. \\
				MM-VDN: \textbf{A man is doing a skateboard.}\end{tabular}
				\\ \hline
				\bf MM-VDN predicts partially correct sentences & \\ \hline 
                \\
				\begin{tabular}{c}
				\includegraphics[width=0.15\textwidth]{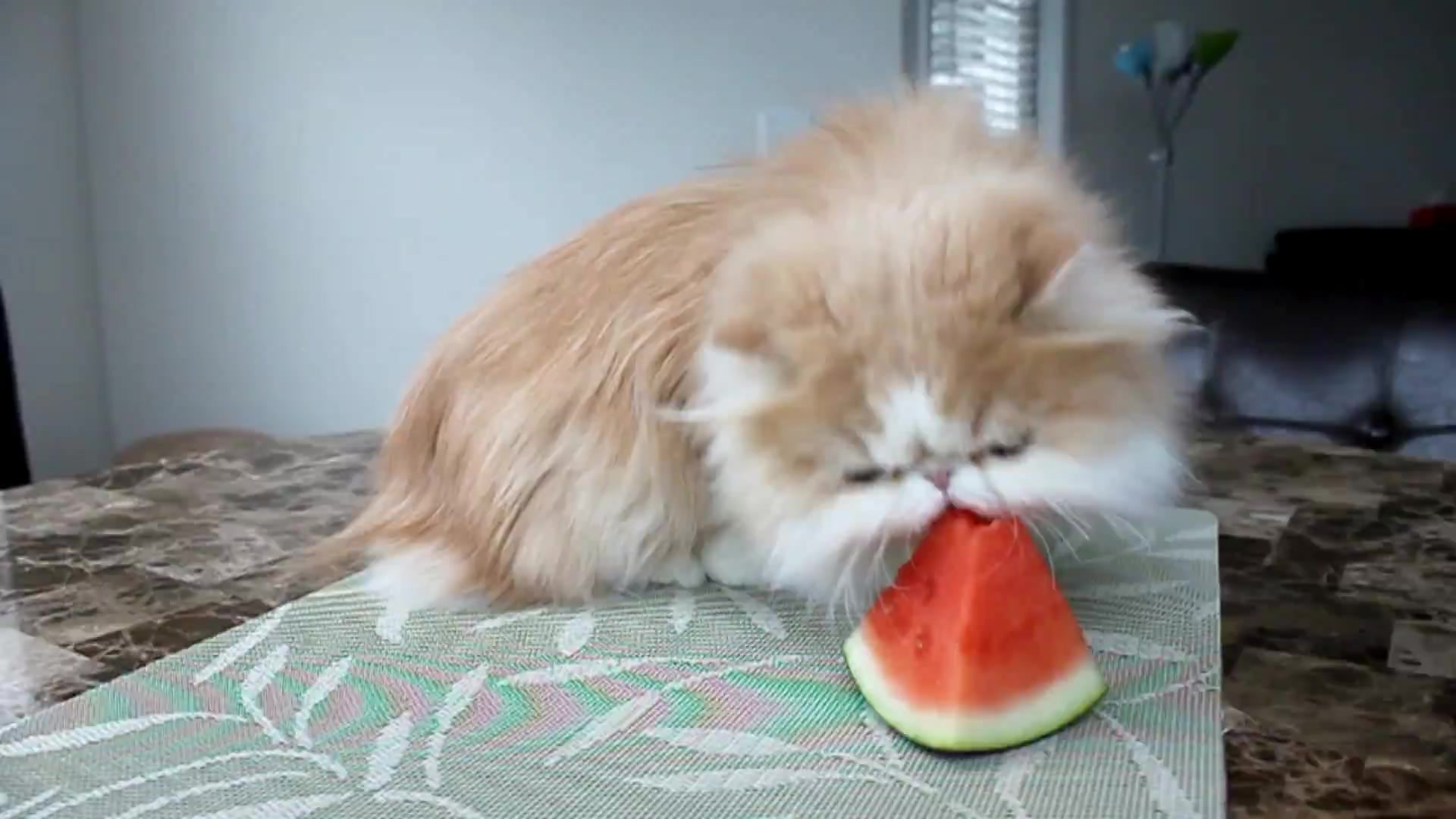}
				\includegraphics[width=0.15\textwidth]{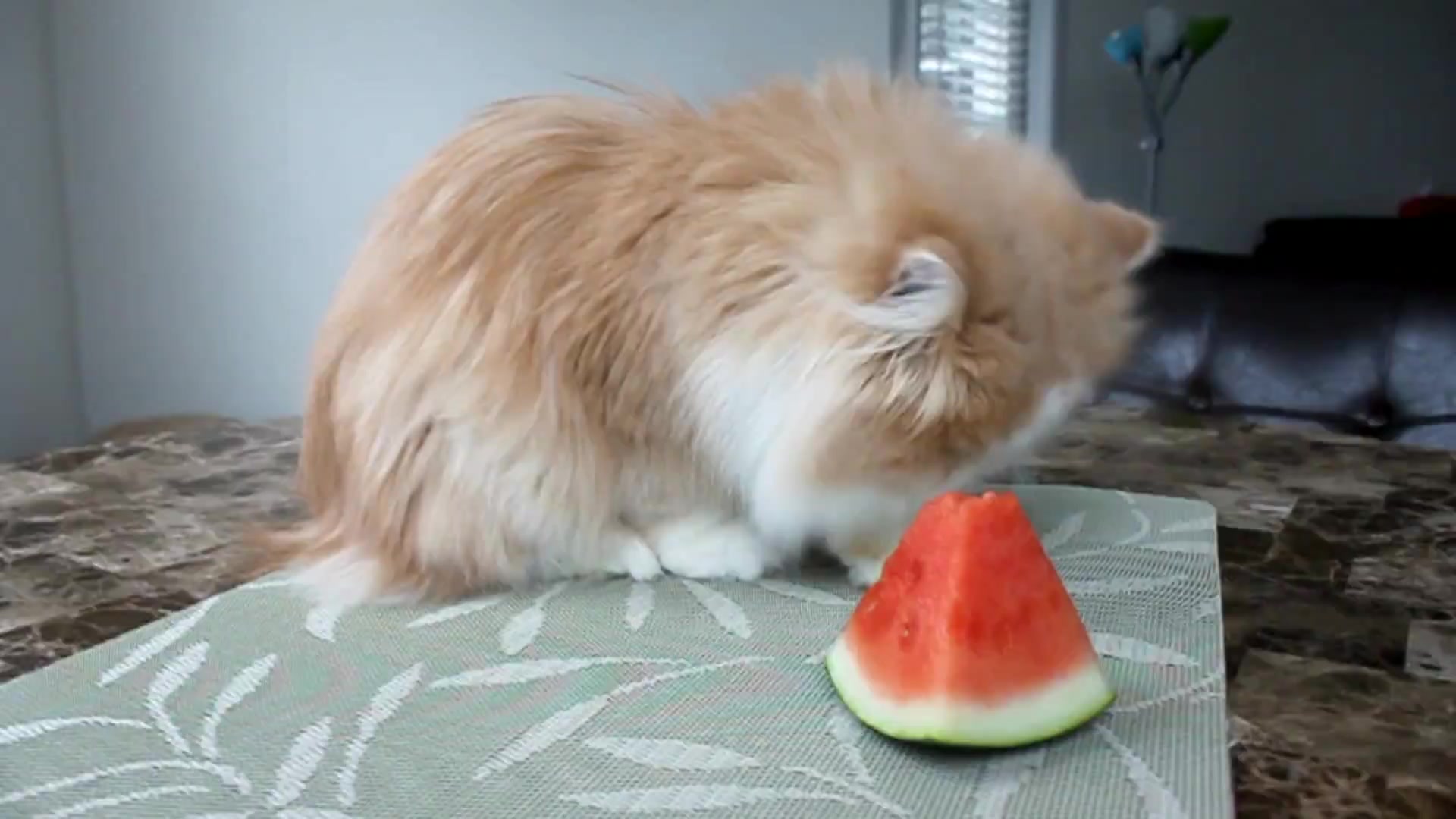}
				\includegraphics[width=0.15\textwidth]{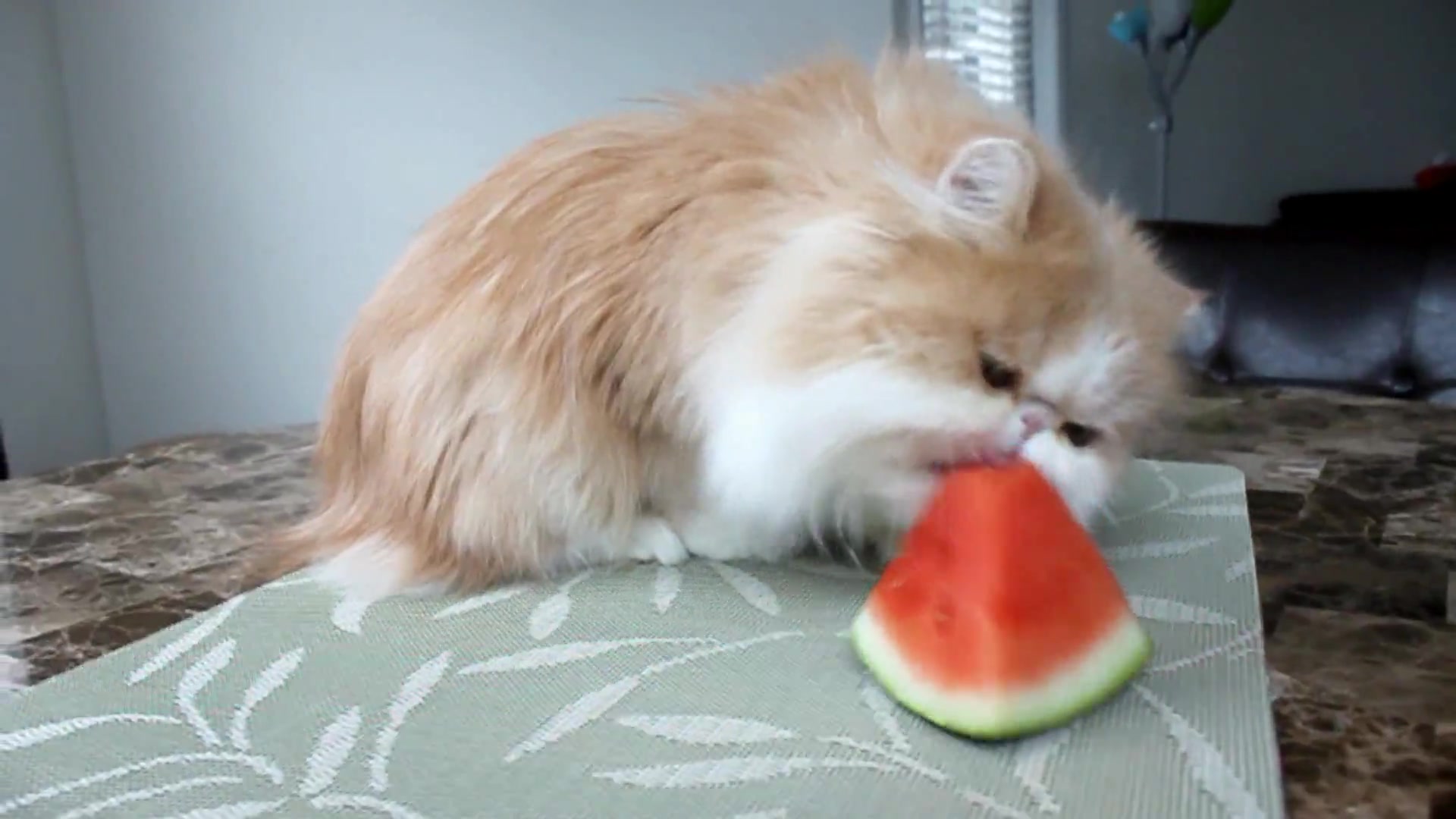}
				\includegraphics[width=0.15\textwidth]{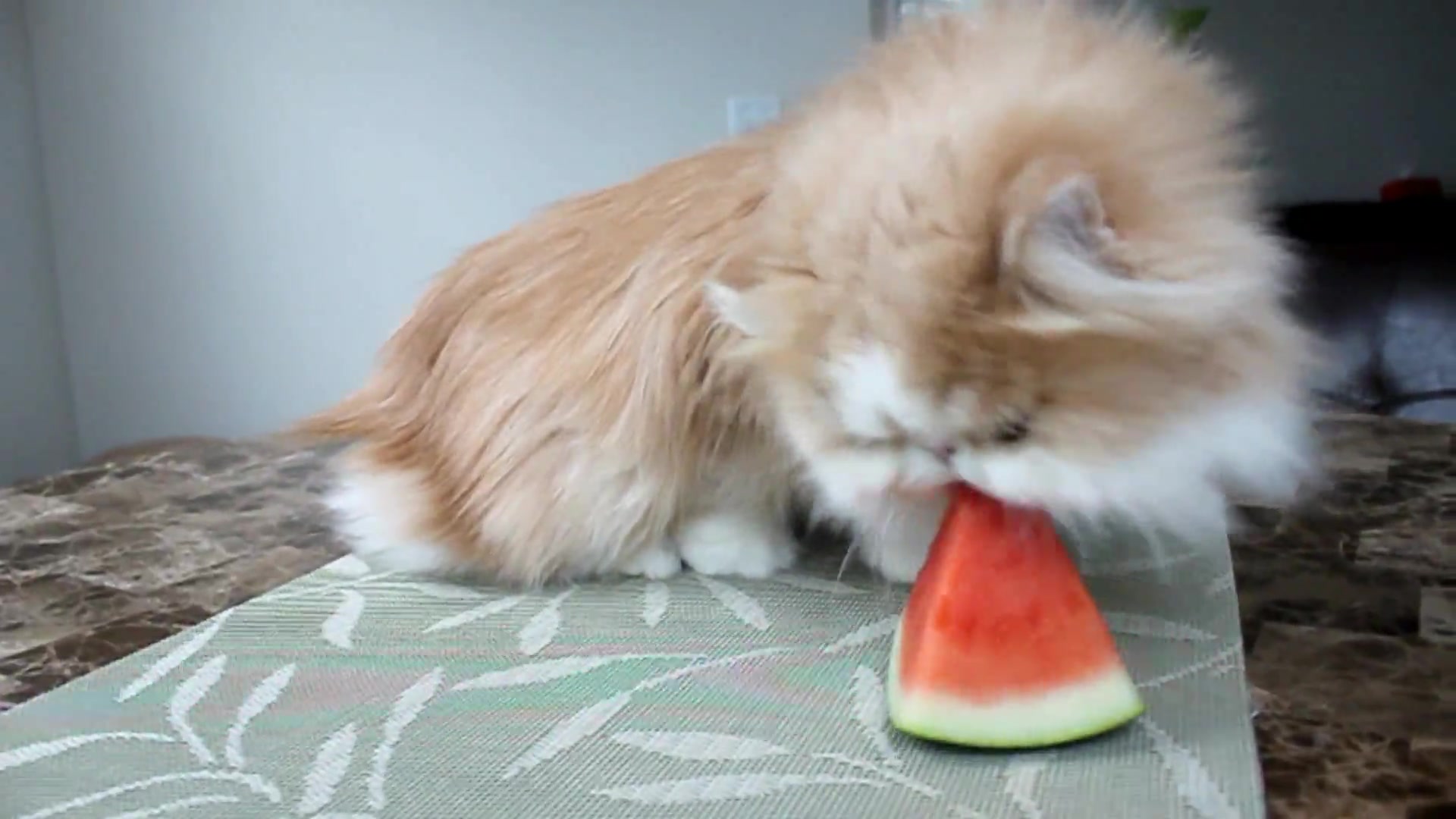}
				\end{tabular}
				&
				\begin{tabular}[c]{@{}l@{}l@{}}
				GT: \textit{A cat is eating a small wedge of watermelon} \\  FGM: A person is cutting the water.\\  
				LSTM-YT: A woman is eating an egg.  \\AlexNet: A baby is drinking. \\
				MM-VDN: \textbf{A cat is eating.}\end{tabular}
				\\ \cline{2-2}
				\begin{tabular}{c}
				\includegraphics[width=0.15\textwidth]{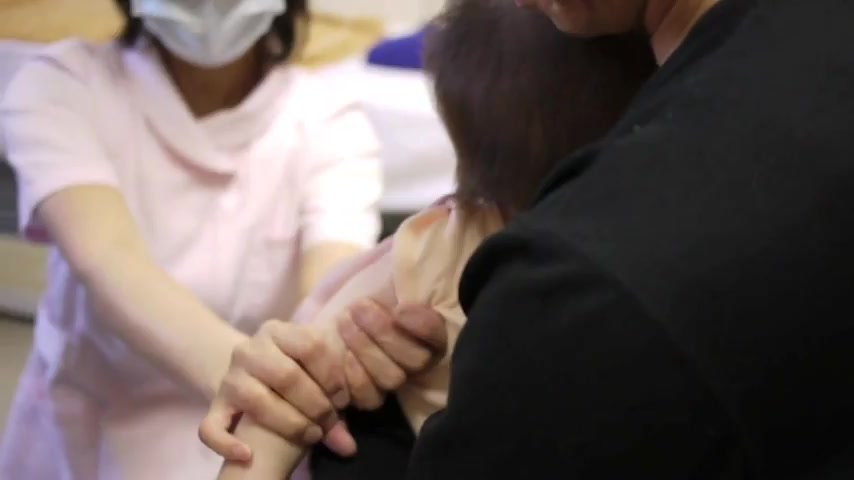}
				\includegraphics[width=0.15\textwidth]{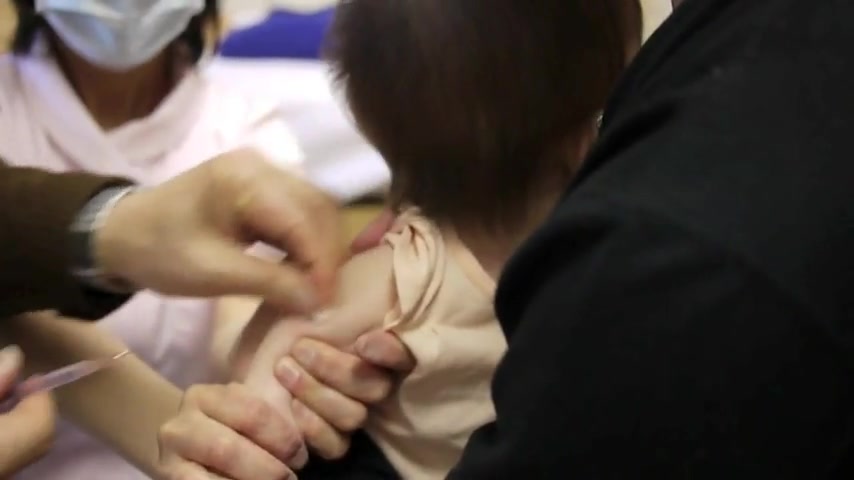}
				\includegraphics[width=0.15\textwidth]{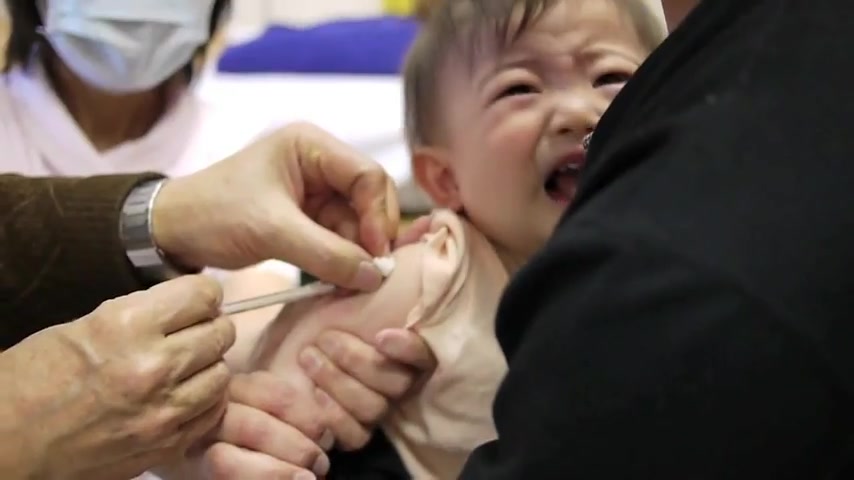}
				\includegraphics[width=0.15\textwidth]{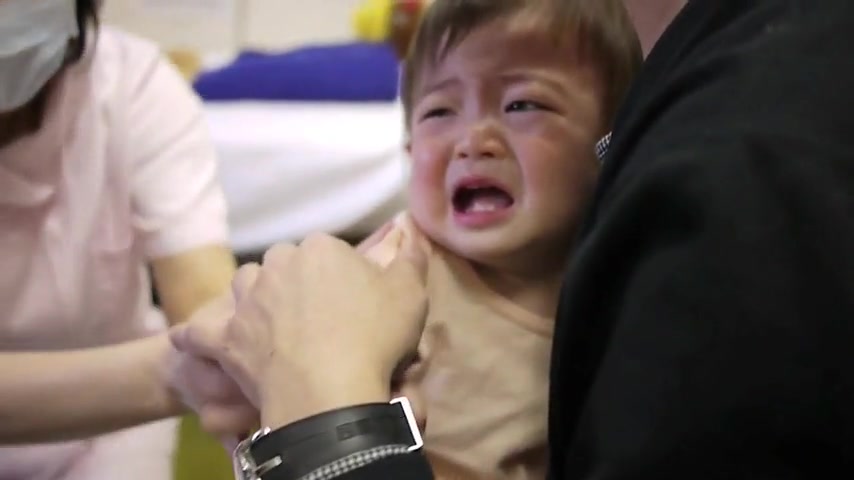}
				\end{tabular}
				&
				\begin{tabular}[c]{@{}l@{}l@{}}
				GT: \textit{A doctor gives a shot to a baby} \\  FGM: A  person is cutting a person.\\  
				LSTM-YT: A  man is putting a piece of hands.  \\AlexNet: A man is putting a stick. \\
				MM-VDN: \textbf{A  man} is playing with \textbf{a baby}. \end{tabular}
				\\ \hline
				\bf MM-VDN \textbf{makes errors} & \\ \hline 
                \\
				\begin{tabular}{c}
				\includegraphics[width=0.15\textwidth]{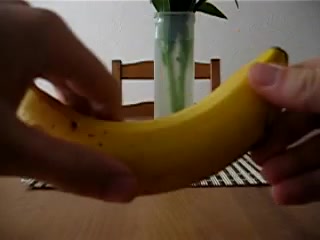}
				\includegraphics[width=0.15\textwidth]{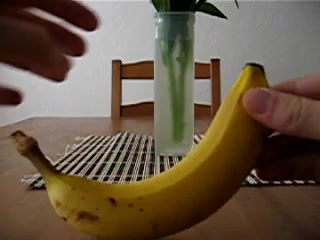}
				\includegraphics[width=0.15\textwidth]{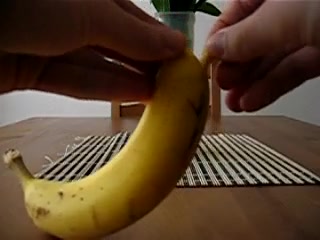}
				\includegraphics[width=0.15\textwidth]{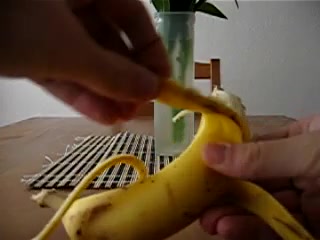}
				\end{tabular}
				&
				\begin{tabular}[c]{@{}l@{}l@{}}
				GT: \textit{A person is peeling a banana from the bottom.} \\ FGM: A person is cutting an onion.\\  
				LSTM-YT: A woman is doing a card.  \\AlexNet: A woman is peeling a apple.\\
				MM-VDN: A woman is cutting a potato. \end{tabular}
				\\  \cline{2-2}
				\begin{tabular}{c}
				\includegraphics[width=0.15\textwidth]{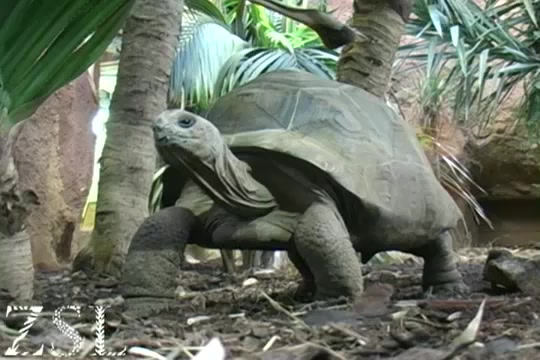}
				\includegraphics[width=0.15\textwidth]{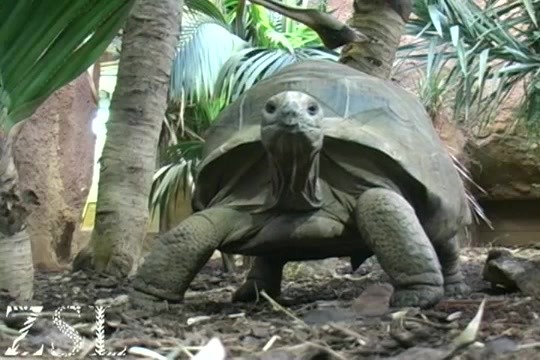}
				\includegraphics[width=0.15\textwidth]{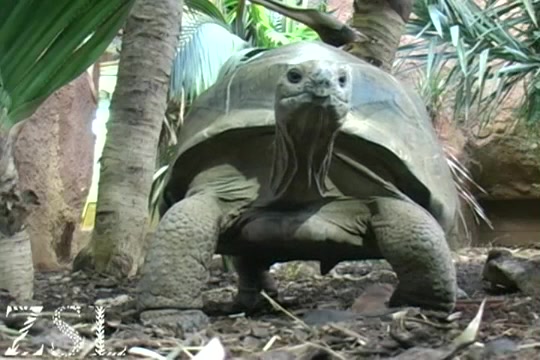}
				\includegraphics[width=0.15\textwidth]{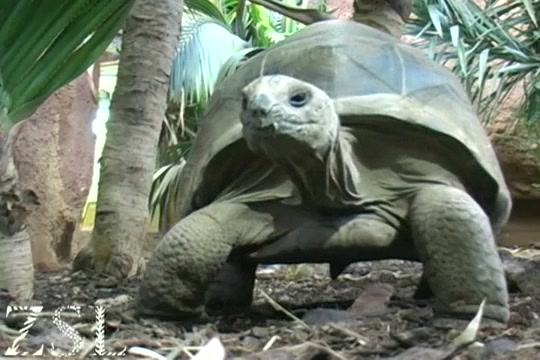}
				\end{tabular}
				&
				\begin{tabular}[c]{@{}l@{}l@{}}
				GT: \textit{A turtle is walking.} \\ FGM: A person is walking in the food.\\  
				LSTM-YT: A baby is eating.  \\ AlexNet: A panda is eating. \\
				MM-VDN: A panda is walking.\end{tabular}
				\\ \hline
			\end{tabular}
		\caption{Some example videos and predicted sentences. (GT) shows the ground truth sentences;
		 (FGM) is the factor graph model in~\cite{thomason:coling14}; (LSTM-YT) is the model in \cite{video_lstm:naccl}; (AlexNet) is the basic CNN + LSTM model; (MM-VDN) is our model. 
		 The top section shows videos where our MM-VDN model improves on others; the middle section shows videos where our model predicts part of the sentence correctly; the bottom shows videos where our model makes errors.}
		\label{tab:example_videos}
		\end{center}
		\end{table*}

%% file: supplement.tex
\begin{table*}[t]
	\begin{center}
	\caption{Visualization of the heat maps and the histograms of multiscale feature value for sampled frames. }
		\begin{tabular}{c}
			\hline \\
			\begin{tabular}{c}
			\includegraphics[width=0.15\textwidth]{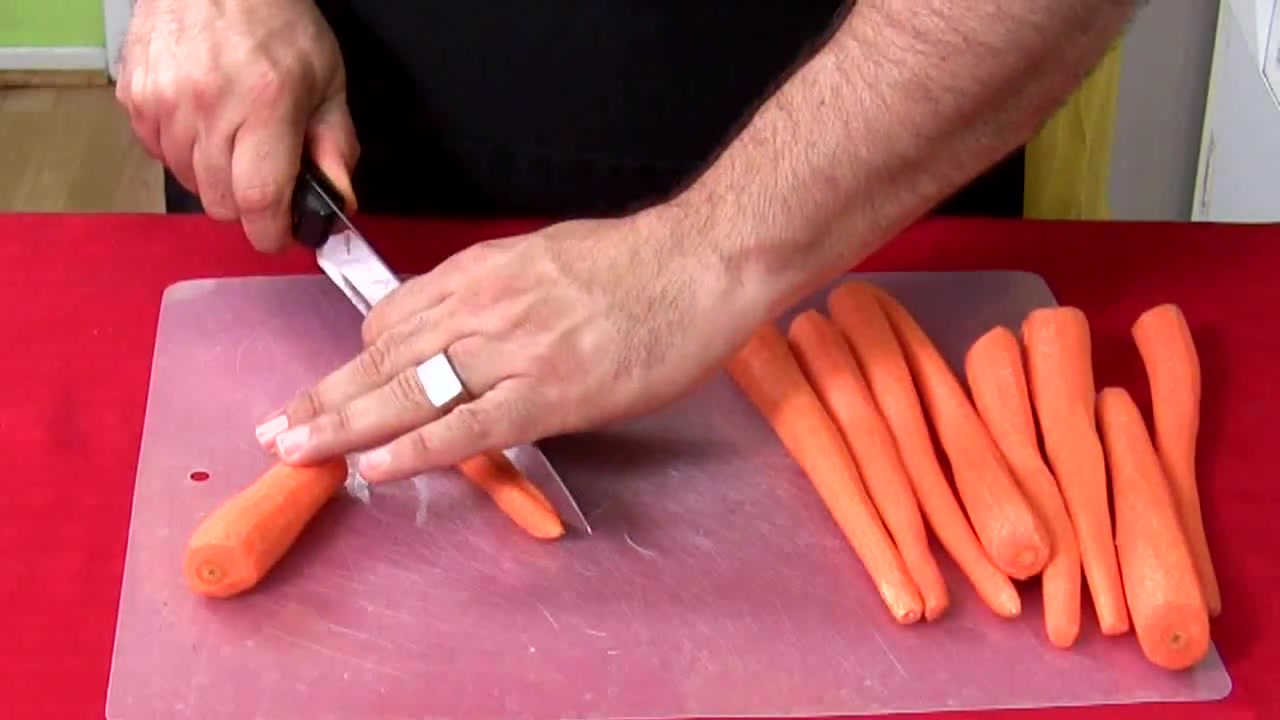}
			\includegraphics[width=0.15\textwidth]{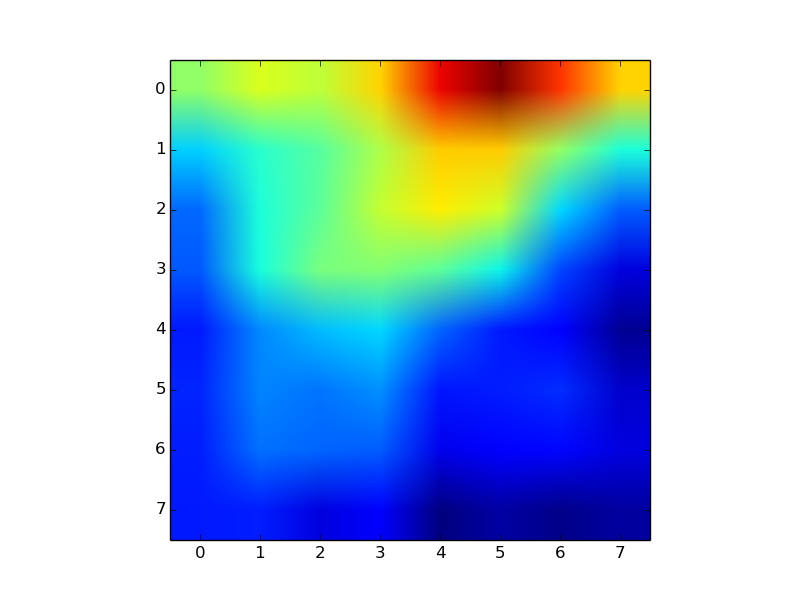}
			\includegraphics[width=0.15\textwidth]{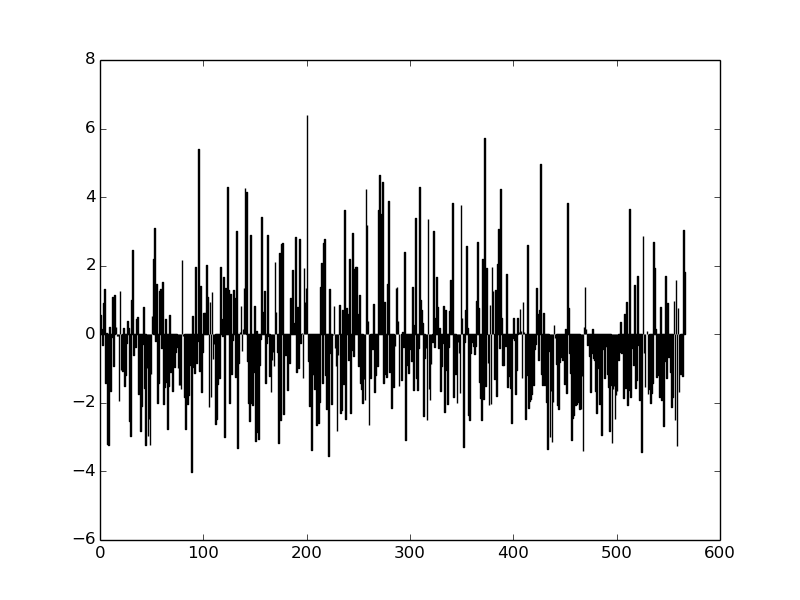}
			\includegraphics[width=0.15\textwidth]{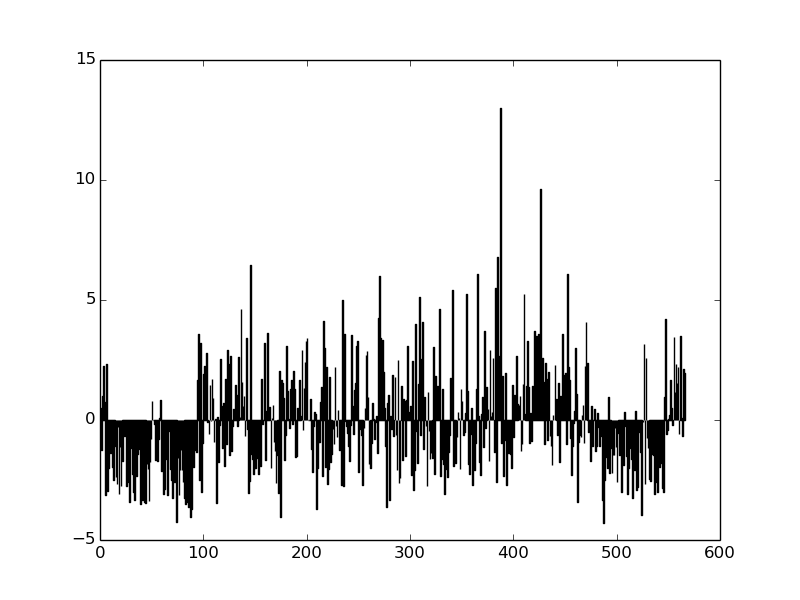}
			\includegraphics[width=0.15\textwidth]{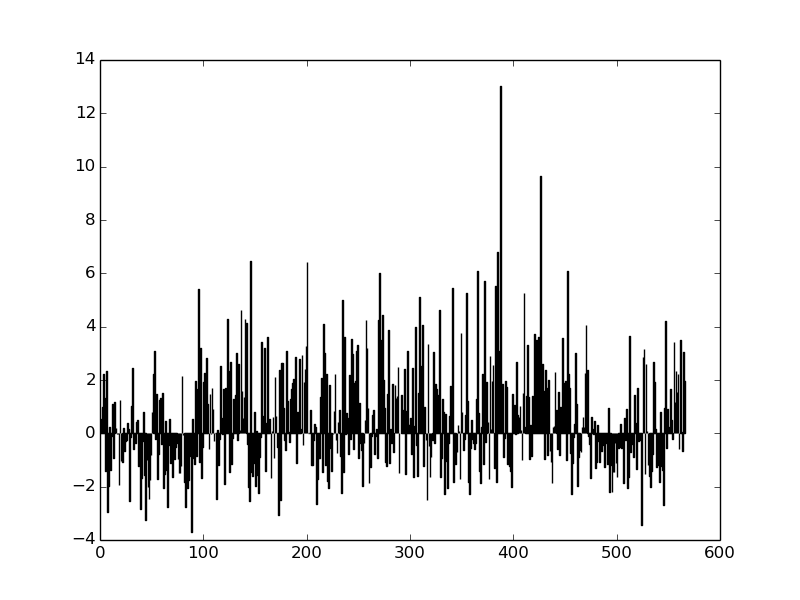}
			\end{tabular} \\
			\begin{tabular}{cc}
			GT: \textit{A man is cutting a carrot.}  &
			MM-VDN: \textit{A man is slicing a carrot.}
			\end{tabular} \\~\\
			
			\begin{tabular}{c}
			\includegraphics[width=0.15\textwidth]{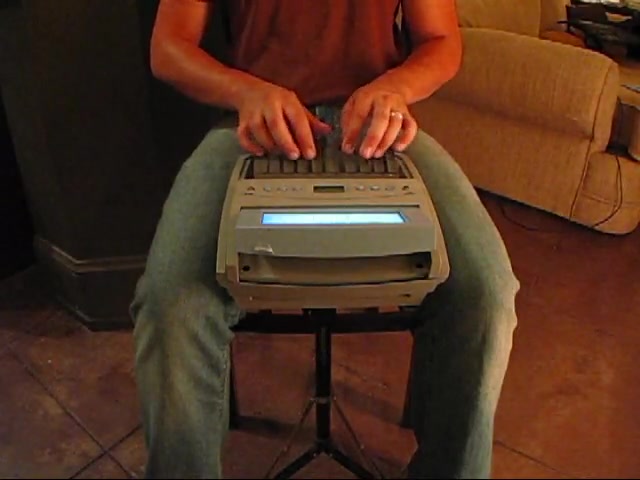}
			\includegraphics[width=0.15\textwidth]{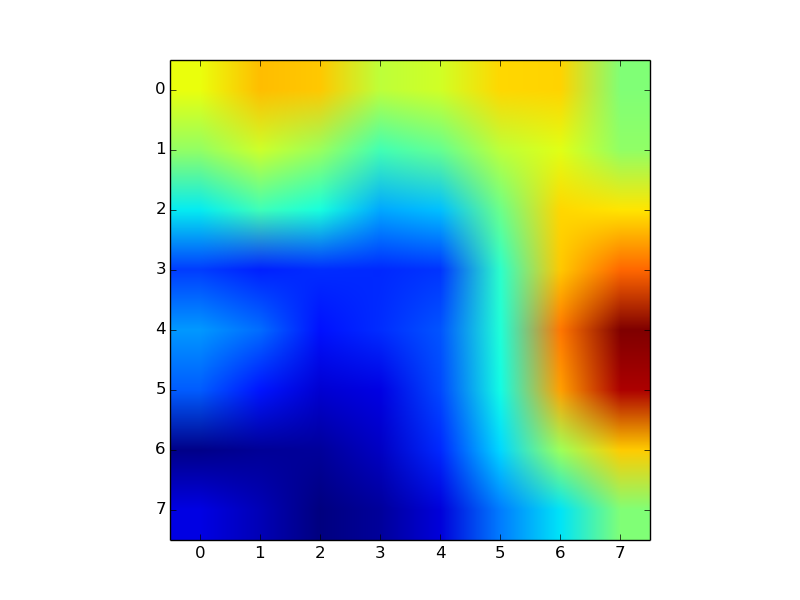}
			\includegraphics[width=0.15\textwidth]{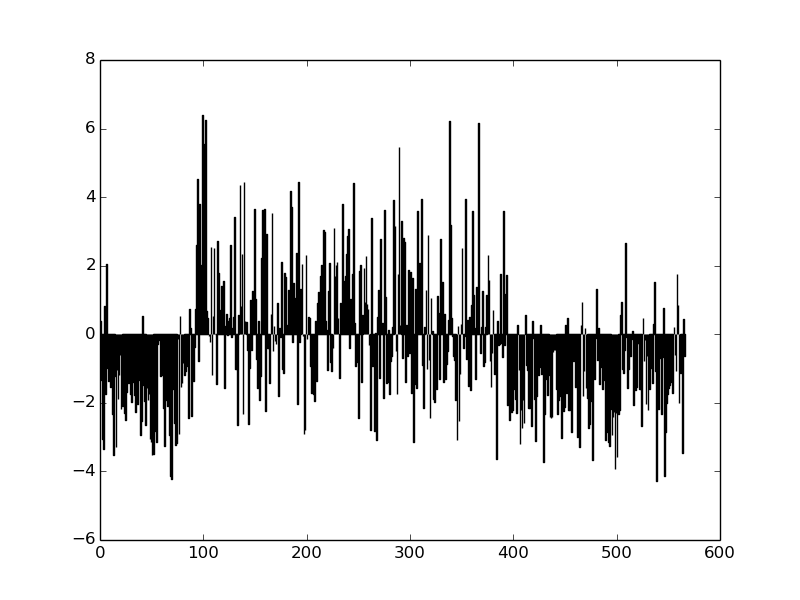}
			\includegraphics[width=0.15\textwidth]{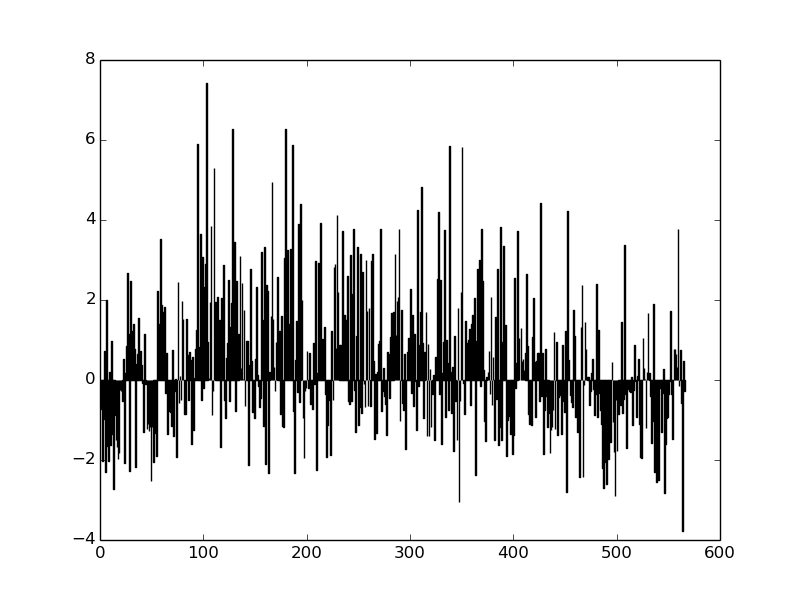}
			\includegraphics[width=0.15\textwidth]{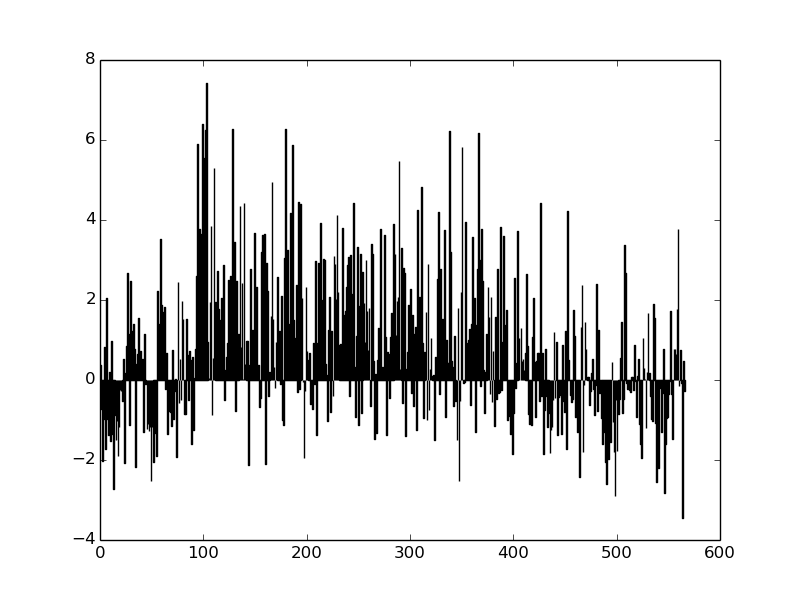}
			\end{tabular} \\			
			\begin{tabular}{cc}
			GT: \textit{A man is tying in the machine.}  &
			MM-VDN: \textit{A woman is typing.}
			\end{tabular} \\~\\
			
			\begin{tabular}{c}
			\includegraphics[width=0.15\textwidth]{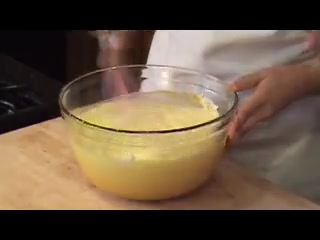}
			\includegraphics[width=0.15\textwidth]{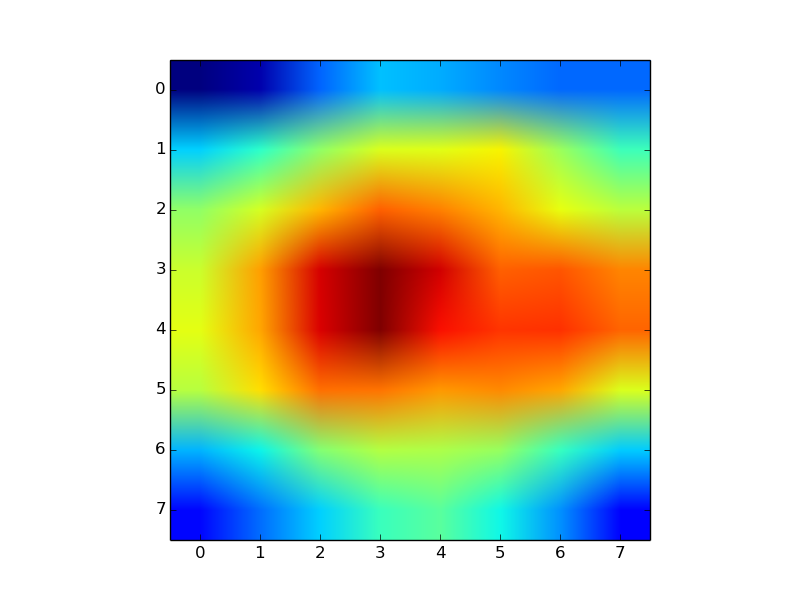}
			\includegraphics[width=0.15\textwidth]{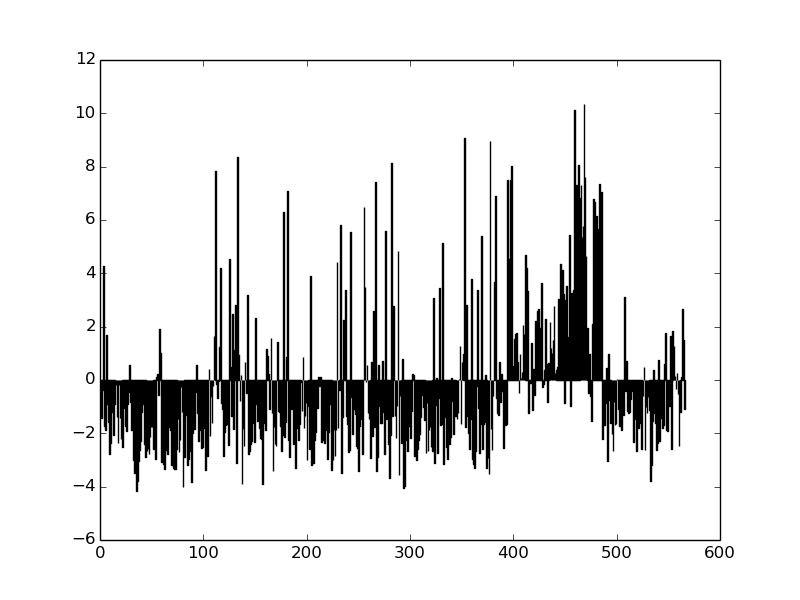}
			\includegraphics[width=0.15\textwidth]{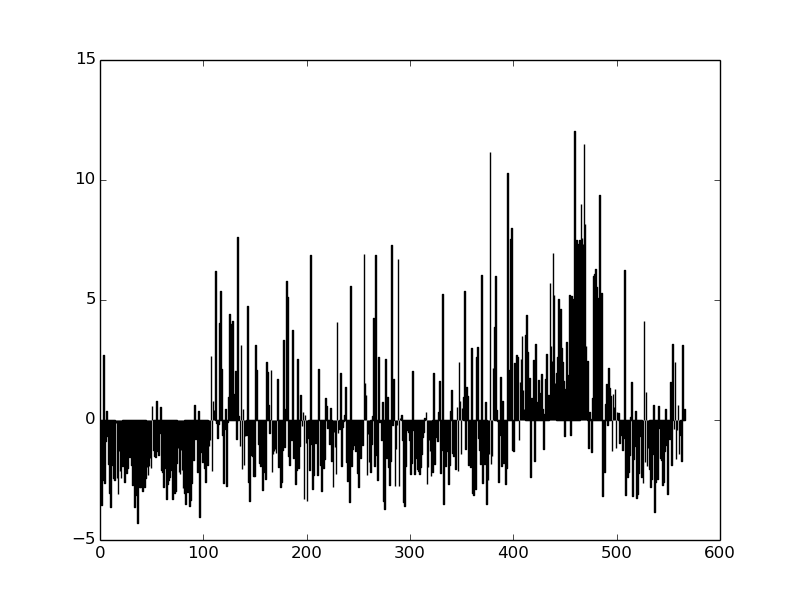}
			\includegraphics[width=0.15\textwidth]{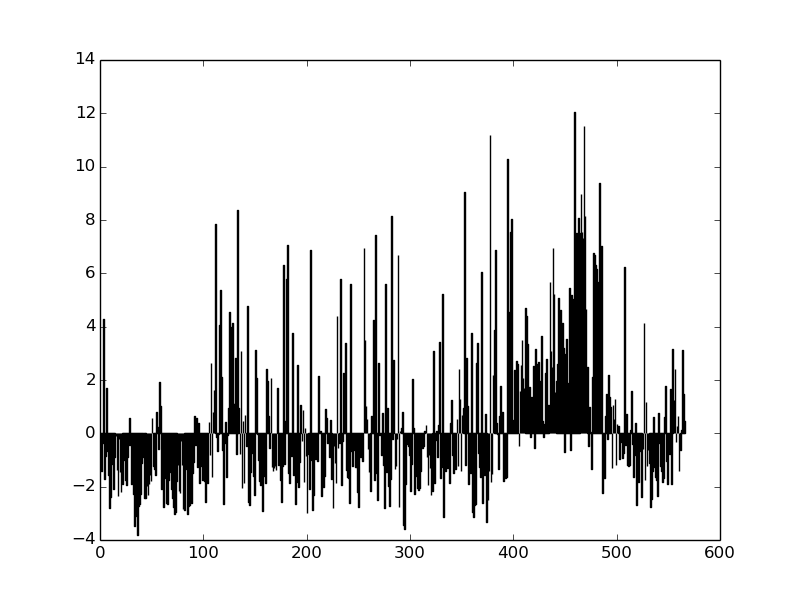}
			\end{tabular} \\			
			\begin{tabular}{cc}
			GT: \textit{A woman is mixing eggs.}  &
			MM-VDN: \textit{A woman is mixing an egg.}
			\end{tabular} \\~\\
			
			\begin{tabular}{c}
			\includegraphics[width=0.15\textwidth]{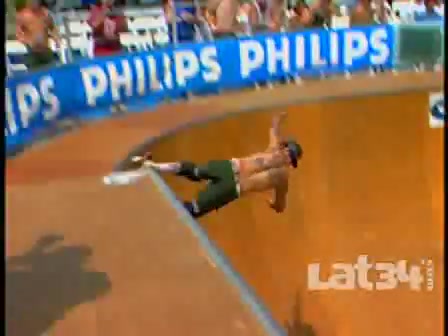}
			\includegraphics[width=0.15\textwidth]{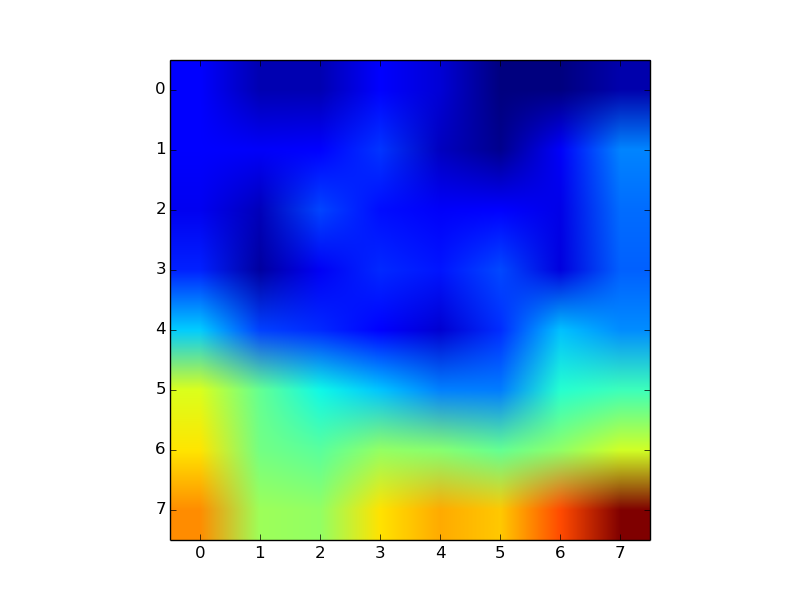}
			\includegraphics[width=0.15\textwidth]{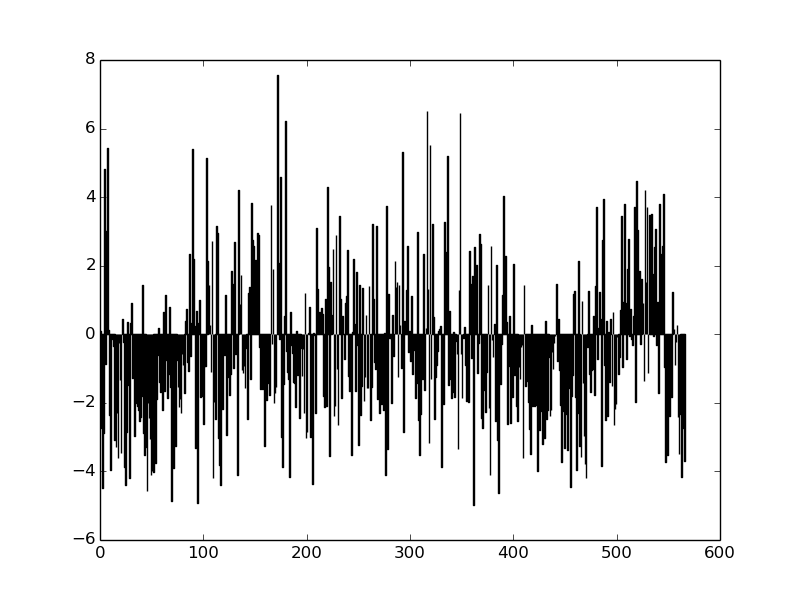}
			\includegraphics[width=0.15\textwidth]{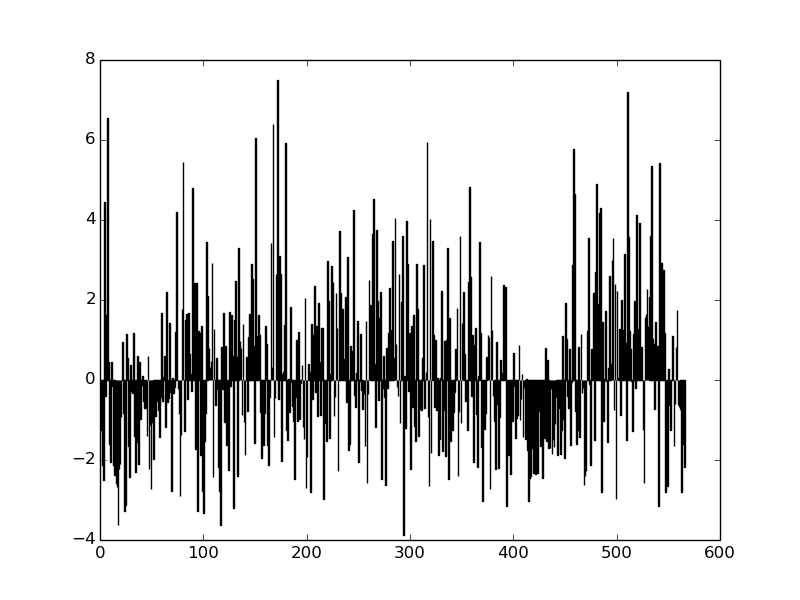}
			\includegraphics[width=0.15\textwidth]{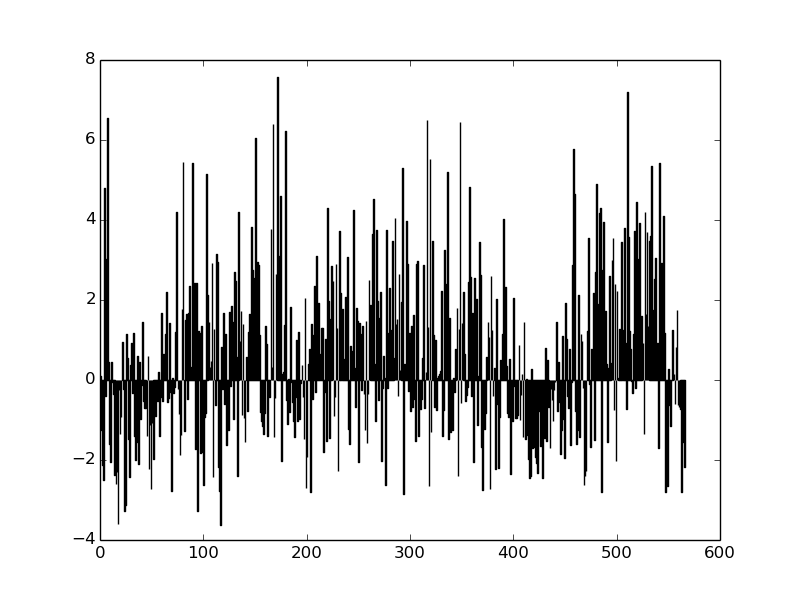}
			\end{tabular} \\			
			\begin{tabular}{cc}
			GT: \textit{A man is riding a skateboard}  &
			MM-VDN: \textit{A man is doing a skateboard.}
			\end{tabular} \\~\\
			
			\begin{tabular}{c}
			\includegraphics[width=0.15\textwidth]{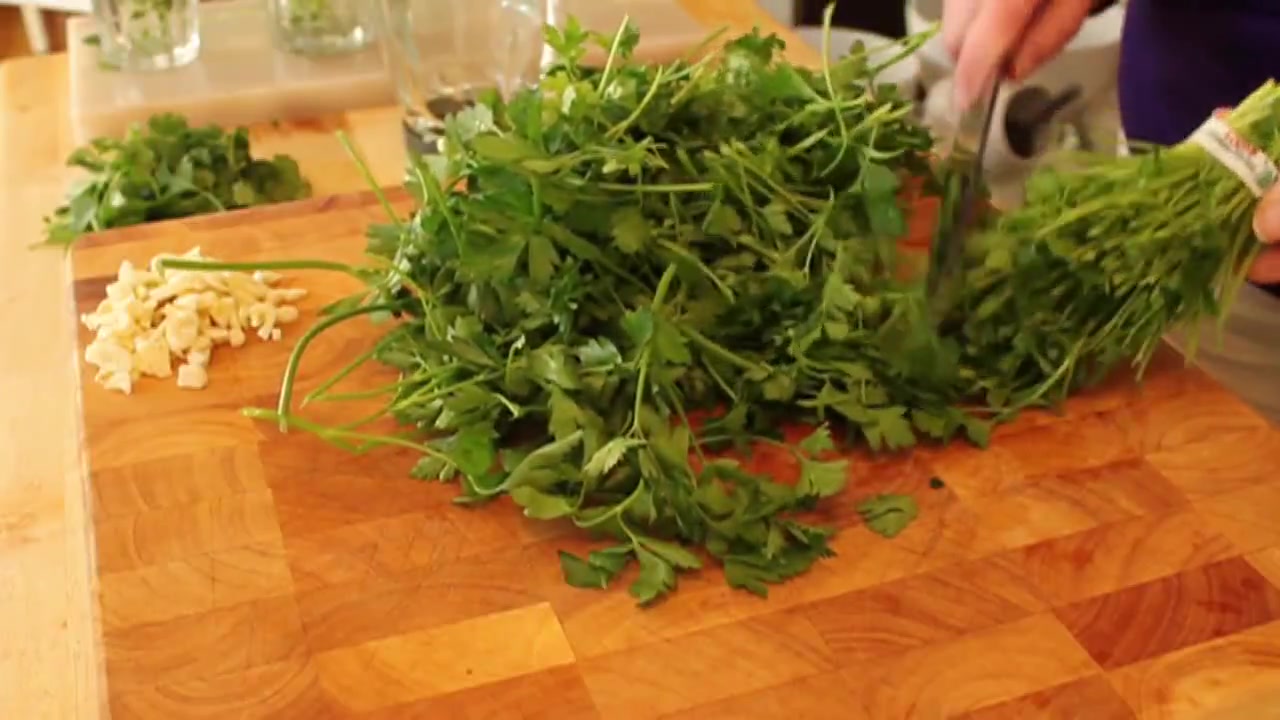}
			\includegraphics[width=0.15\textwidth]{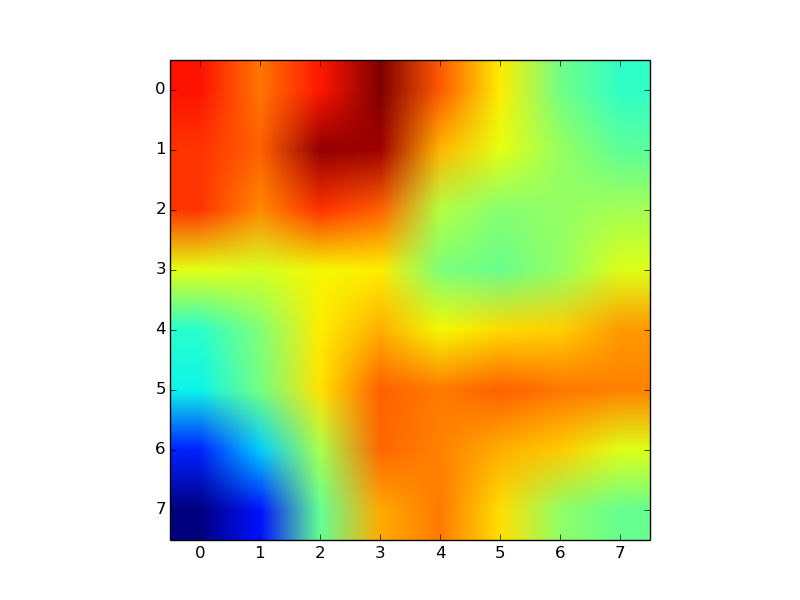}
			\includegraphics[width=0.15\textwidth]{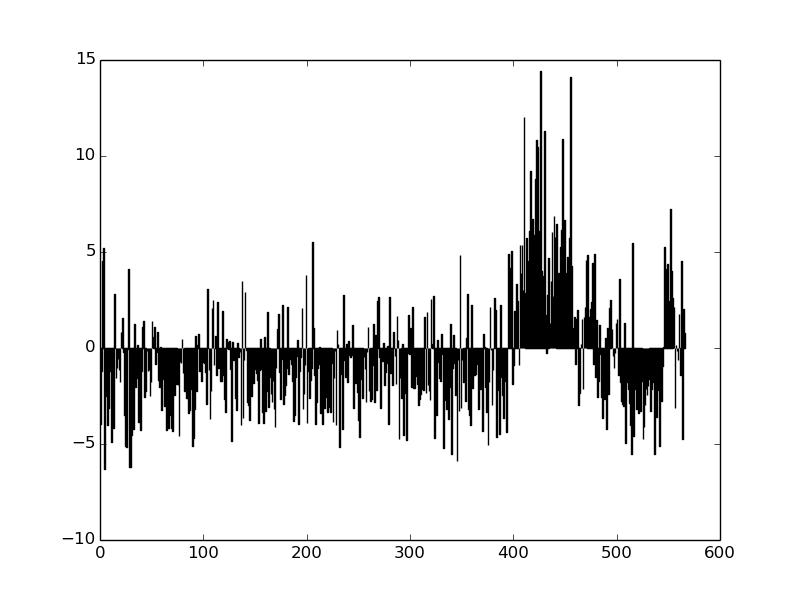}
			\includegraphics[width=0.15\textwidth]{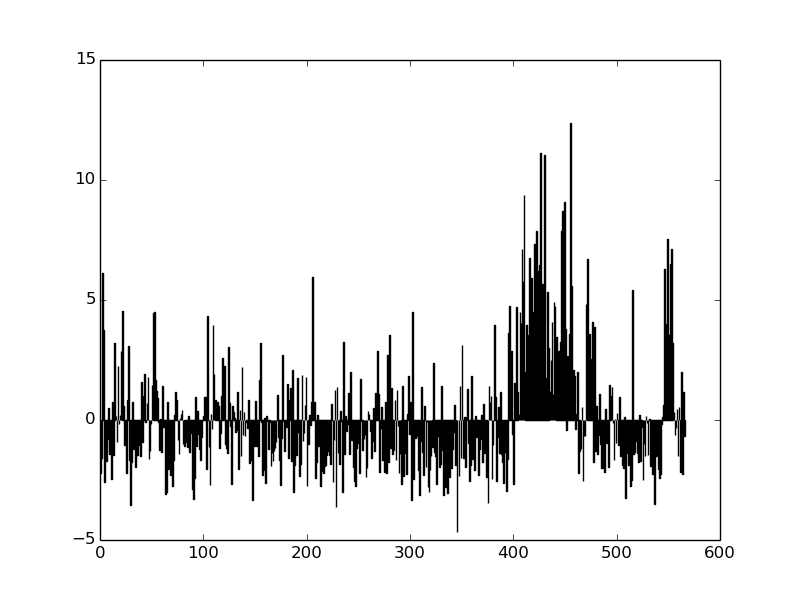}
			\includegraphics[width=0.15\textwidth]{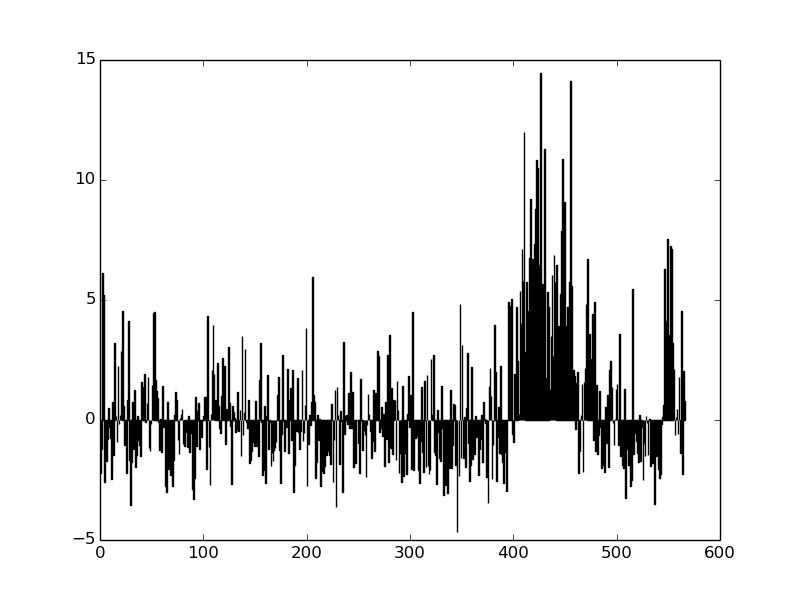}
			\end{tabular} \\			
			\begin{tabular}{cc}
			GT: \textit{A woman is cutting parsley.}  &
			MM-VDN: \textit{A woman is cutting a vegetable.}
			\end{tabular} \\~\\
			
			\begin{tabular}{c}
			\includegraphics[width=0.15\textwidth]{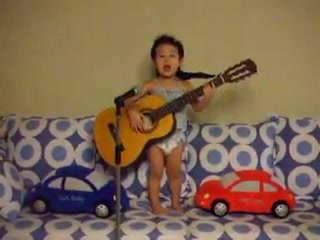}
			\includegraphics[width=0.15\textwidth]{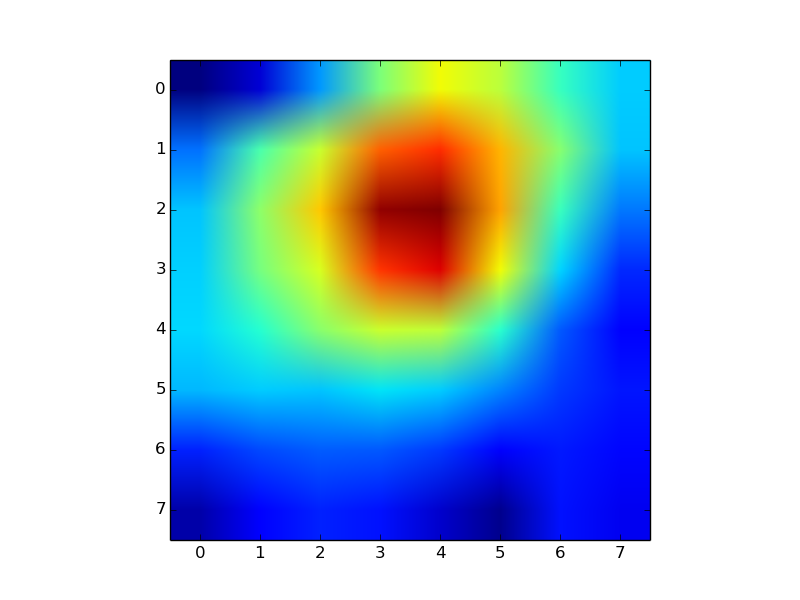}
			\includegraphics[width=0.15\textwidth]{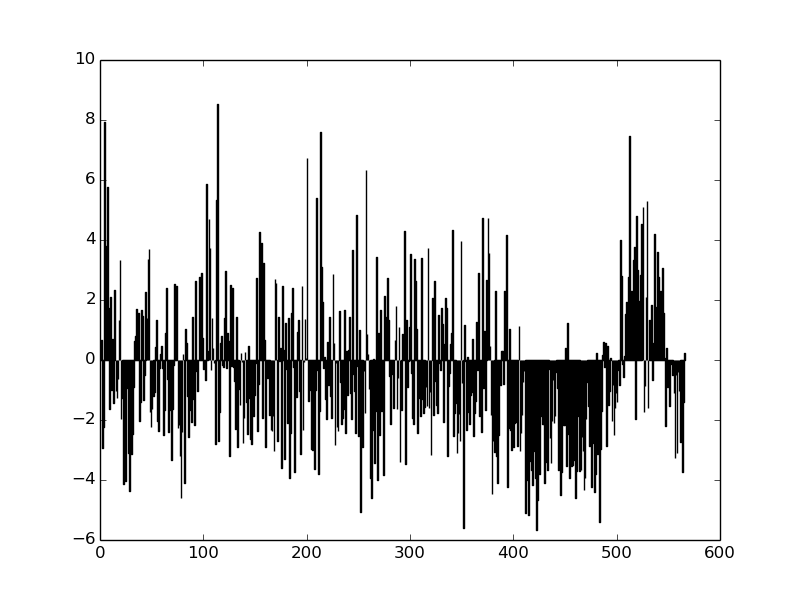}
			\includegraphics[width=0.15\textwidth]{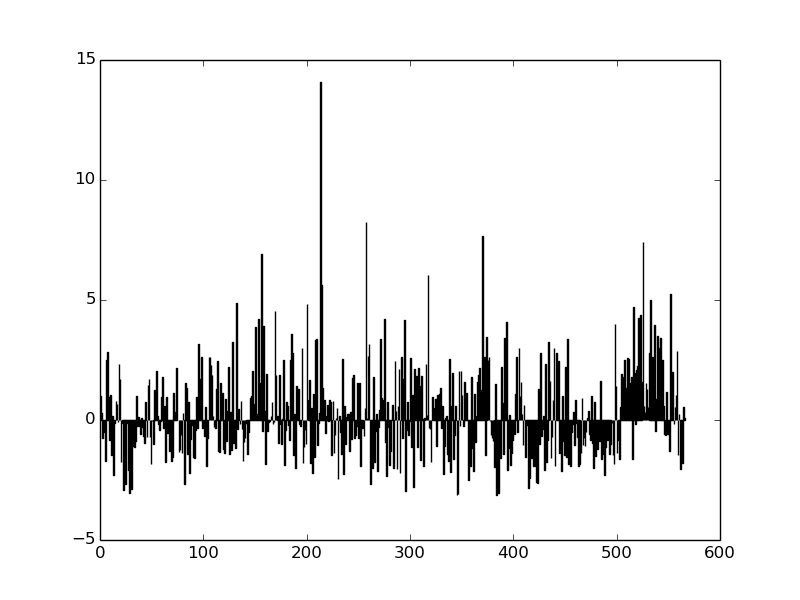}
			\includegraphics[width=0.15\textwidth]{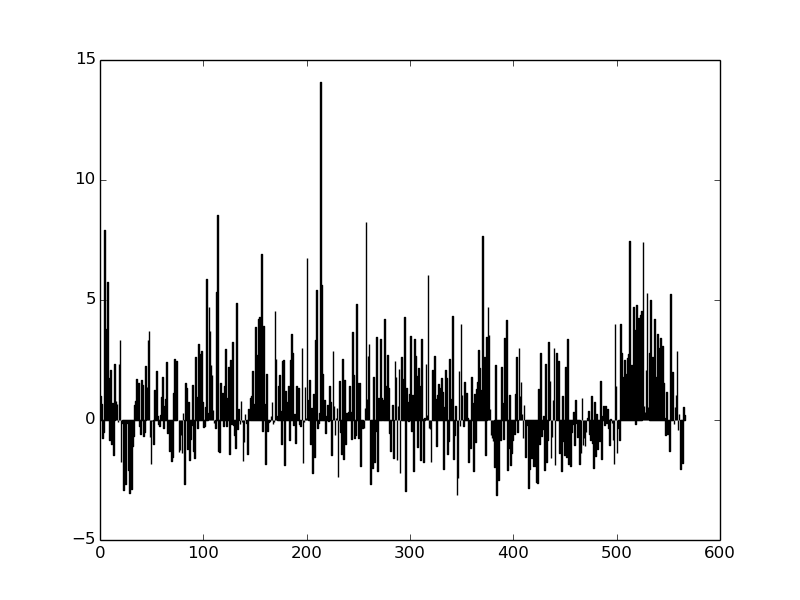}
			\end{tabular} \\			
			\begin{tabular}{cc}
			GT: \textit{A boy is playing a guitar.}  &
			MM-VDN: \textit{A man is playing a guitar.}
			\end{tabular} \\~\\
			
			\begin{tabular}{c}
			\includegraphics[width=0.15\textwidth]{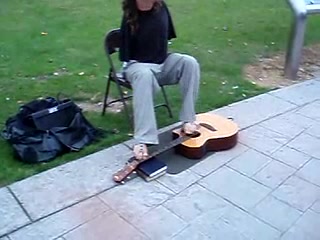}
			\includegraphics[width=0.15\textwidth]{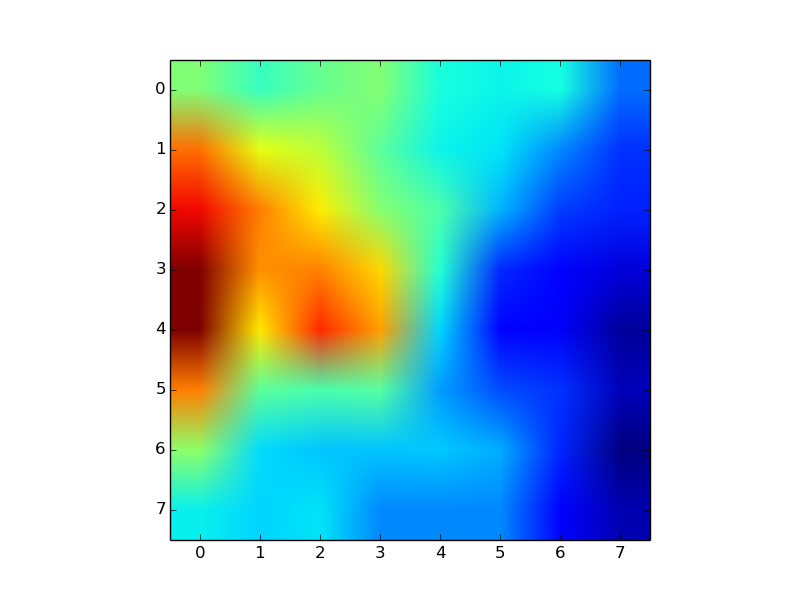}
			\includegraphics[width=0.15\textwidth]{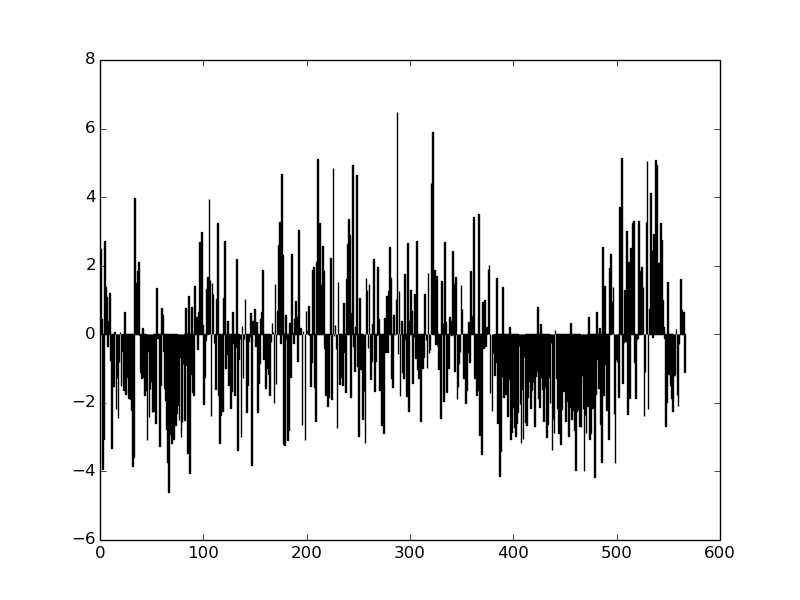}
			\includegraphics[width=0.15\textwidth]{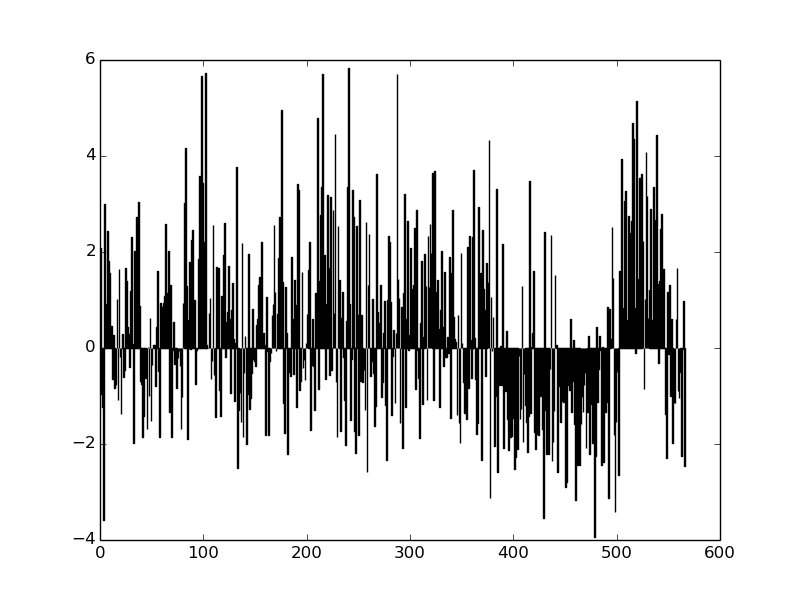}
			\includegraphics[width=0.15\textwidth]{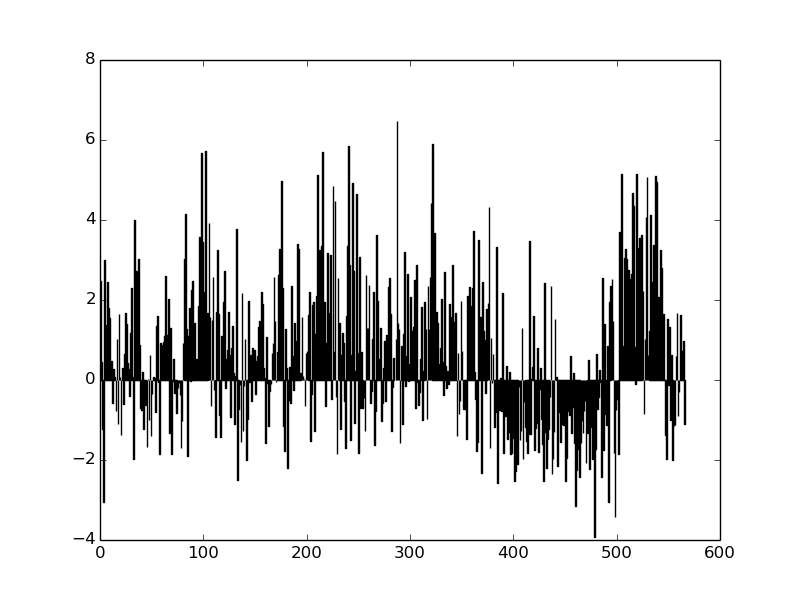}
			\end{tabular} \\			
			\begin{tabular}{cc}
			GT: \textit{A man is playing a guitar with his feet.}  &
			MM-VDN: \textit{A man is playing a guitar.}
			\end{tabular}  
		\end{tabular}
	\label{tab:visua1}
	\end{center}
\end{table*}

\begin{table*}[t]
	\begin{center}
		\begin{tabular}{c}
			\begin{tabular}{c}
			\includegraphics[width=0.15\textwidth]{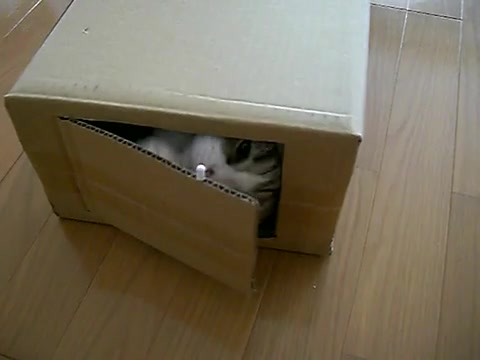}
			\includegraphics[width=0.15\textwidth]{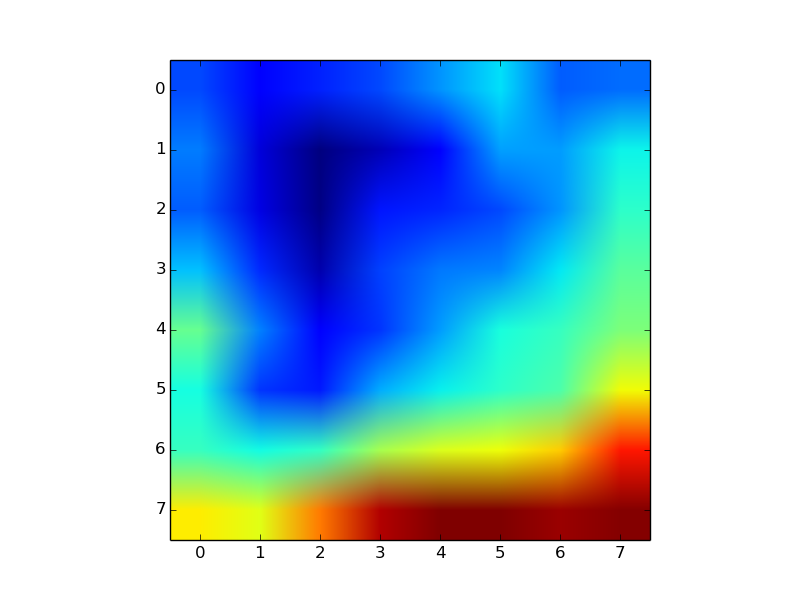}
			\includegraphics[width=0.15\textwidth]{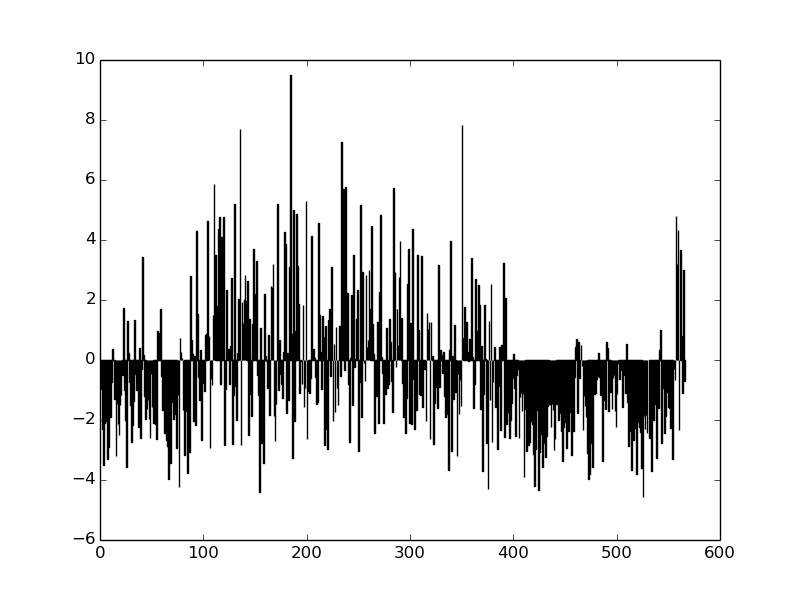}
			\includegraphics[width=0.15\textwidth]{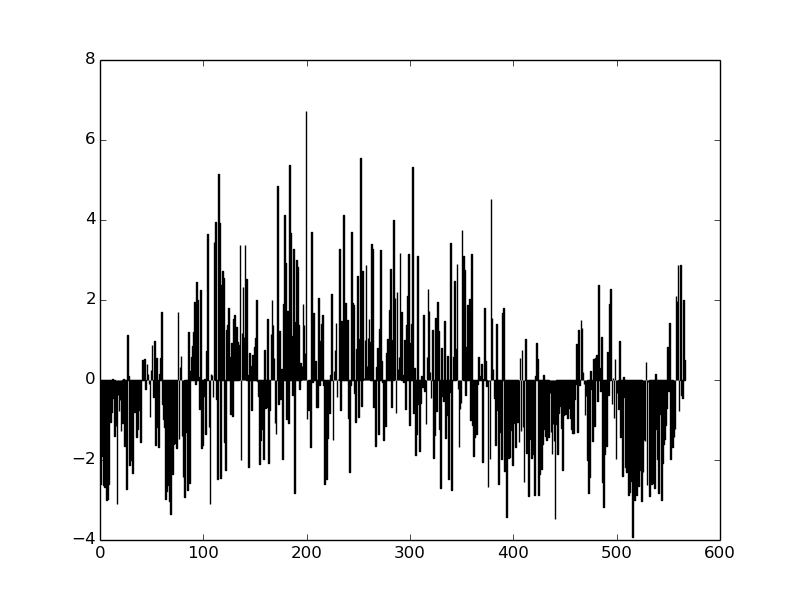}
			\includegraphics[width=0.15\textwidth]{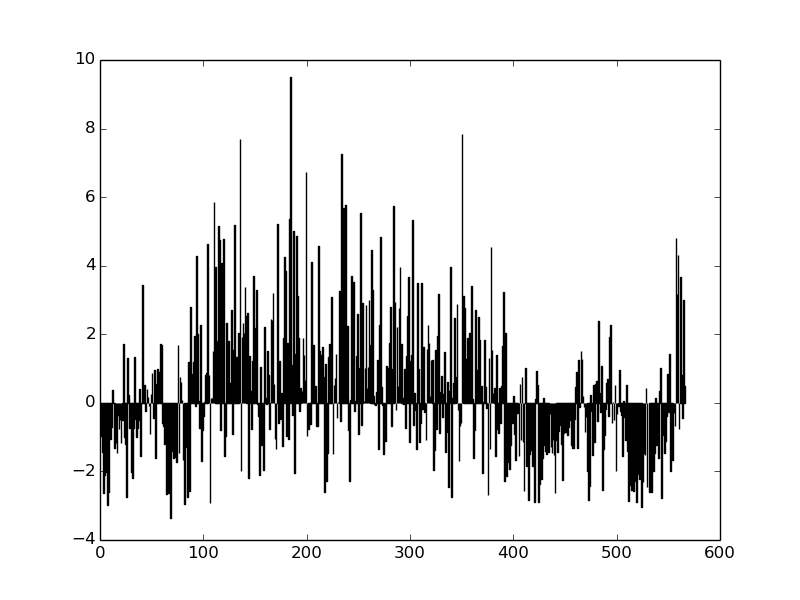}
			\end{tabular} \\			
			\begin{tabular}{cc}
			GT: \textit{A cat is playing in a box.}  &
			MM-VDN: \textit{A cat is playing.}
			\end{tabular}  \\~\\
			
			\begin{tabular}{c}
			\includegraphics[width=0.15\textwidth]{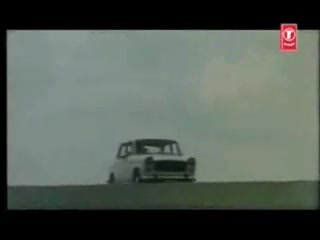}
			\includegraphics[width=0.15\textwidth]{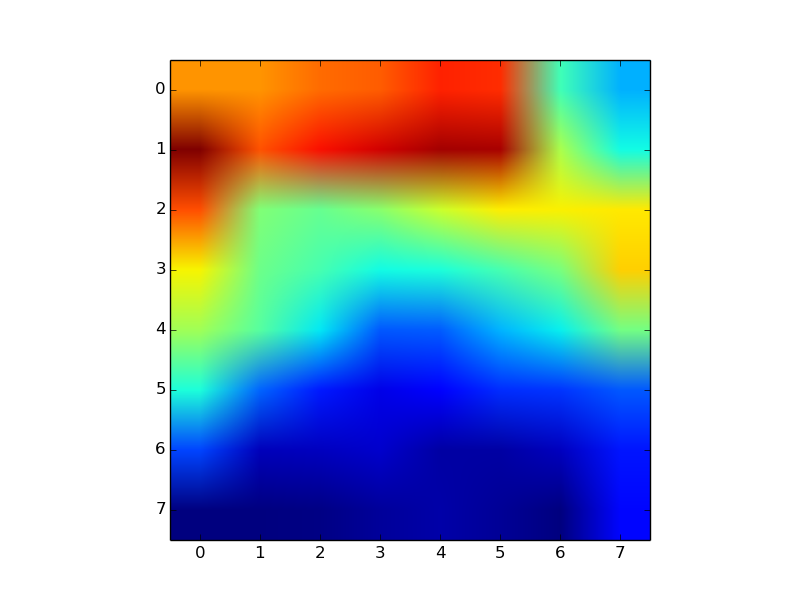}
			\includegraphics[width=0.15\textwidth]{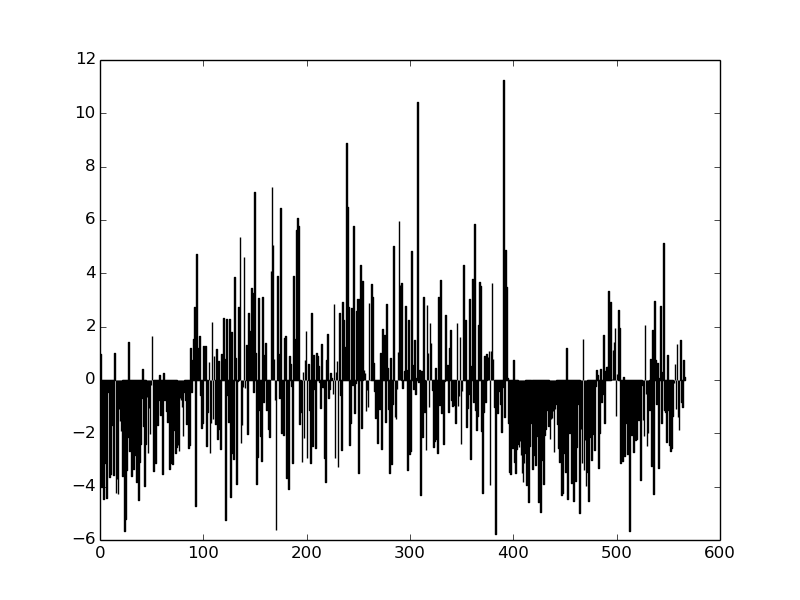}
			\includegraphics[width=0.15\textwidth]{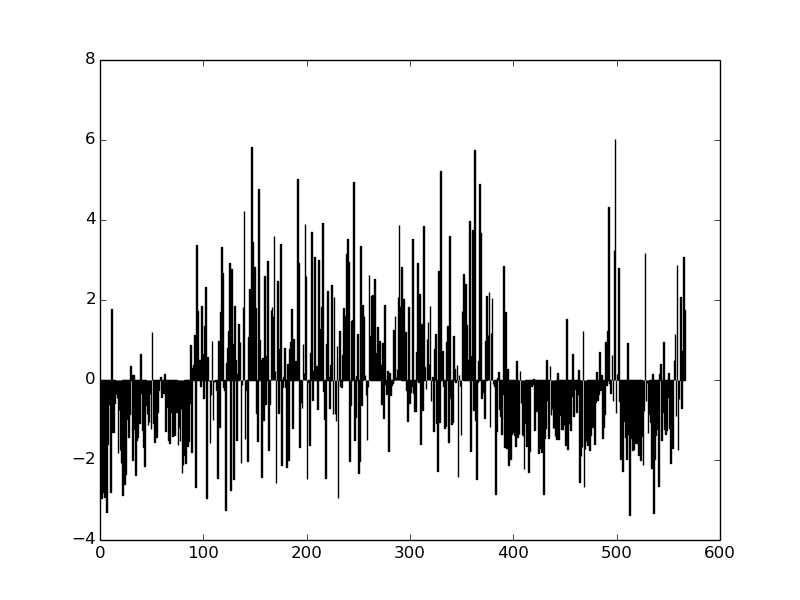}
			\includegraphics[width=0.15\textwidth]{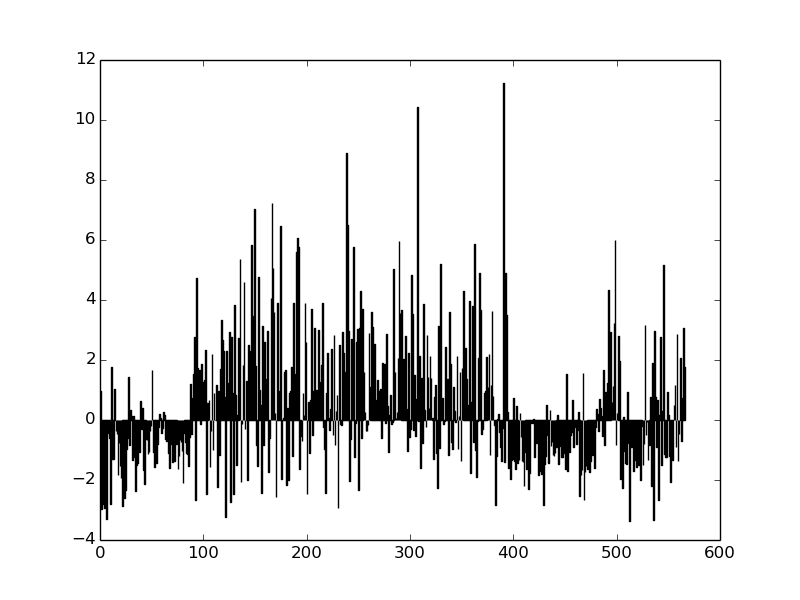}
			\end{tabular} \\			
			\begin{tabular}{cc}
			GT: \textit{A car is going down a hill.}  &
			MM-VDN: \textit{A car is driving down the road.}
			\end{tabular}  \\~\\
			
			\begin{tabular}{c}
			\includegraphics[width=0.15\textwidth]{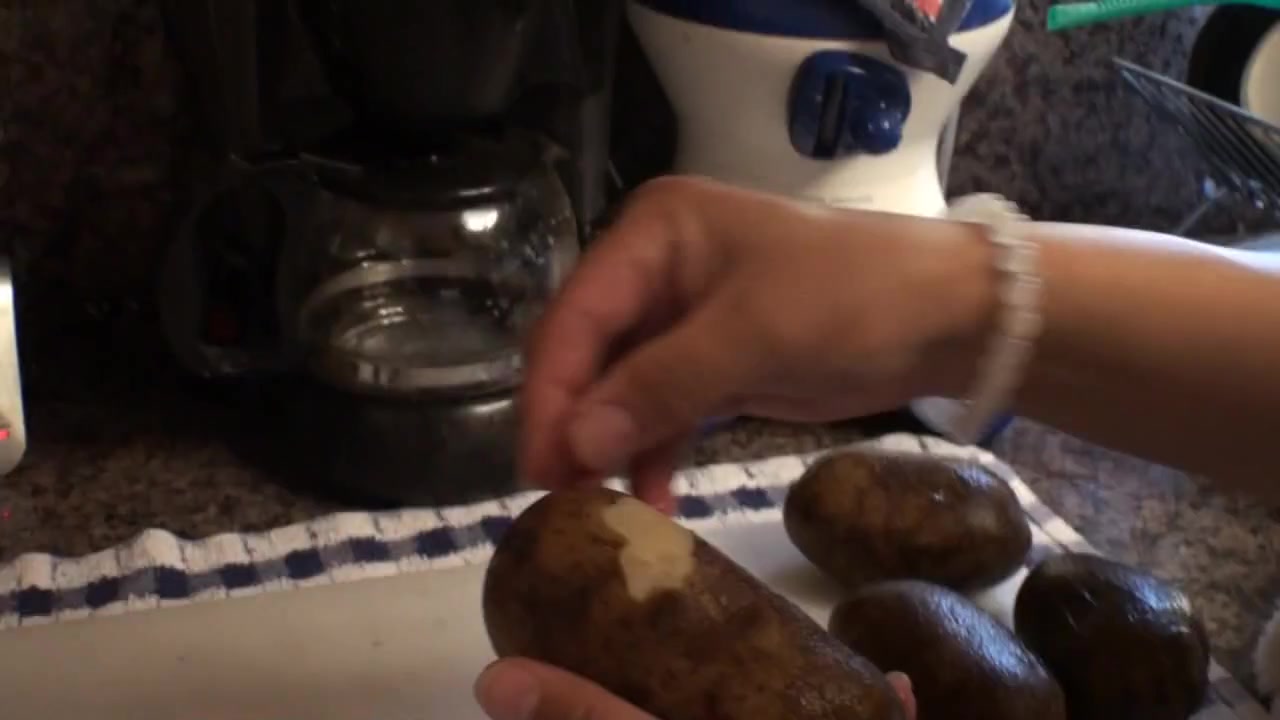}
			\includegraphics[width=0.15\textwidth]{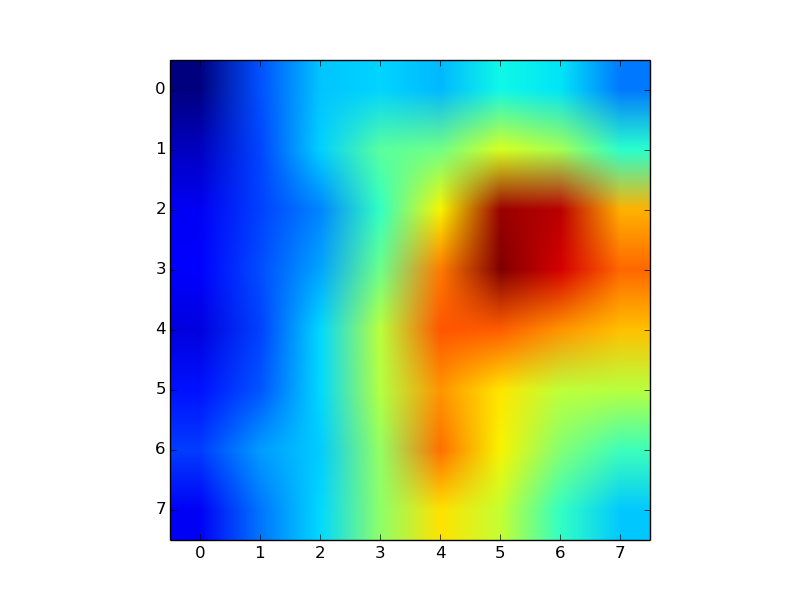}
			\includegraphics[width=0.15\textwidth]{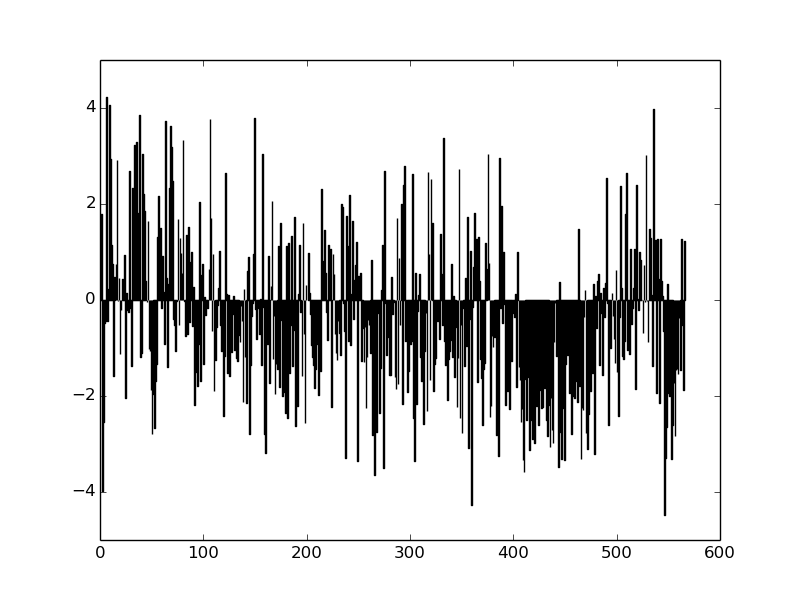}
			\includegraphics[width=0.15\textwidth]{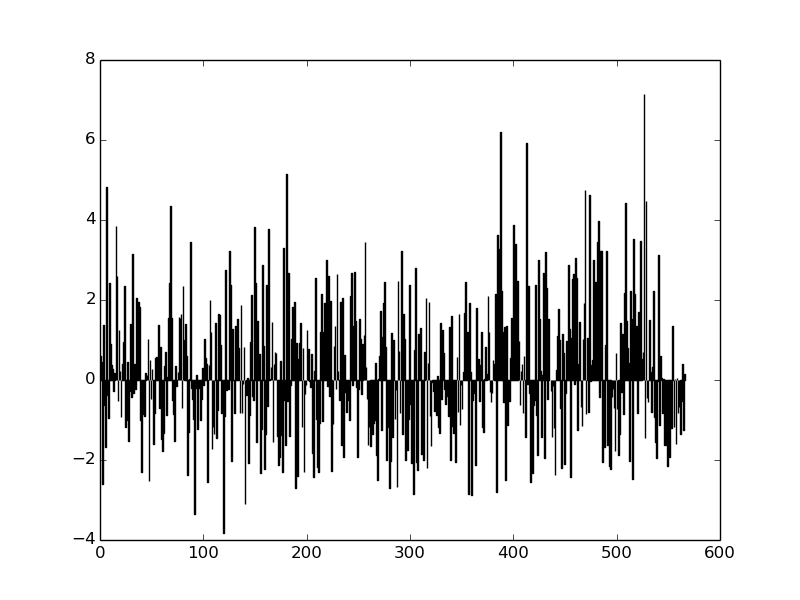}
			\includegraphics[width=0.15\textwidth]{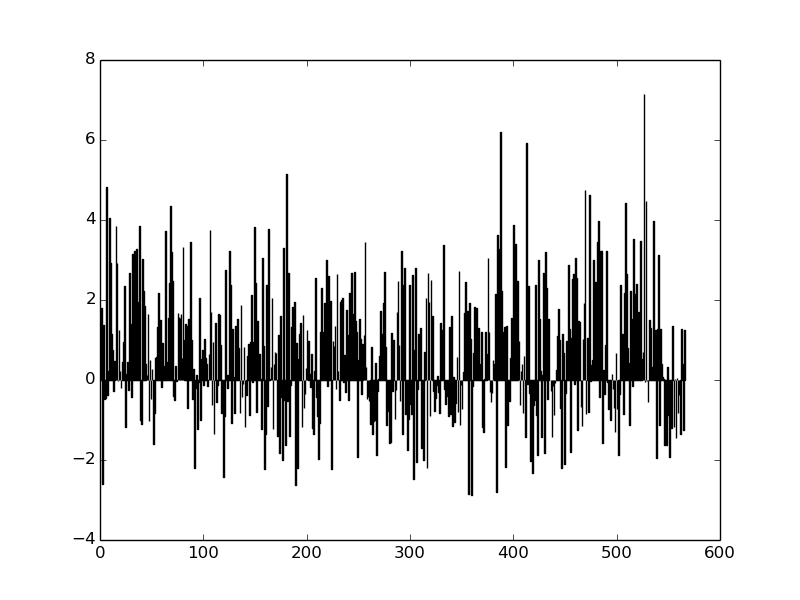}
			\end{tabular} \\			
			\begin{tabular}{cc}
			GT: \textit{A person is peeling a potato}  &
			MM-VDN: \textit{A man is peeling a potato.}
			\end{tabular}  \\~\\
			
			\begin{tabular}{c}
			\includegraphics[width=0.15\textwidth]{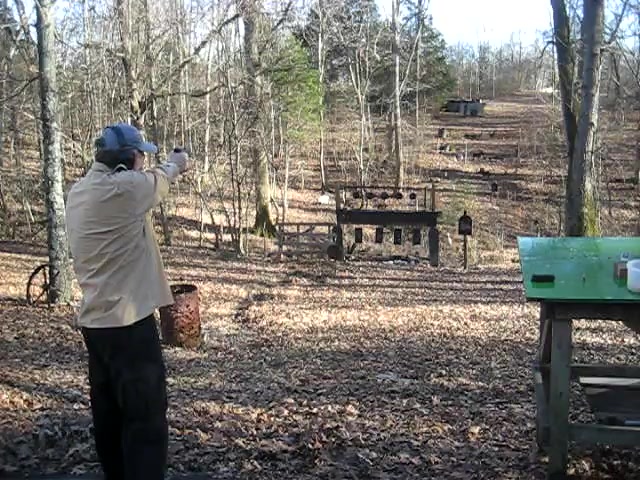}
			\includegraphics[width=0.15\textwidth]{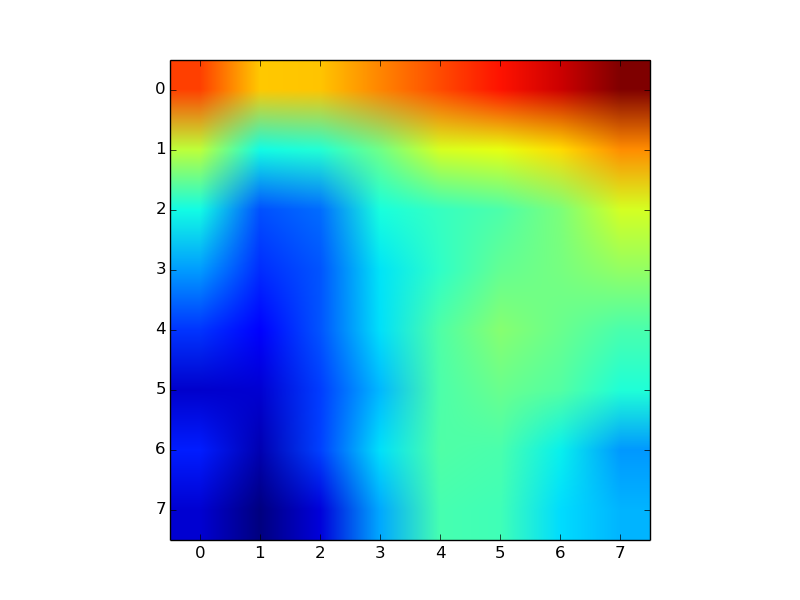}
			\includegraphics[width=0.15\textwidth]{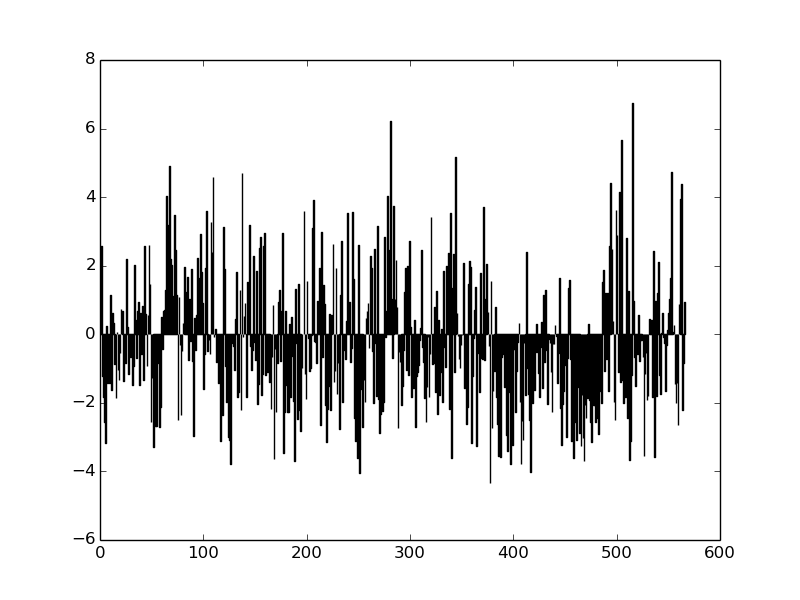}
			\includegraphics[width=0.15\textwidth]{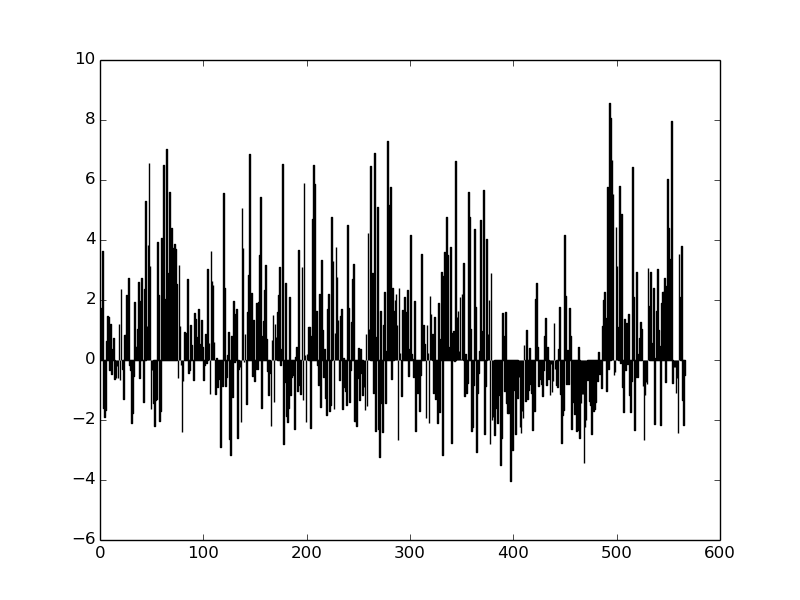}
			\includegraphics[width=0.15\textwidth]{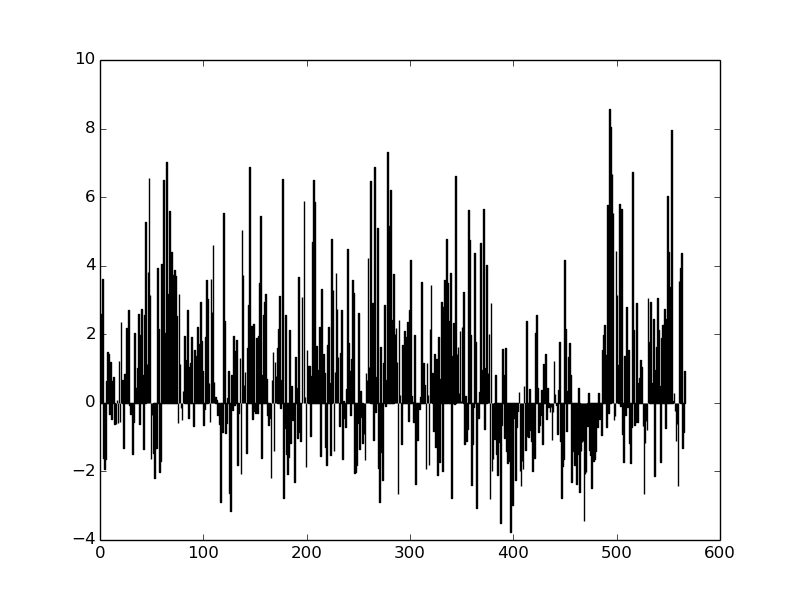}
			\end{tabular} \\			
			\begin{tabular}{cc}
			GT: \textit{A man is shooting at a target.}  &
			MM-VDN: \textit{A man is shooting a target.}
			\end{tabular}  \\~\\
			
			\begin{tabular}{c}
			\includegraphics[width=0.15\textwidth]{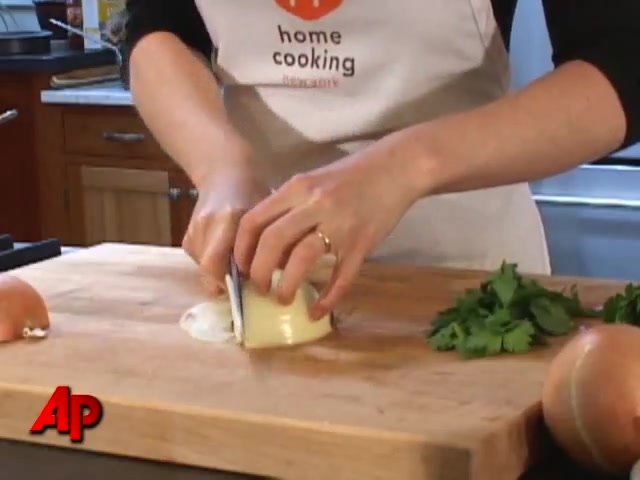}
			\includegraphics[width=0.15\textwidth]{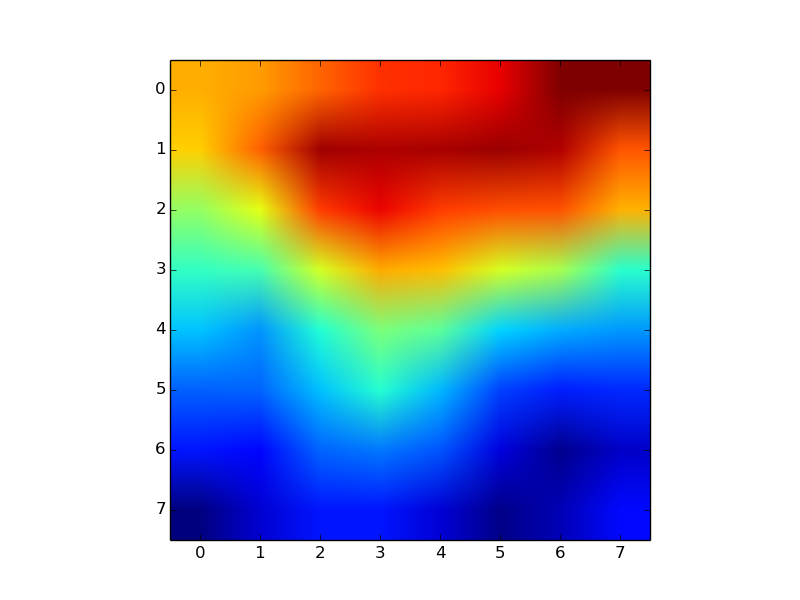}
			\includegraphics[width=0.15\textwidth]{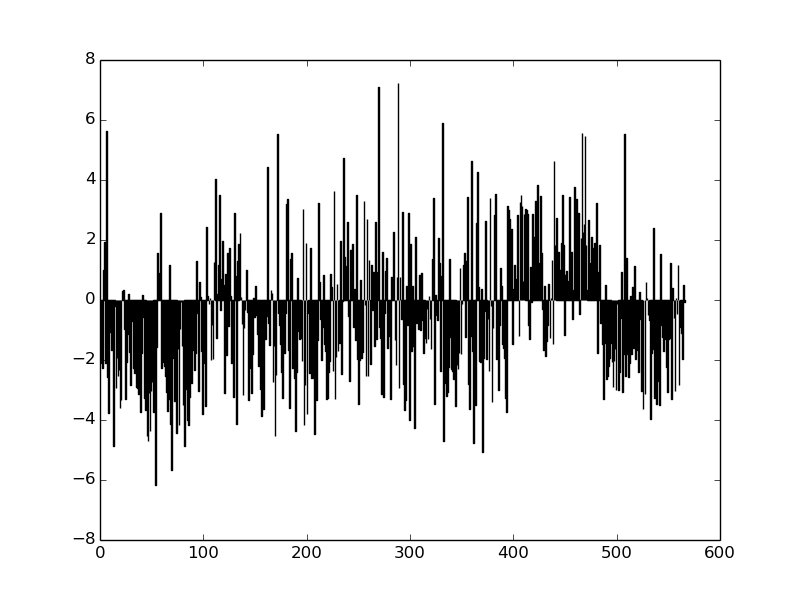}
			\includegraphics[width=0.15\textwidth]{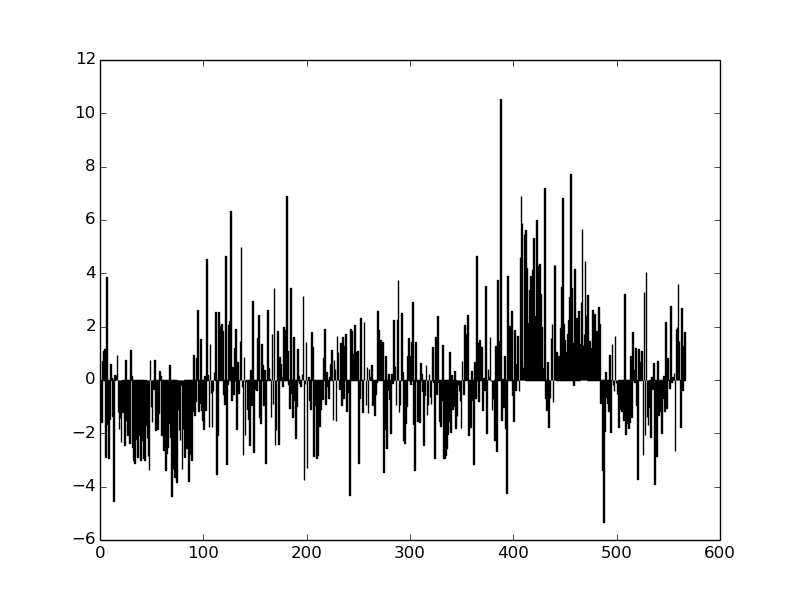}
			\includegraphics[width=0.15\textwidth]{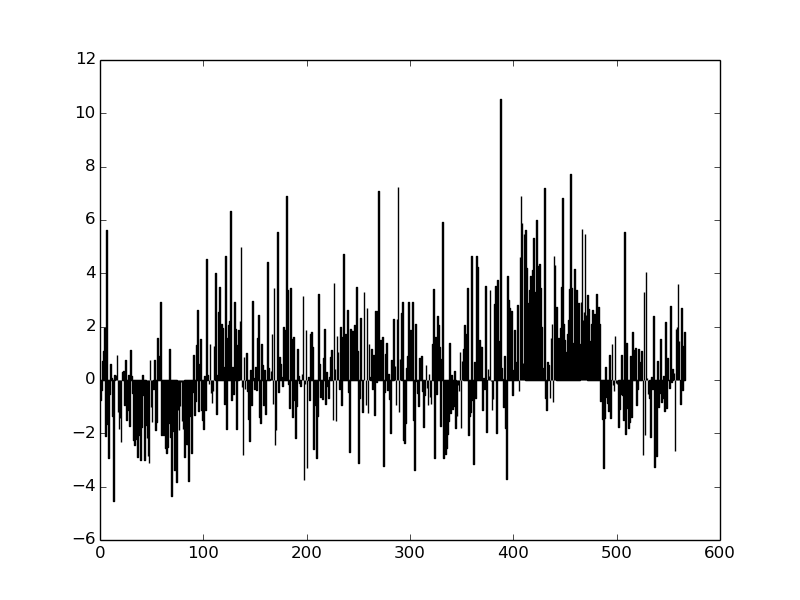}
			\end{tabular} \\			
			\begin{tabular}{cc}
			GT: \textit{A person is slicing an onion}  &
			MM-VDN: \textit{A man is slicing an onion.}
			\end{tabular}  \\~\\
			
			\begin{tabular}{c}
			\includegraphics[width=0.15\textwidth]{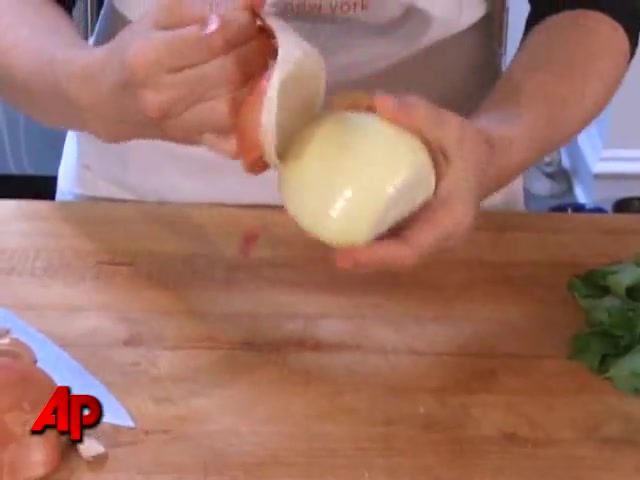}
			\includegraphics[width=0.15\textwidth]{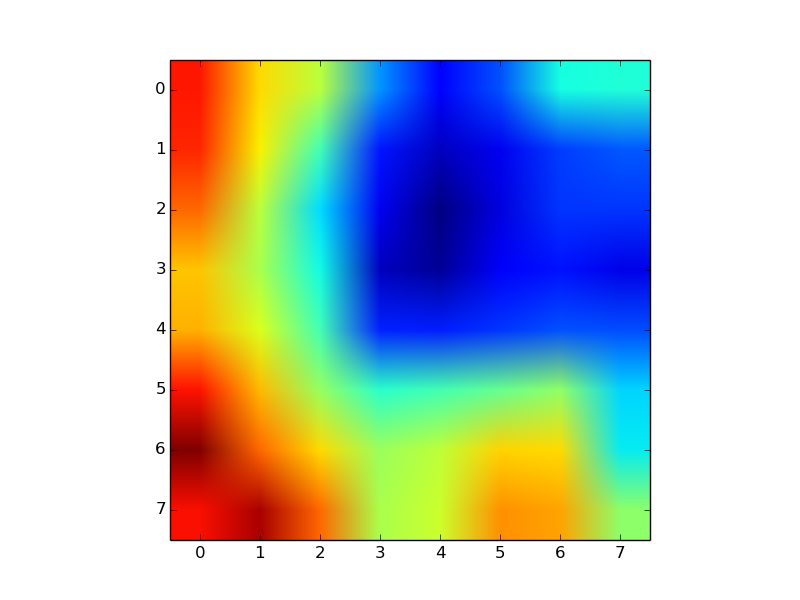}
			\includegraphics[width=0.15\textwidth]{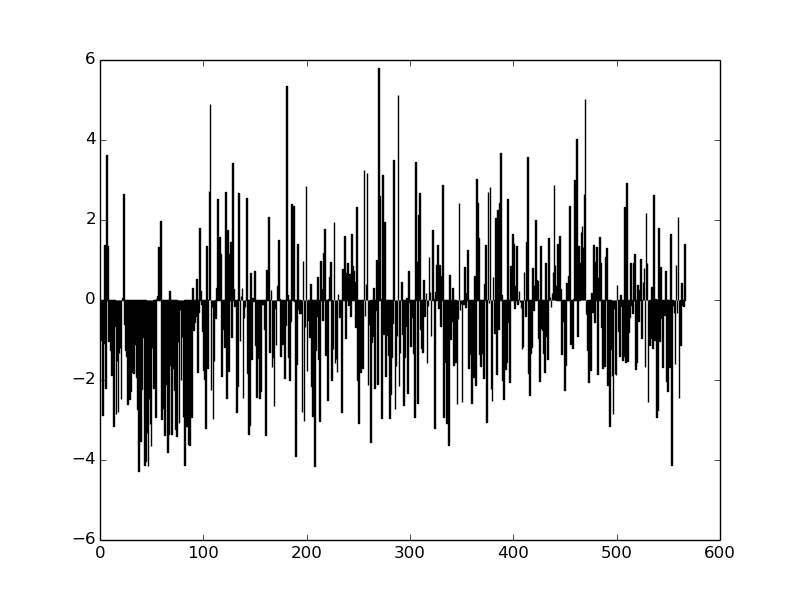}
			\includegraphics[width=0.15\textwidth]{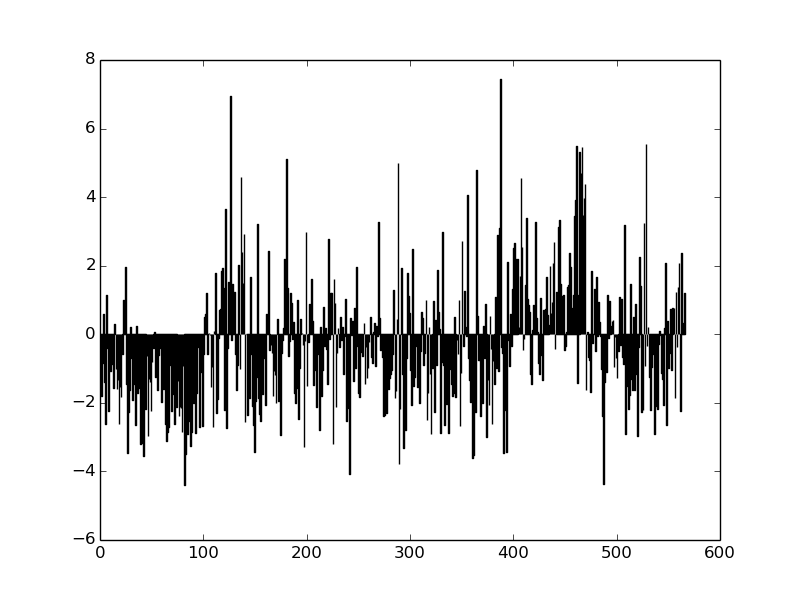}
			\includegraphics[width=0.15\textwidth]{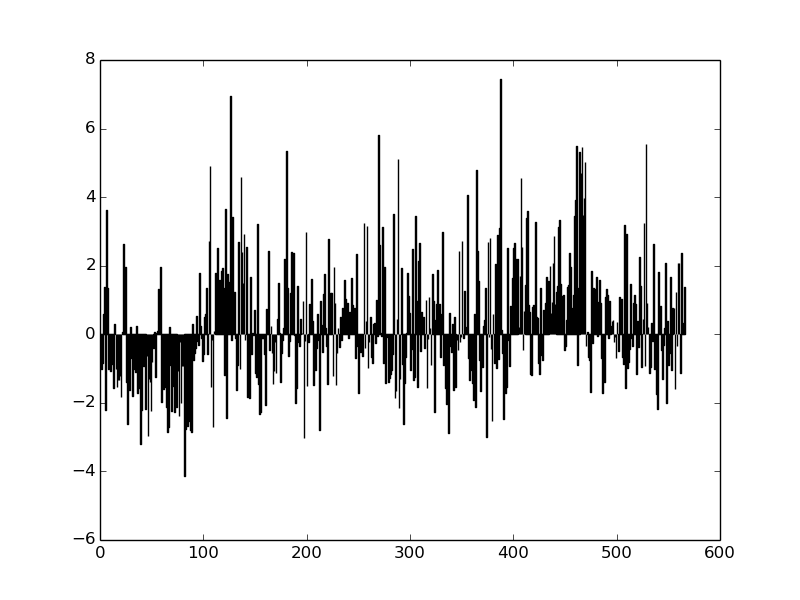}
			\end{tabular} \\			
			\begin{tabular}{cc}
			GT: \textit{A person is peeling an onion.}  &
			MM-VDN: \textit{A man is peeling an onion.}
			\end{tabular}  \\~\\
			
			\begin{tabular}{c}
			\includegraphics[width=0.15\textwidth]{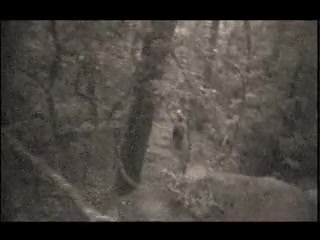}
			\includegraphics[width=0.15\textwidth]{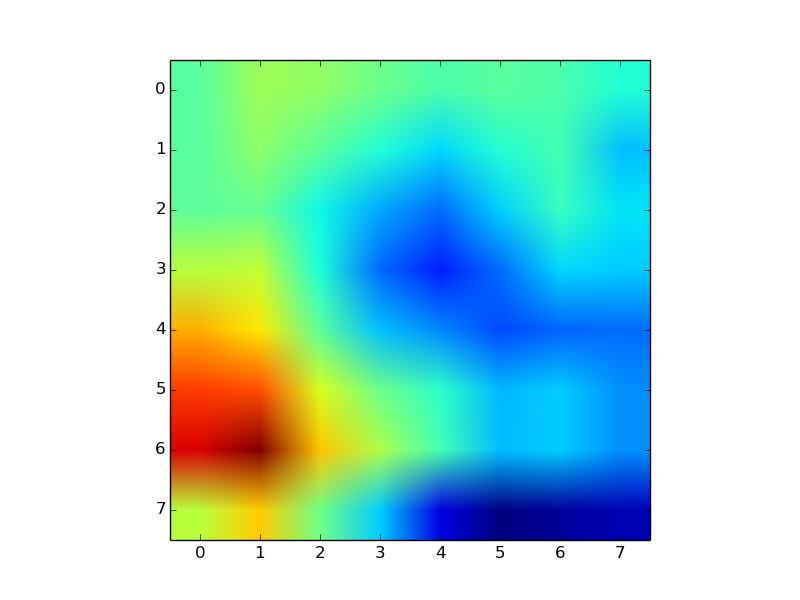}
			\includegraphics[width=0.15\textwidth]{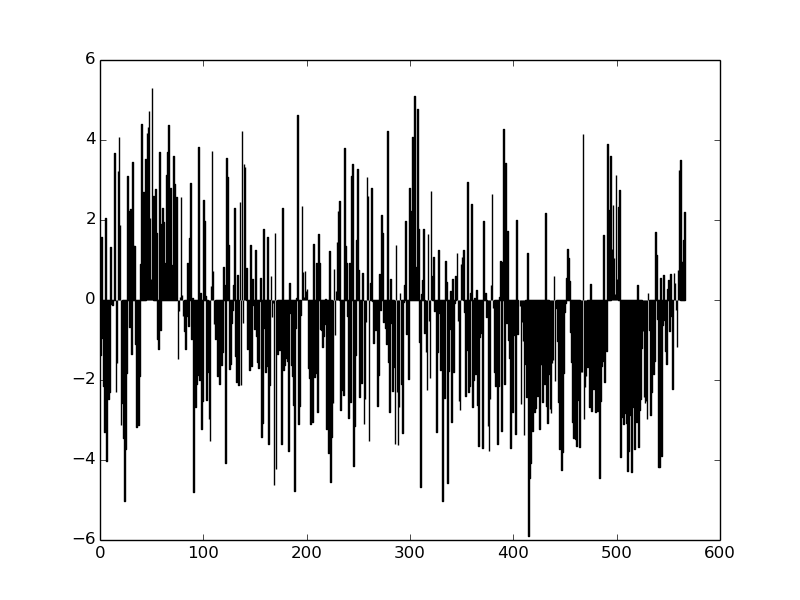}
			\includegraphics[width=0.15\textwidth]{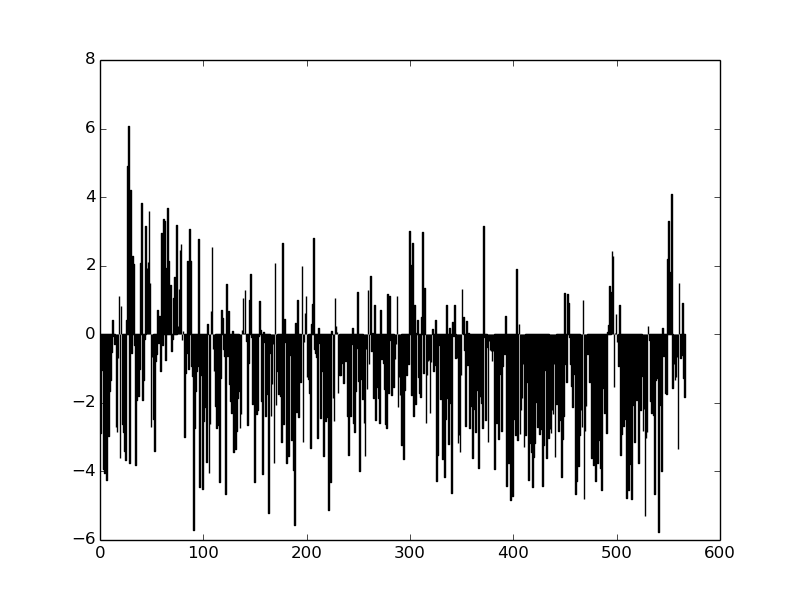}
			\includegraphics[width=0.15\textwidth]{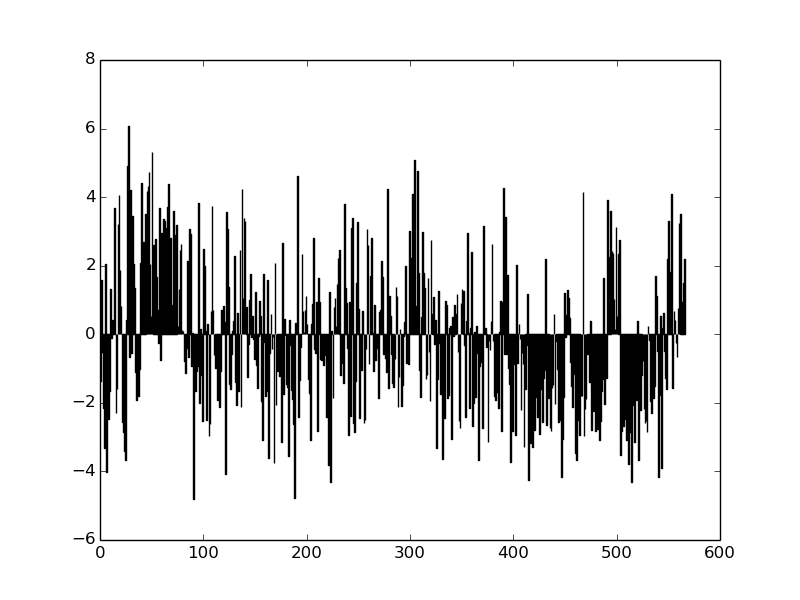}
			\end{tabular} \\			
			\begin{tabular}{cc}
			GT: \textit{A man is riding a bicycle.}  &
			MM-VDN: \textit{A man is riding a bicycle.}
			\end{tabular}
		\end{tabular}
	\end{center}
\end{table*}

\begin{table*}[t]
	\begin{center}
		\begin{tabular}{c}
			\begin{tabular}{c}
			\includegraphics[width=0.15\textwidth]{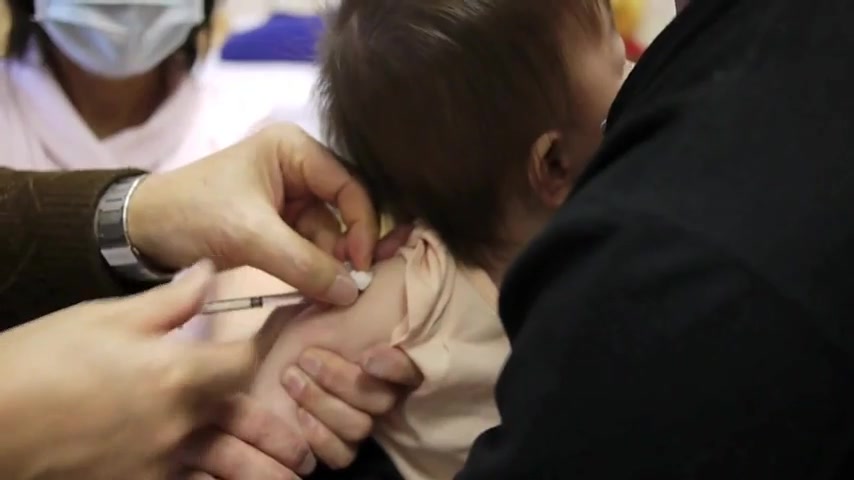}
			\includegraphics[width=0.15\textwidth]{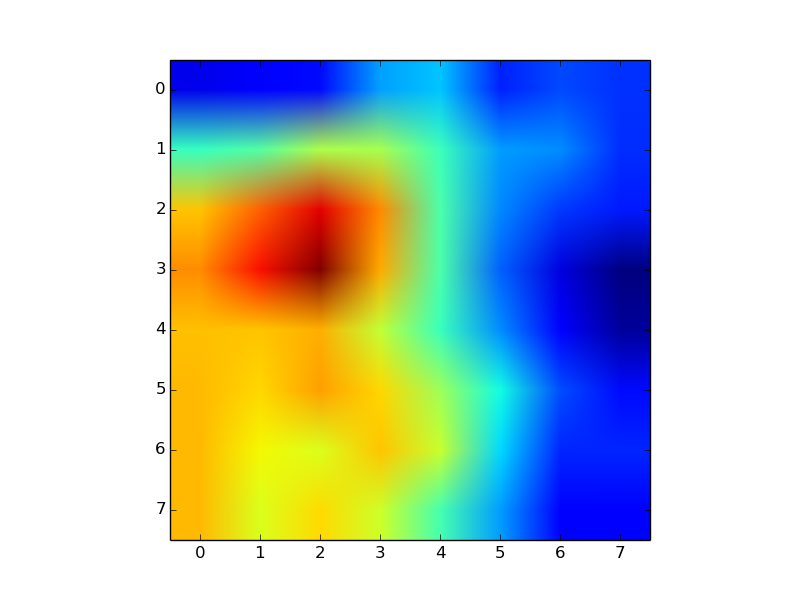}
			\includegraphics[width=0.15\textwidth]{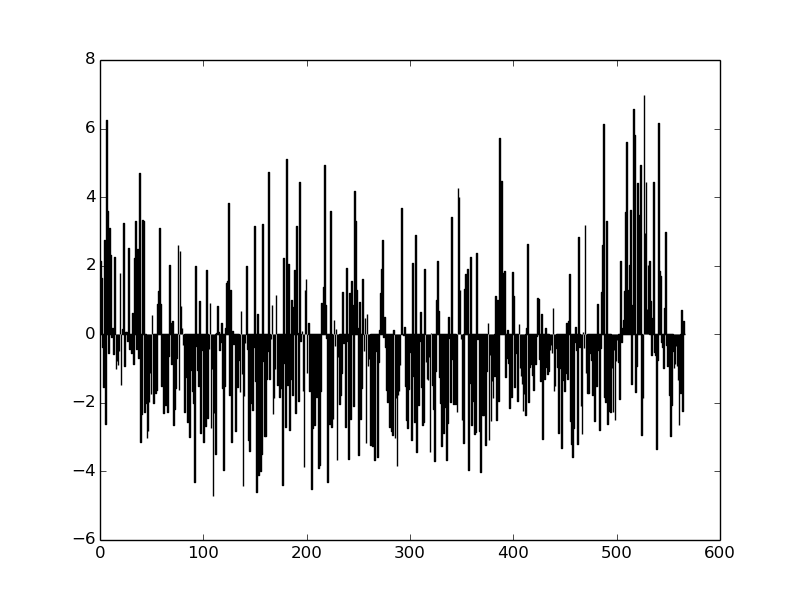}
			\includegraphics[width=0.15\textwidth]{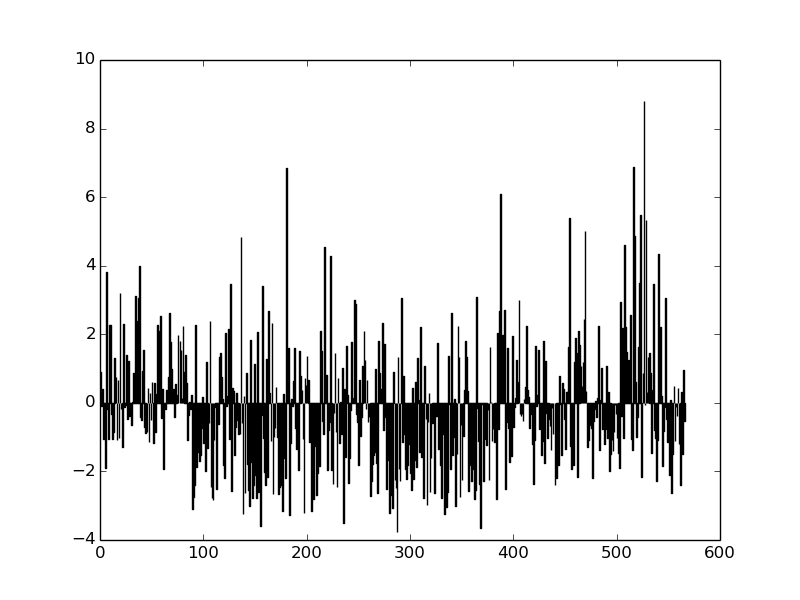}
			\includegraphics[width=0.15\textwidth]{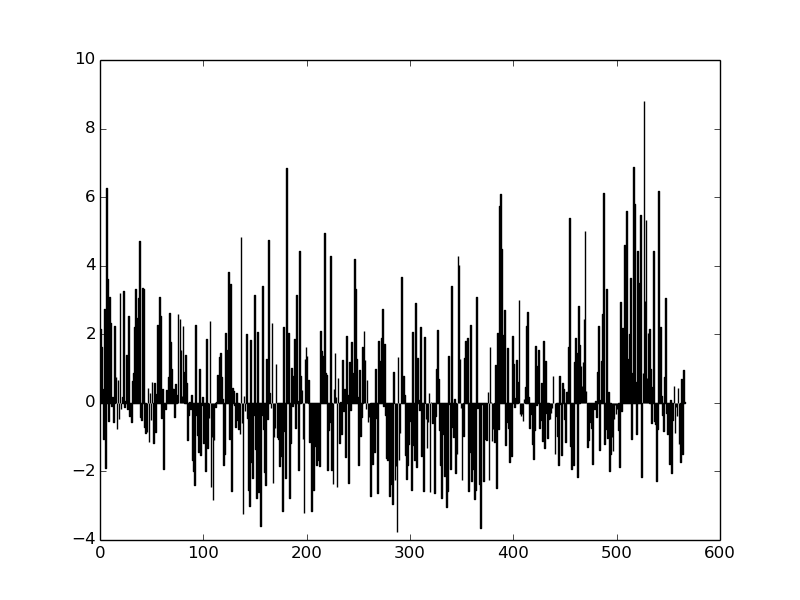}
			\end{tabular} \\			
			\begin{tabular}{cc}
			GT: \textit{A doctor gives a shot to a crying baby}  &
			MM-VDN: \textit{A man is playing with a baby.}
			\end{tabular}  \\~\\
			
			\begin{tabular}{c}
			\includegraphics[width=0.15\textwidth]{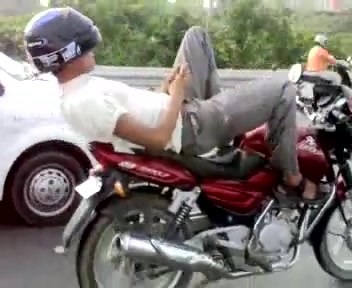}
			\includegraphics[width=0.15\textwidth]{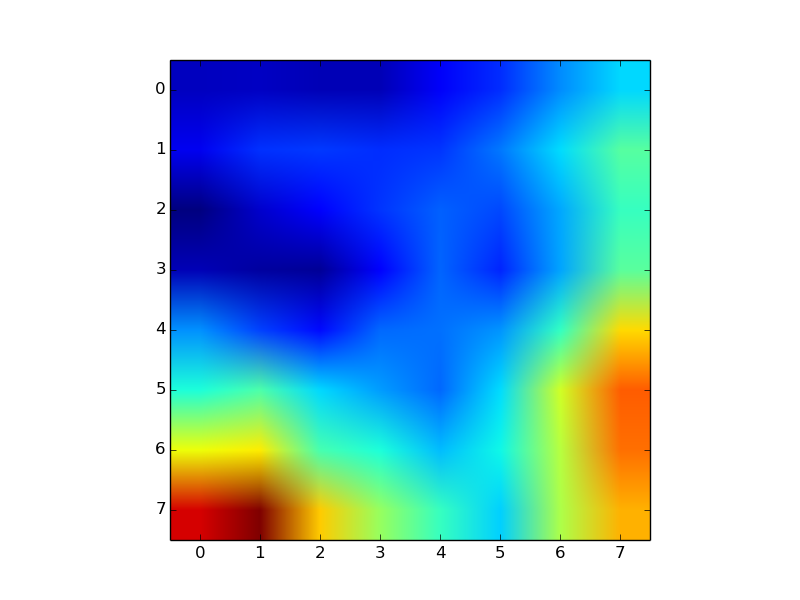}
			\includegraphics[width=0.15\textwidth]{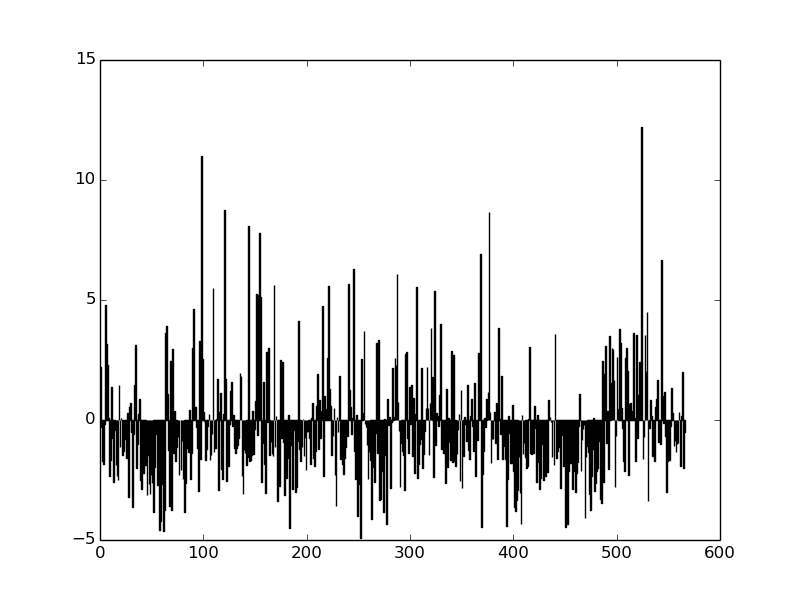}
			\includegraphics[width=0.15\textwidth]{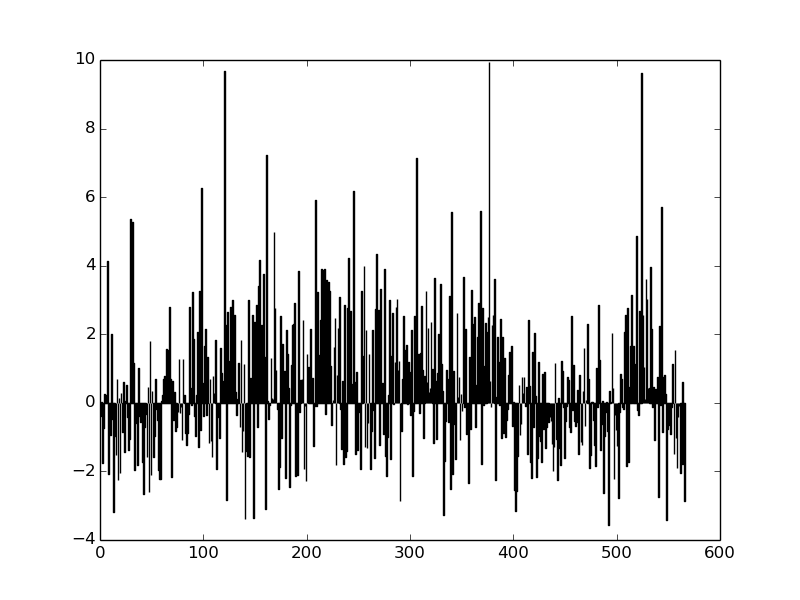}
			\includegraphics[width=0.15\textwidth]{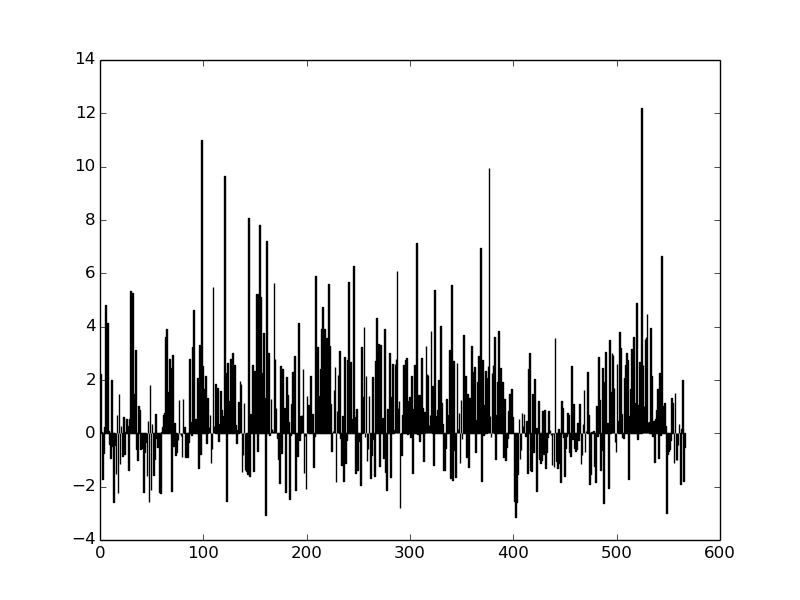}
			\end{tabular} \\			
			\begin{tabular}{cc}
			GT: \textit{A man is riding a motorcycle.}  &
			MM-VDN: \textit{A man is riding a motorcycle.}
			\end{tabular}  \\~\\
			
			\begin{tabular}{c}
			\includegraphics[width=0.15\textwidth]{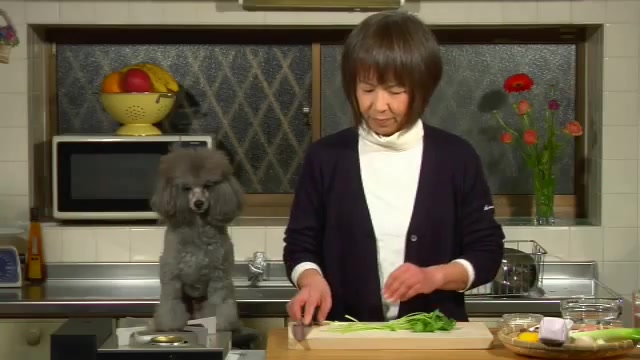}
			\includegraphics[width=0.15\textwidth]{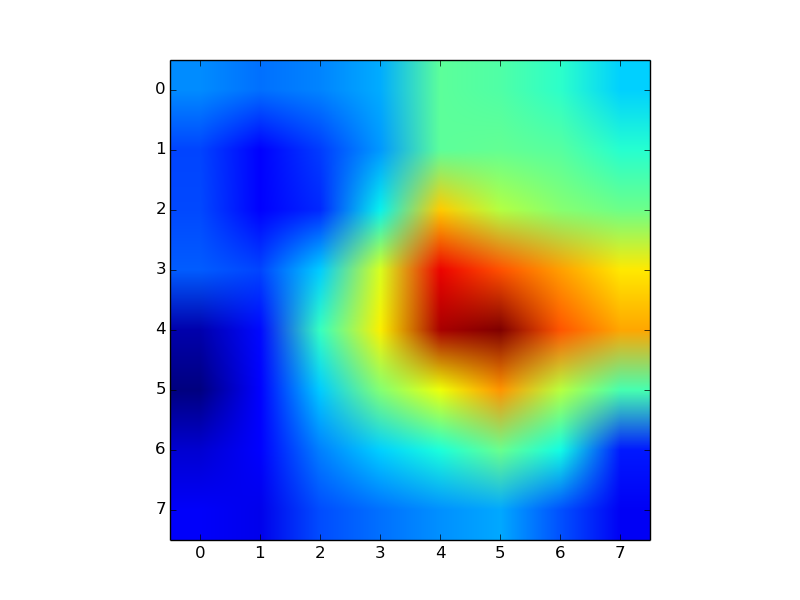}
			\includegraphics[width=0.15\textwidth]{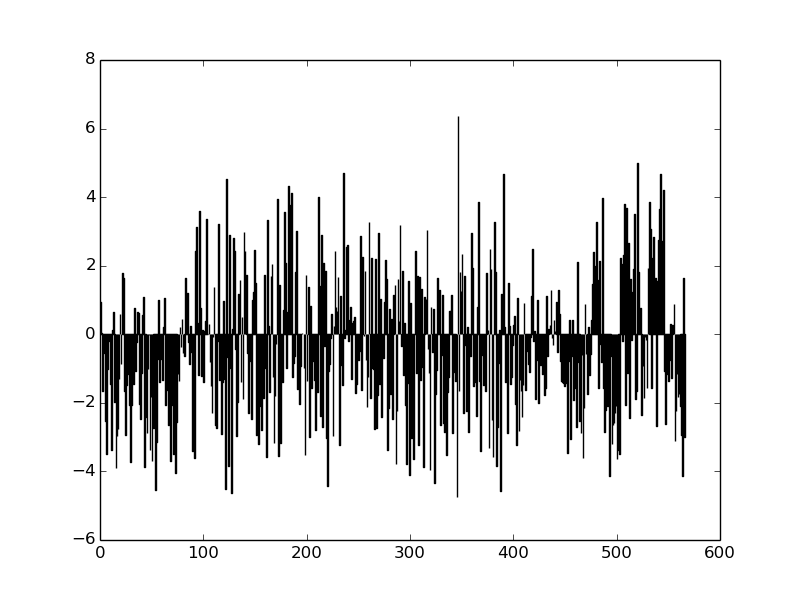}
			\includegraphics[width=0.15\textwidth]{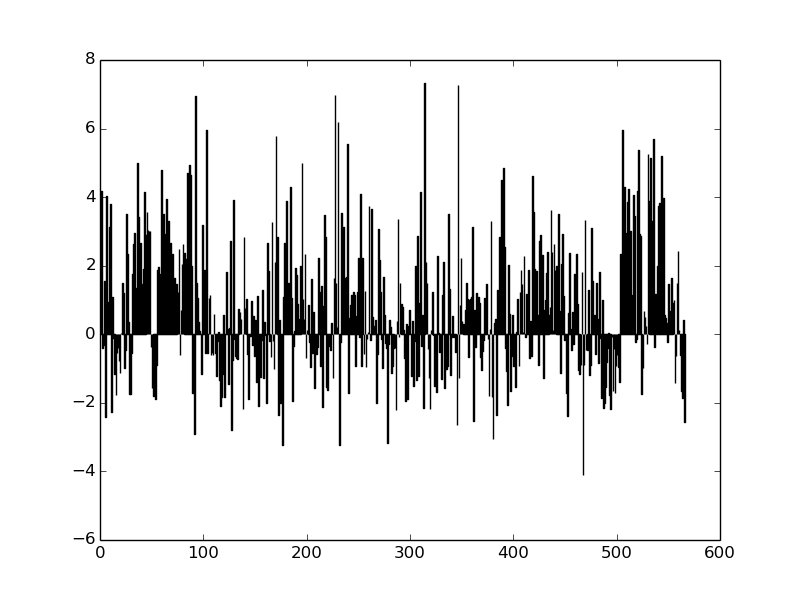}
			\includegraphics[width=0.15\textwidth]{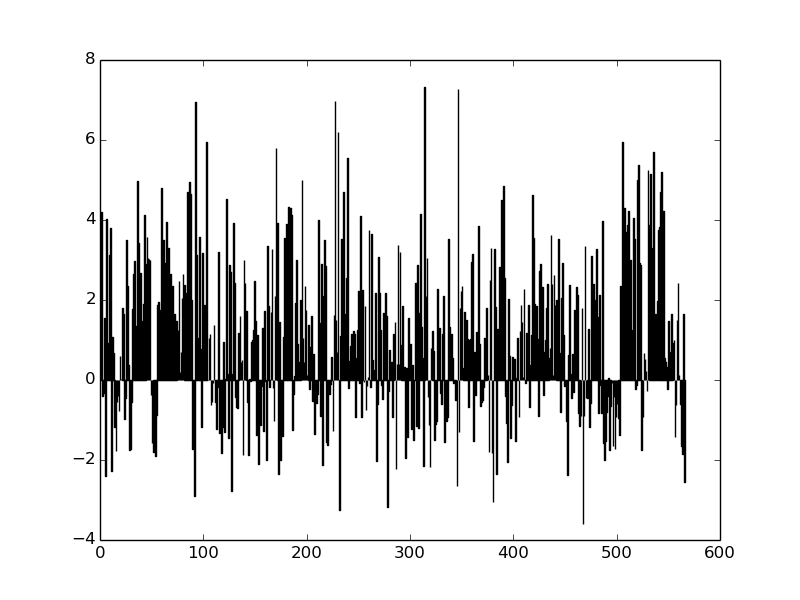}
			\end{tabular} \\			
			\begin{tabular}{cc}
			GT: \textit{A woman is cutting green onion.}  &
			MM-VDN: \textit{A woman is cutting a green onion.}
			\end{tabular}  \\~\\
			
			\begin{tabular}{c}
			\includegraphics[width=0.15\textwidth]{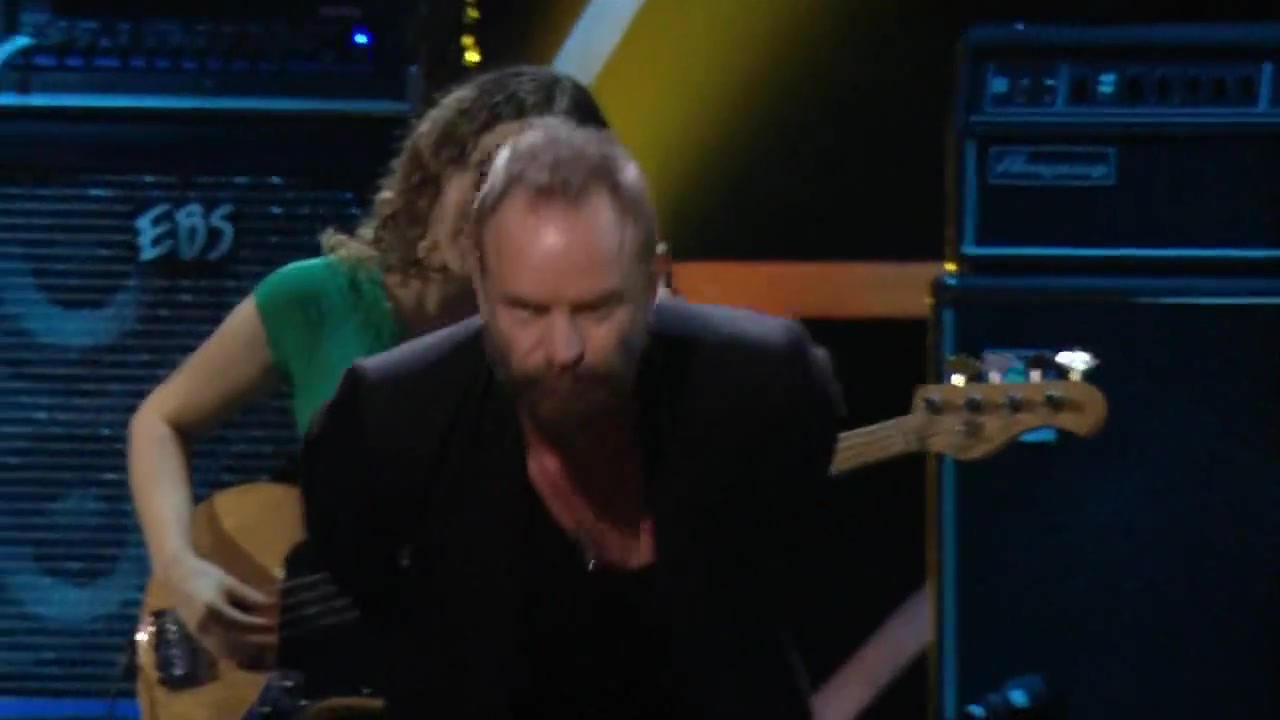}
			\includegraphics[width=0.15\textwidth]{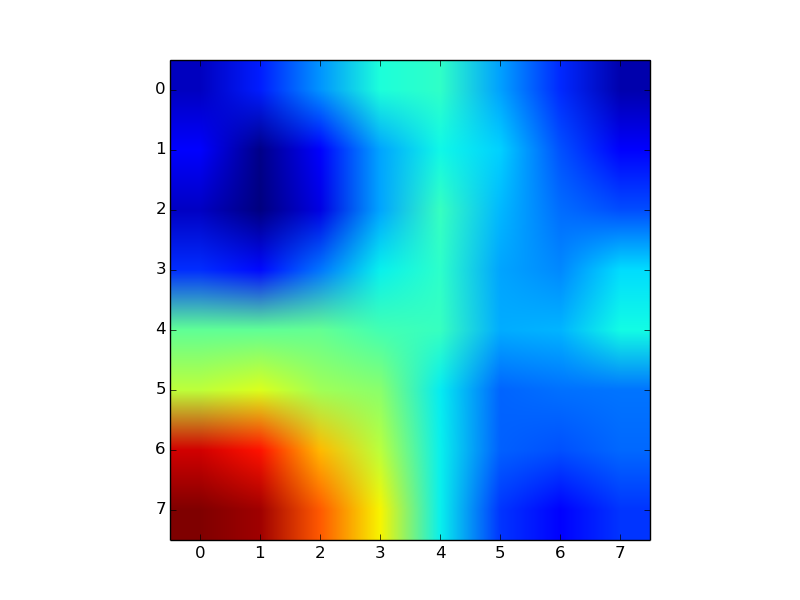}
			\includegraphics[width=0.15\textwidth]{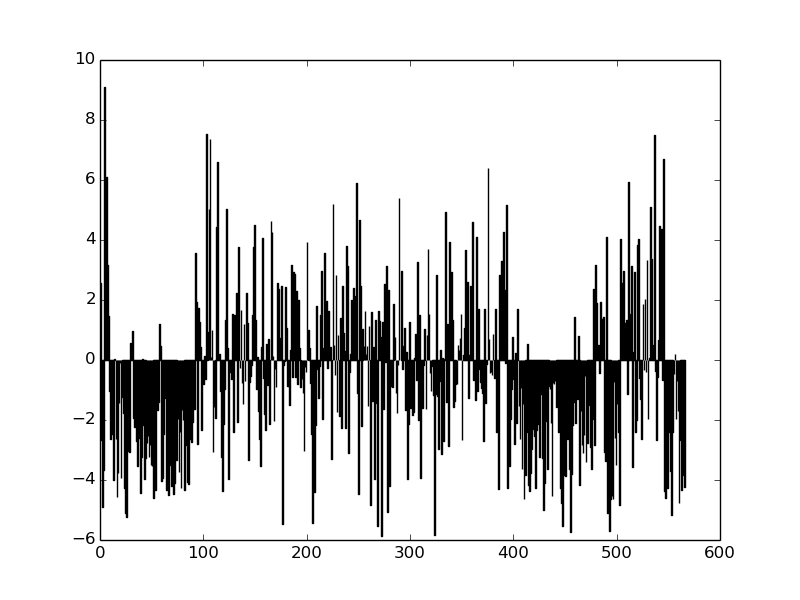}
			\includegraphics[width=0.15\textwidth]{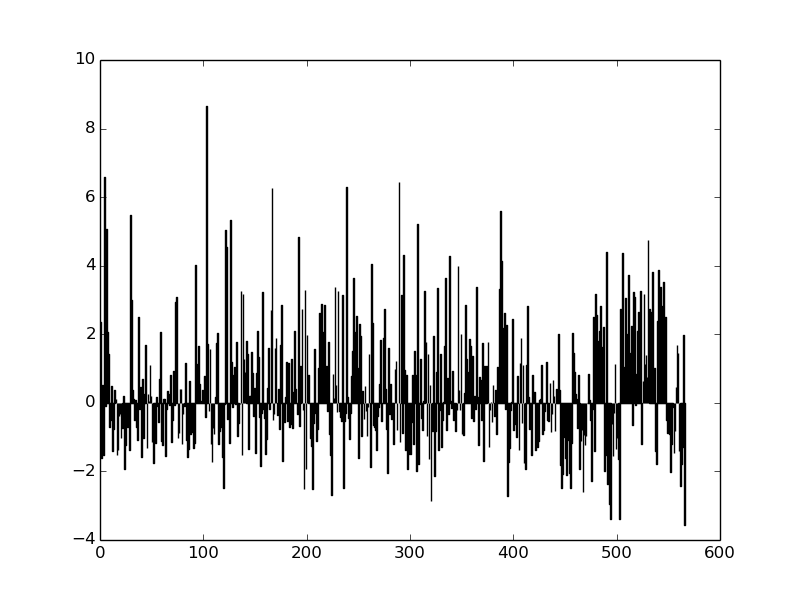}
			\includegraphics[width=0.15\textwidth]{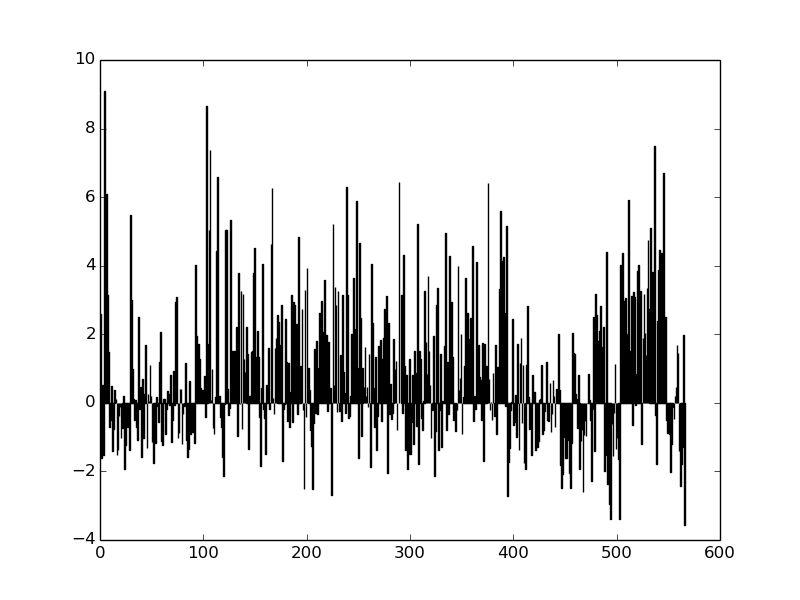}
			\end{tabular} \\			
			\begin{tabular}{cc}
			GT: \textit{A band is performing on the stage.}  &
			MM-VDN: \textit{A band is performing on stage.}
			\end{tabular}  \\~\\
			
			\begin{tabular}{c}
			\includegraphics[width=0.15\textwidth]{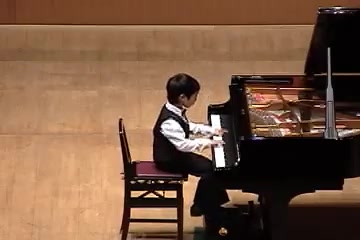}
			\includegraphics[width=0.15\textwidth]{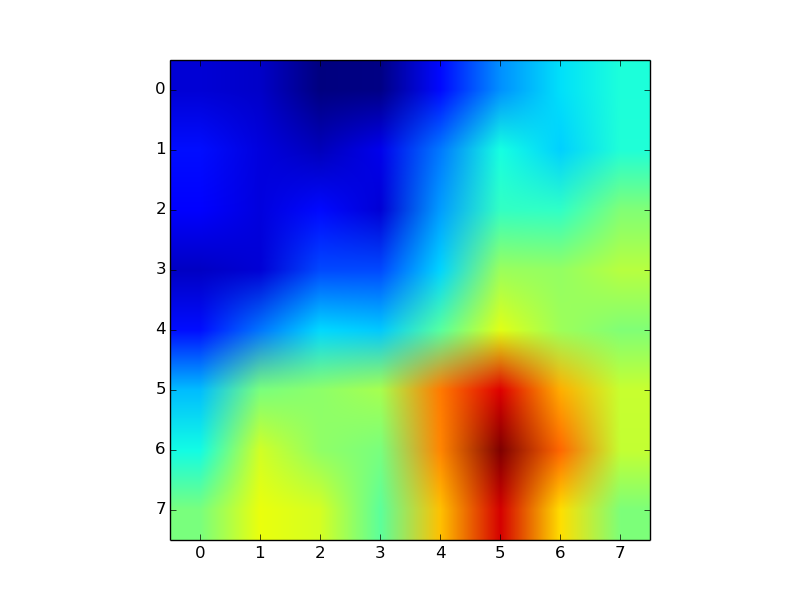}
			\includegraphics[width=0.15\textwidth]{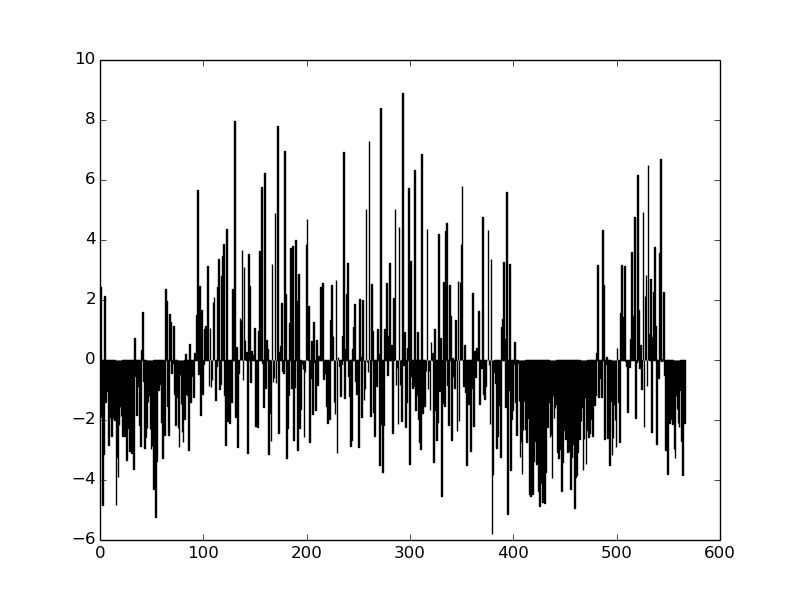}
			\includegraphics[width=0.15\textwidth]{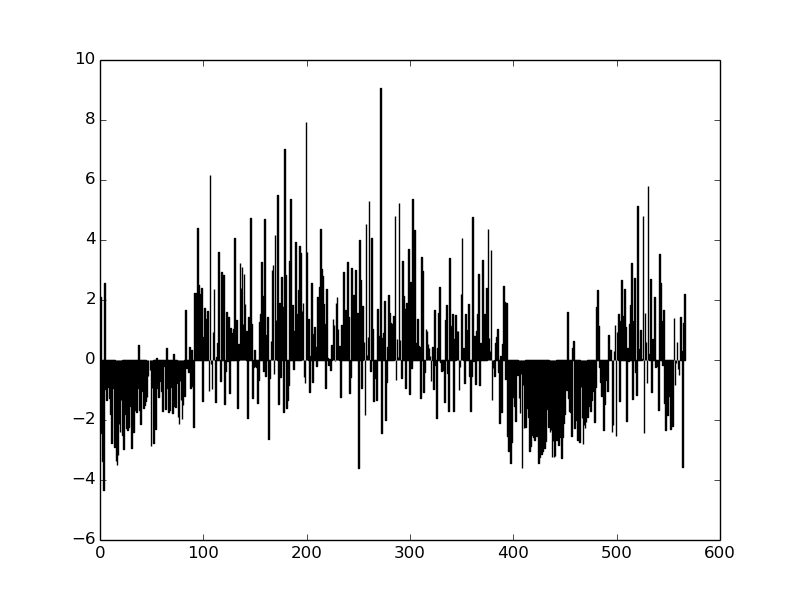}
			\includegraphics[width=0.15\textwidth]{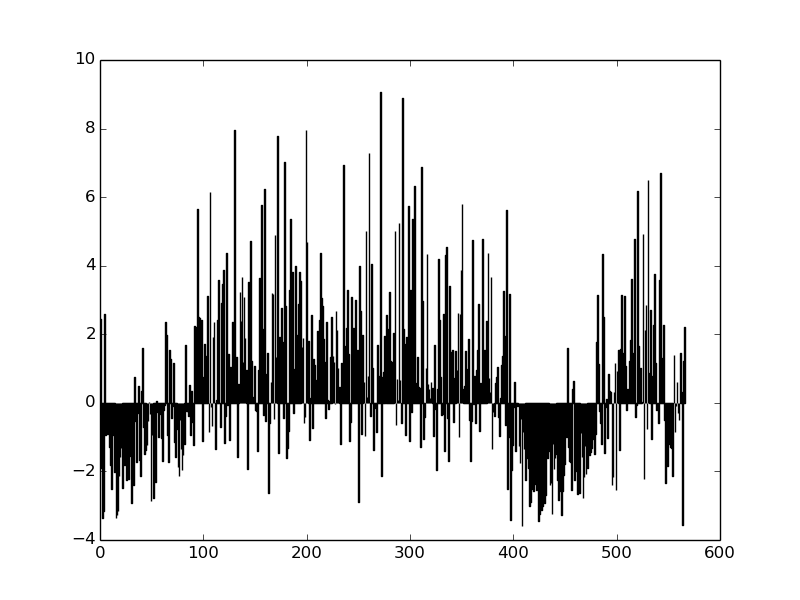}
			\end{tabular} \\			
			\begin{tabular}{cc}
			GT: \textit{A boy is playing a piano}  &
			MM-VDN: \textit{A man is playing a piano.}
			\end{tabular}  \\~\\
			
			\begin{tabular}{c}
			\includegraphics[width=0.15\textwidth]{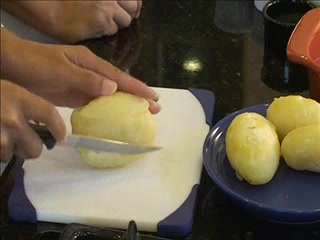}
			\includegraphics[width=0.15\textwidth]{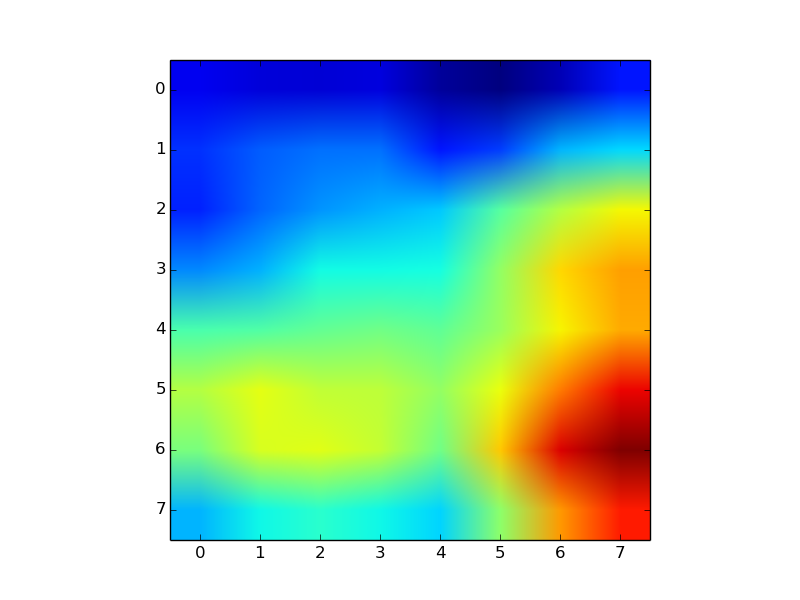}
			\includegraphics[width=0.15\textwidth]{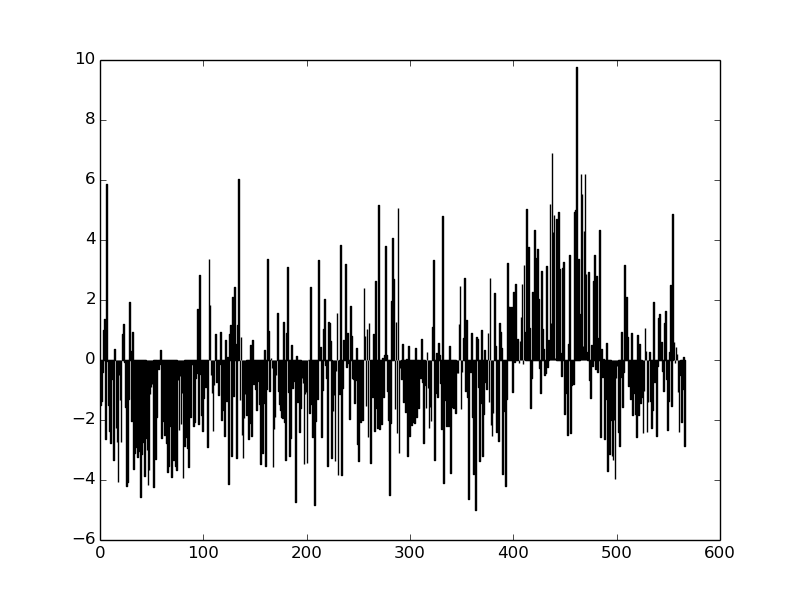}
			\includegraphics[width=0.15\textwidth]{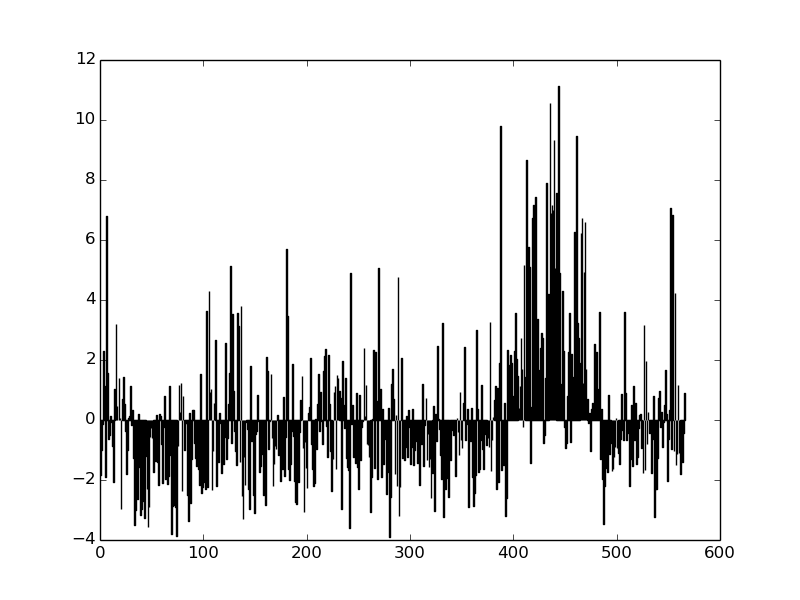}
			\includegraphics[width=0.15\textwidth]{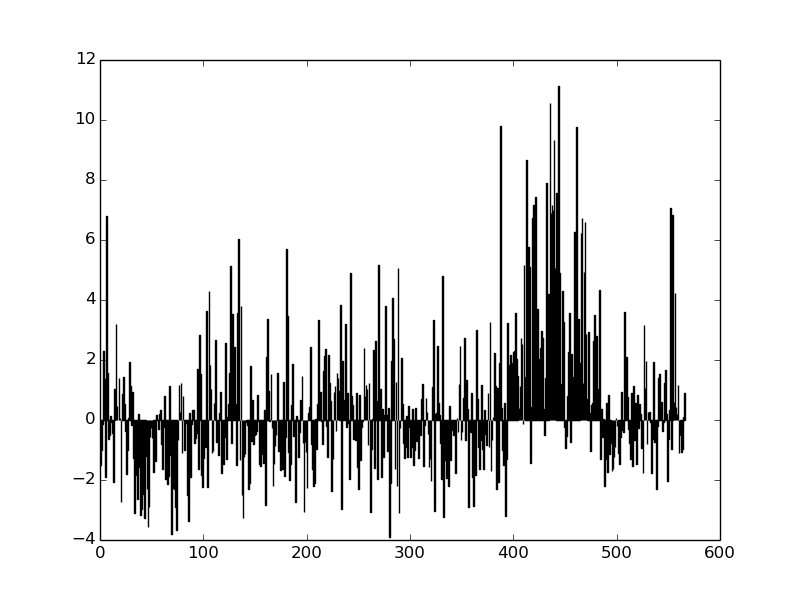}
			\end{tabular} \\			
			\begin{tabular}{cc}
			GT: \textit{A girl is slicing a potato into pieces.}  &
			MM-VDN: \textit{A man is peeling a potato.}
			\end{tabular}  \\~\\
			
			\begin{tabular}{c}
			\includegraphics[width=0.15\textwidth]{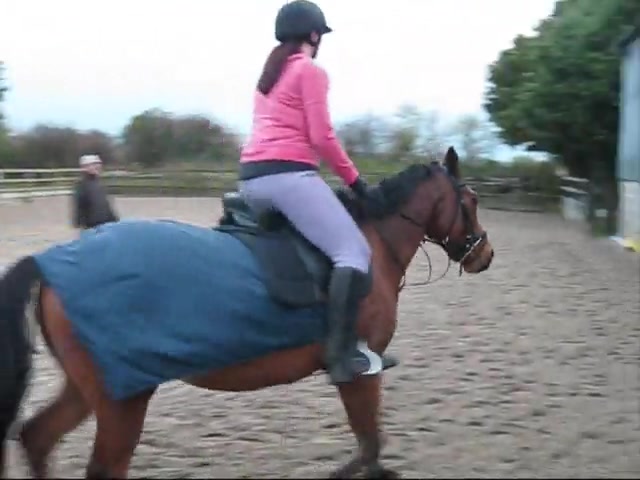}
			\includegraphics[width=0.15\textwidth]{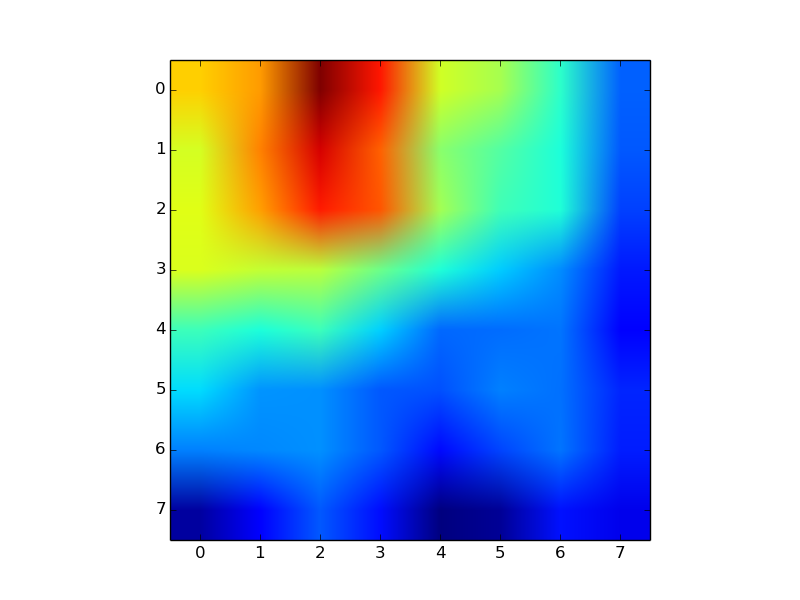}
			\includegraphics[width=0.15\textwidth]{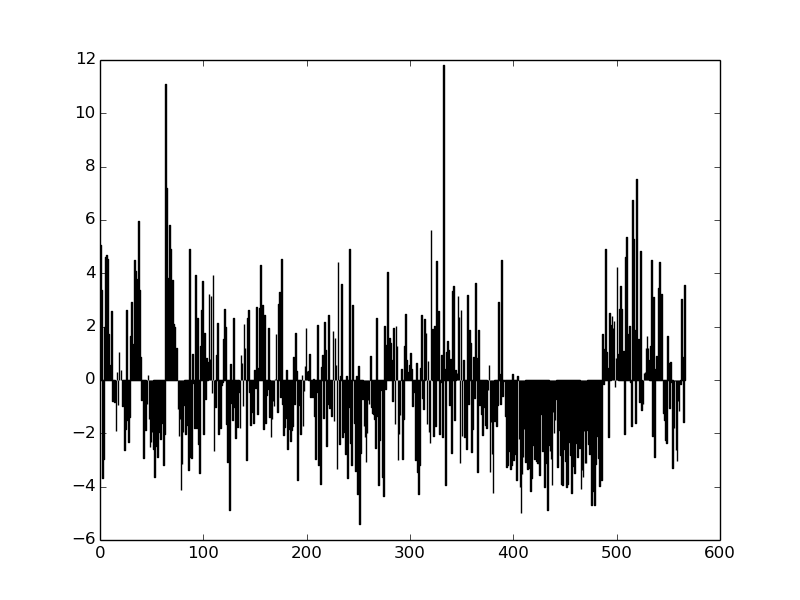}
			\includegraphics[width=0.15\textwidth]{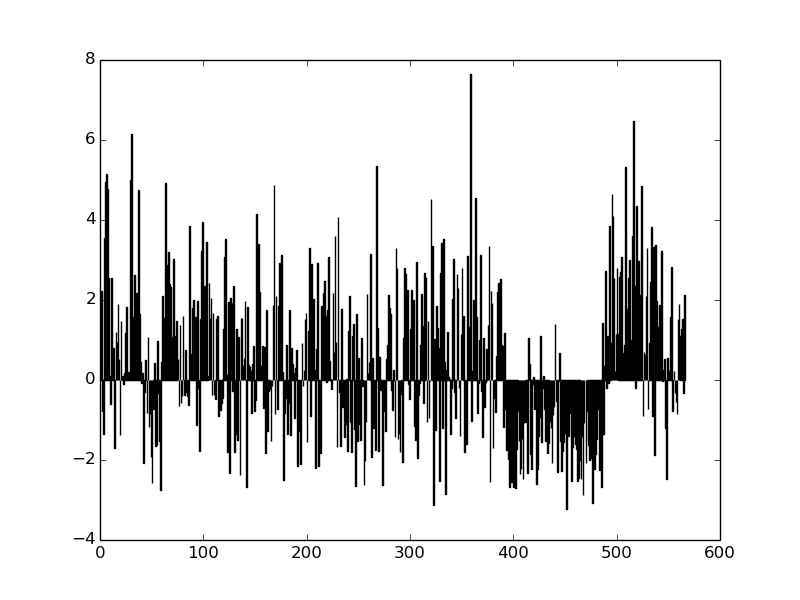}
			\includegraphics[width=0.15\textwidth]{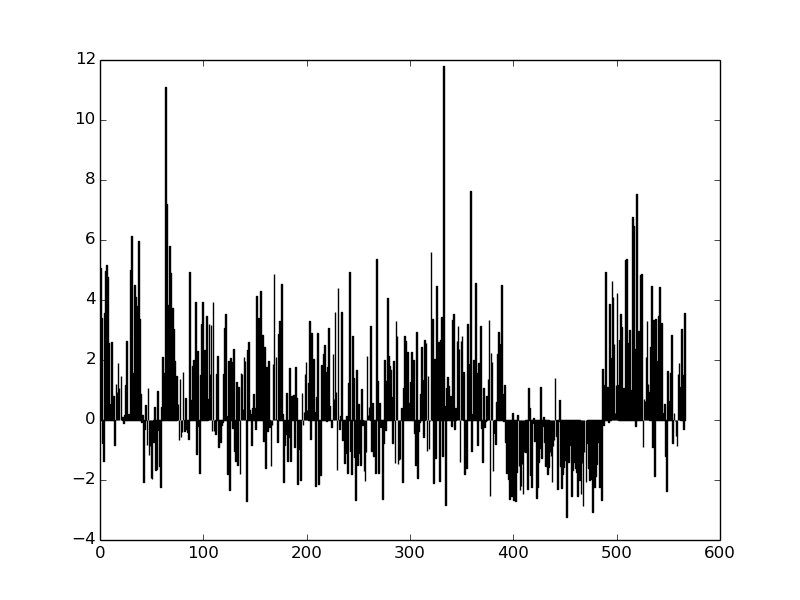}
			\end{tabular} \\			
			\begin{tabular}{cc}
			GT: \textit{A girl is riding a horse.}  &
			MM-VDN: \textit{A girl is riding a horse.}
			\end{tabular}
		\end{tabular}
	\end{center}
\end{table*}

\begin{table*}[t]
	\begin{center}
		\begin{tabular}{c}
			\begin{tabular}{c}
			\includegraphics[width=0.15\textwidth]{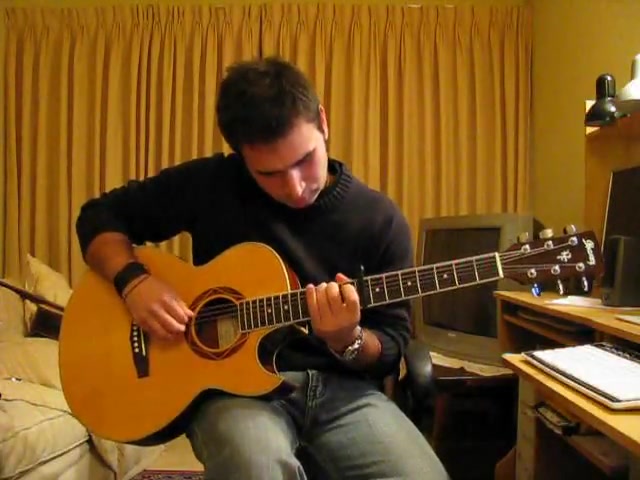}
			\includegraphics[width=0.15\textwidth]{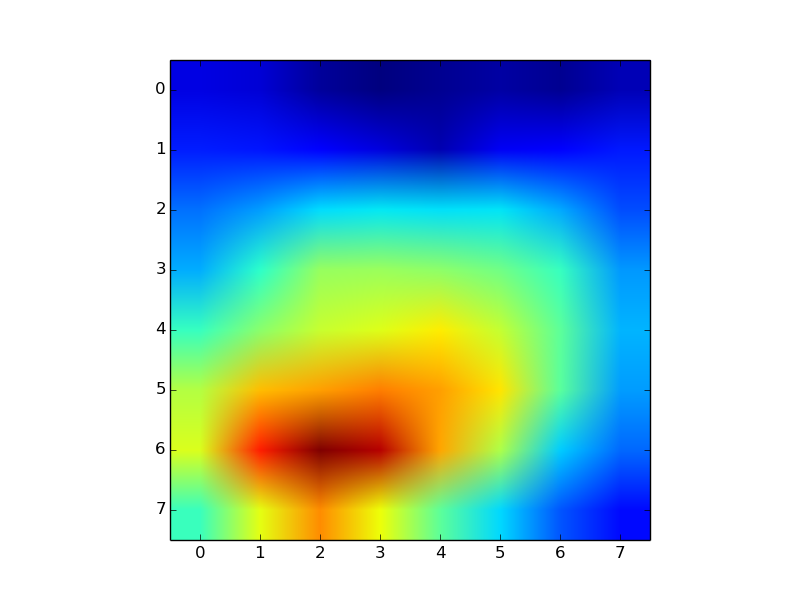}
			\includegraphics[width=0.15\textwidth]{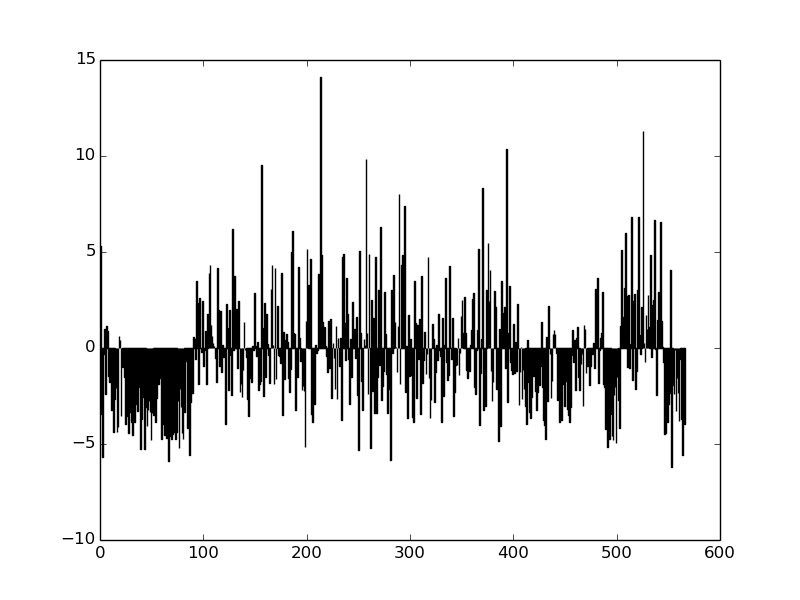}
			\includegraphics[width=0.15\textwidth]{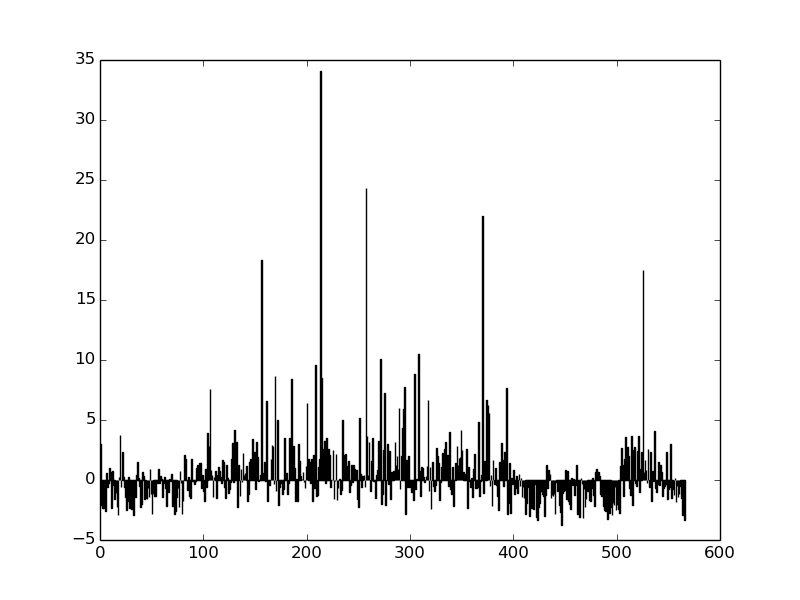}
			\includegraphics[width=0.15\textwidth]{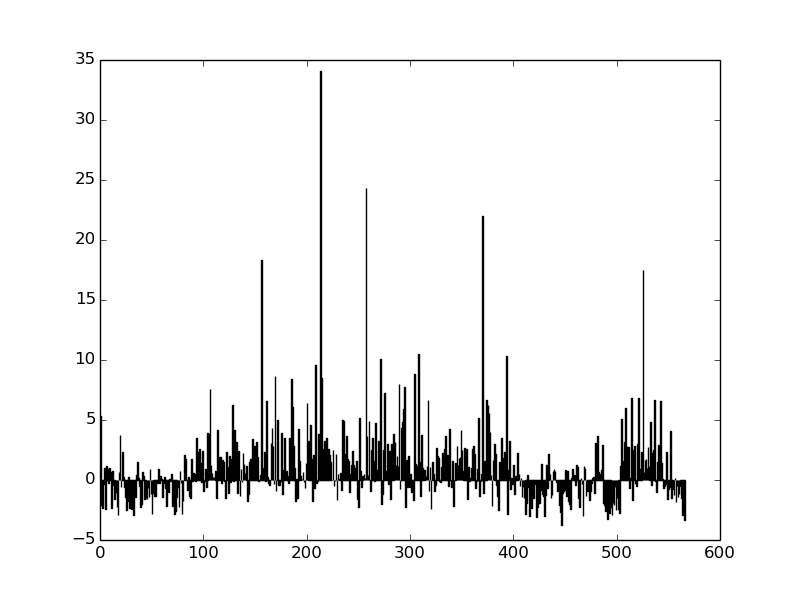}
			\end{tabular} \\			
			\begin{tabular}{cc}
			GT: \textit{A man is playing a guitar.}  &
			MM-VDN: \textit{A man is playing a guitar.}
			\end{tabular}  \\~\\
			
			\begin{tabular}{c}
			\includegraphics[width=0.15\textwidth]{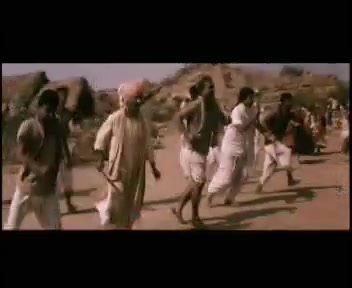}
			\includegraphics[width=0.15\textwidth]{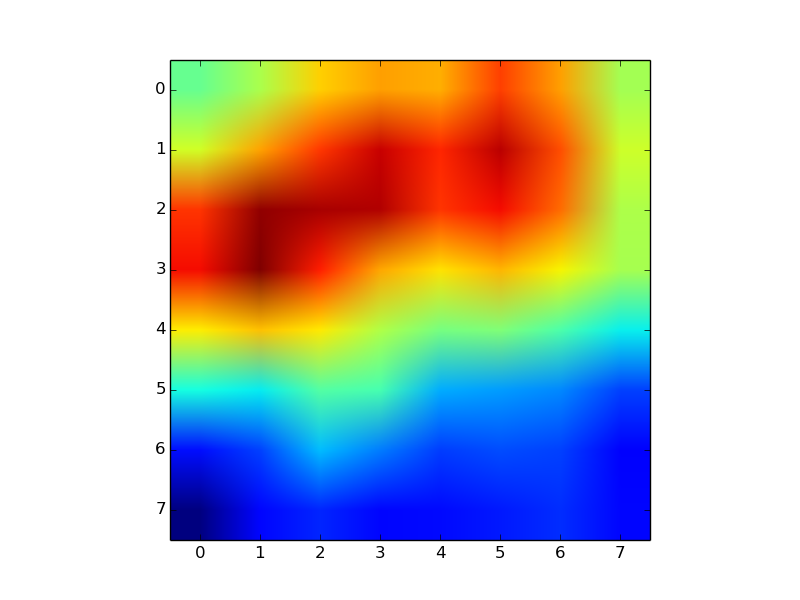}
			\includegraphics[width=0.15\textwidth]{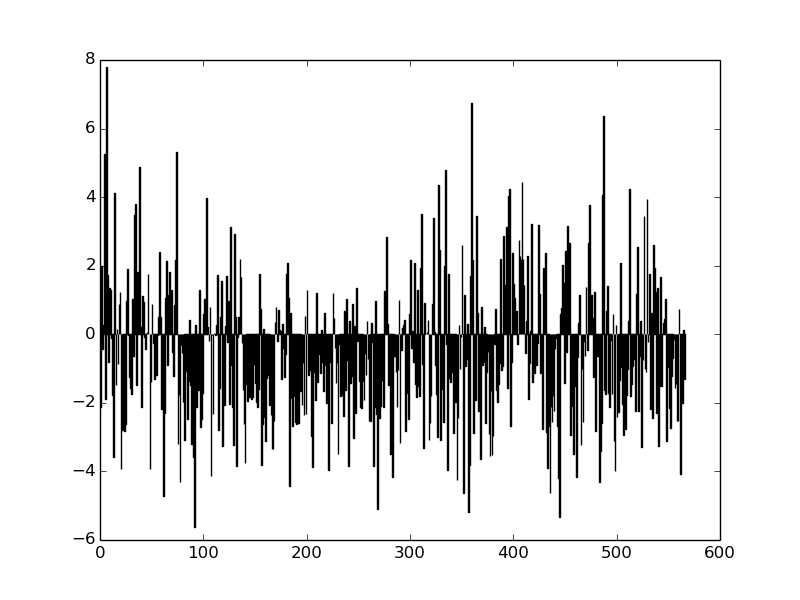}
			\includegraphics[width=0.15\textwidth]{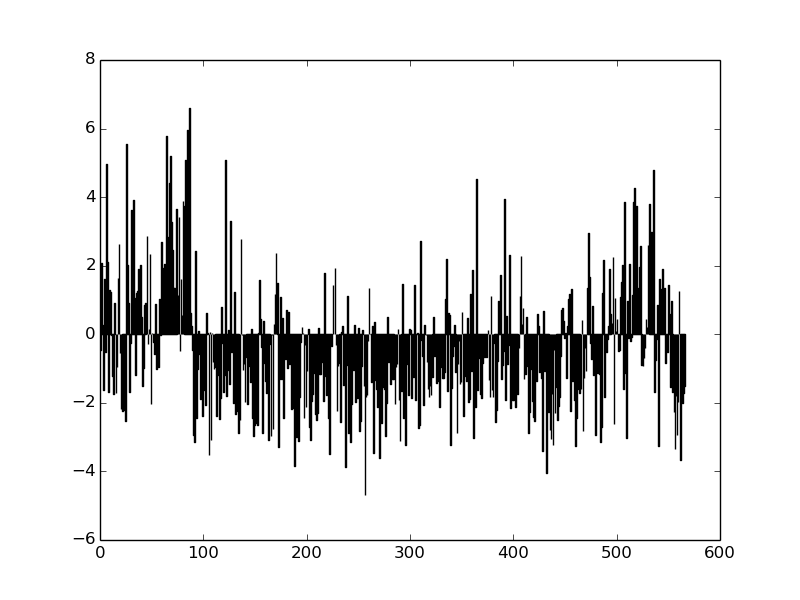}
			\includegraphics[width=0.15\textwidth]{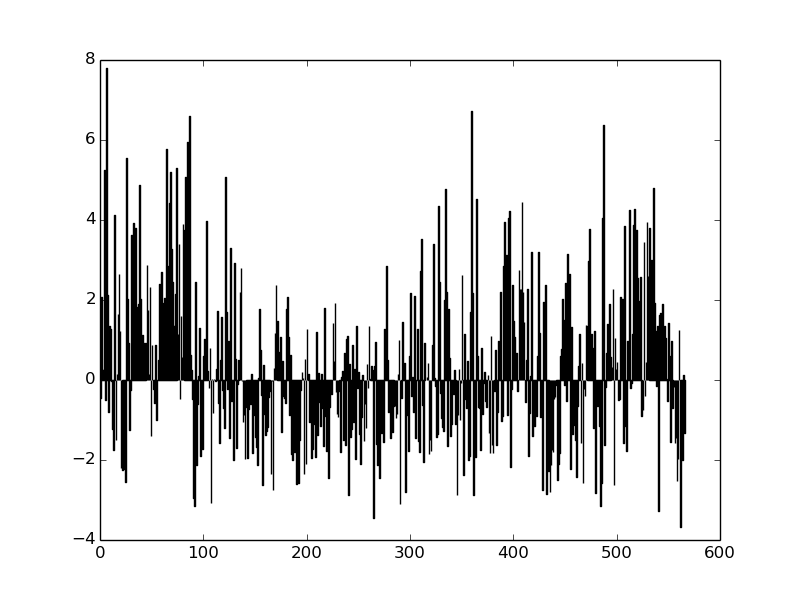}
			\end{tabular} \\			
			\begin{tabular}{cc}
			GT: \textit{People are dancing outside.}  &
			MM-VDN: \textit{People are dancing.}
			\end{tabular}  \\~\\
			
			\begin{tabular}{c}
			\includegraphics[width=0.15\textwidth]{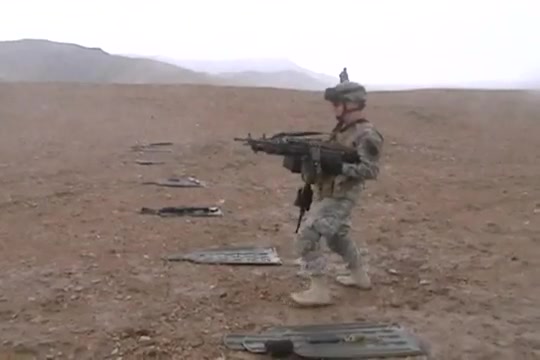}
			\includegraphics[width=0.15\textwidth]{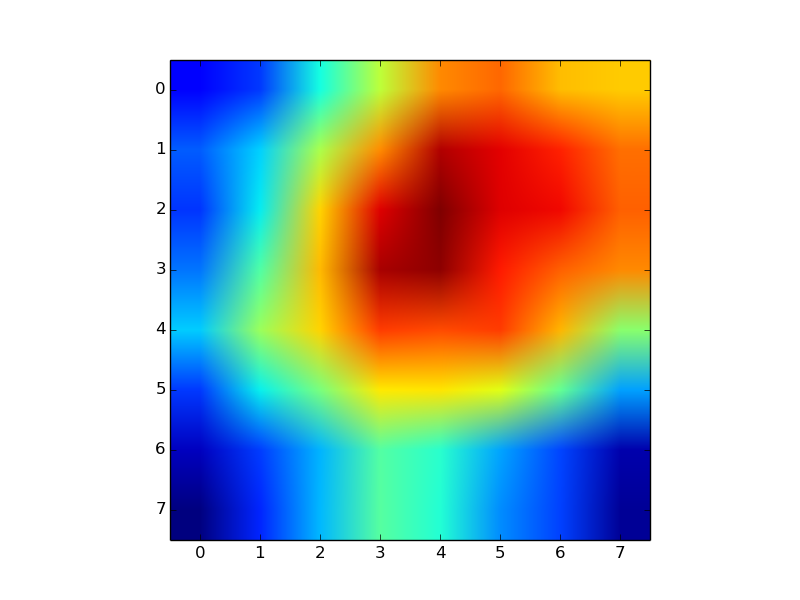}
			\includegraphics[width=0.15\textwidth]{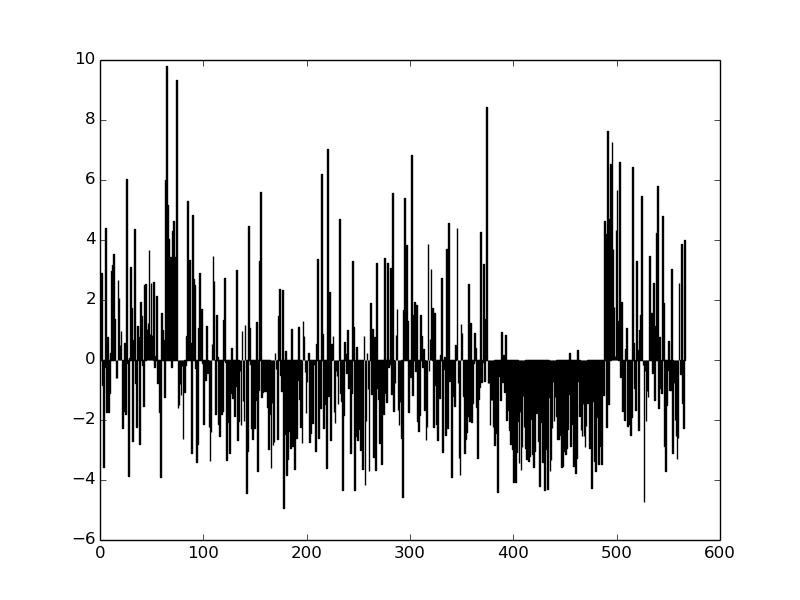}
			\includegraphics[width=0.15\textwidth]{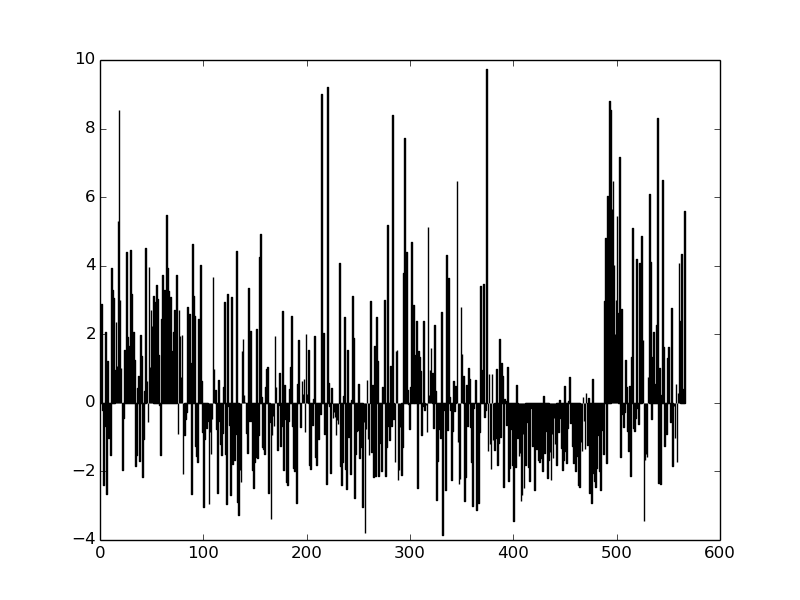}
			\includegraphics[width=0.15\textwidth]{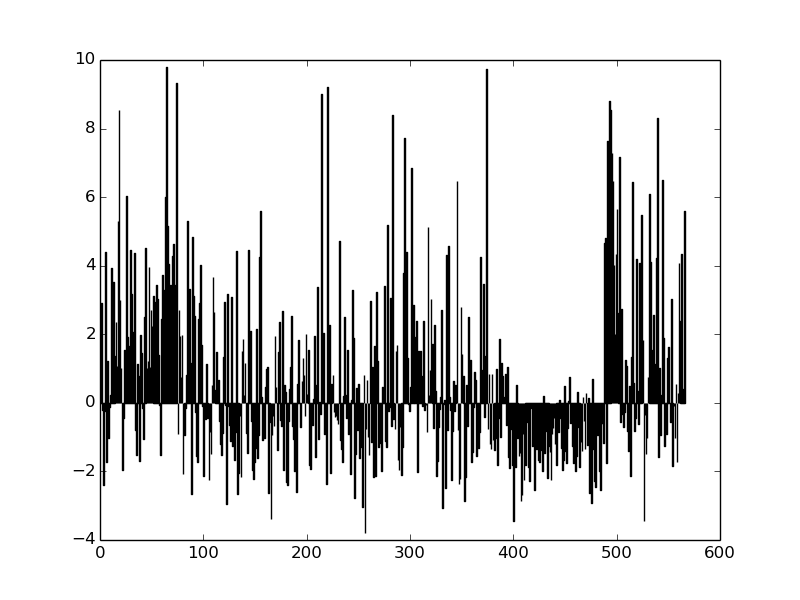}
			\end{tabular} \\			
			\begin{tabular}{cc}
			GT: \textit{A soldier is shooting some target.}  &
			MM-VDN: \textit{A man is shooting a target.}
			\end{tabular}  \\~\\
			
			\begin{tabular}{c}
			\includegraphics[width=0.15\textwidth]{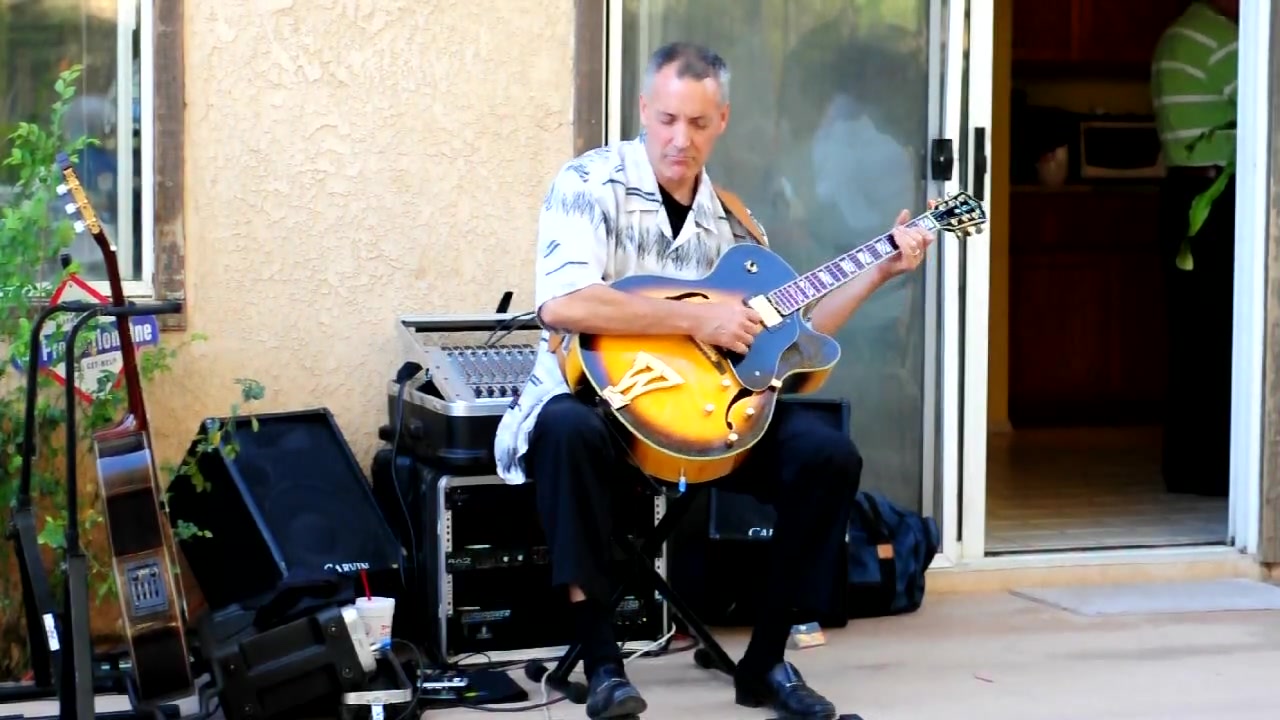}
			\includegraphics[width=0.15\textwidth]{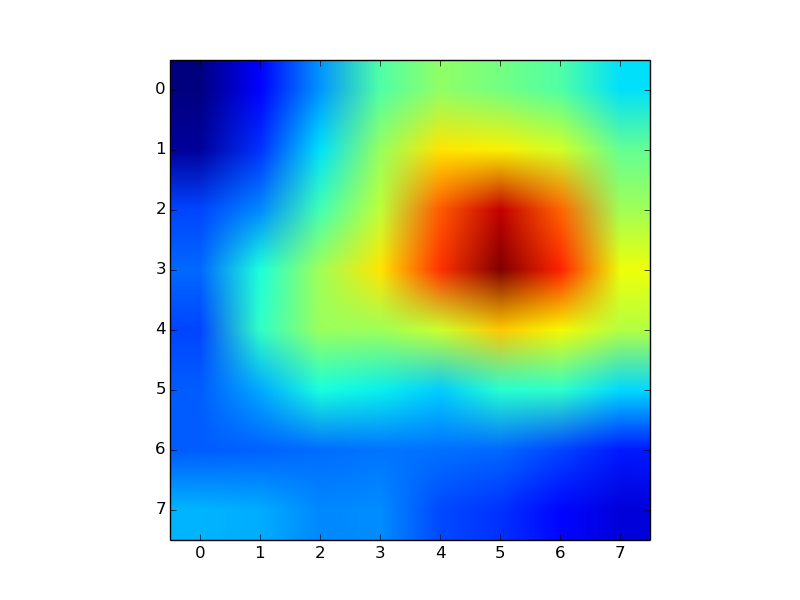}
			\includegraphics[width=0.15\textwidth]{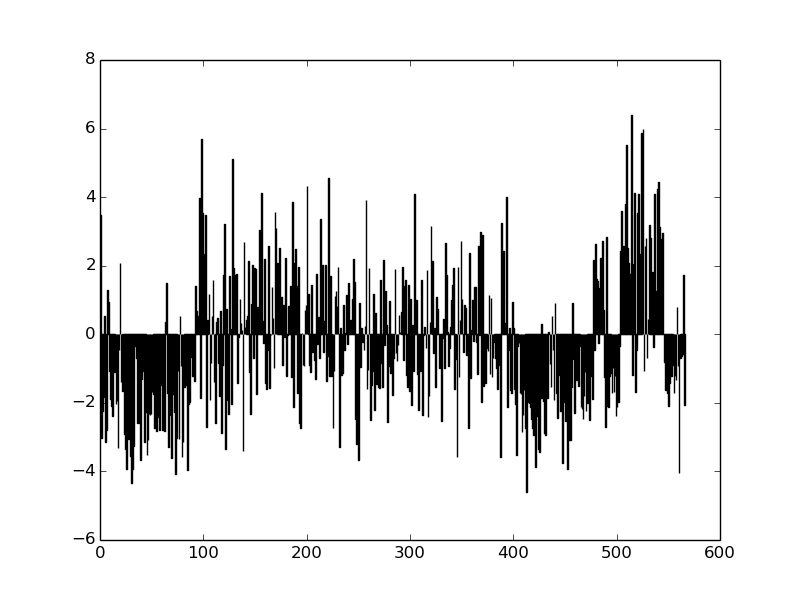}
			\includegraphics[width=0.15\textwidth]{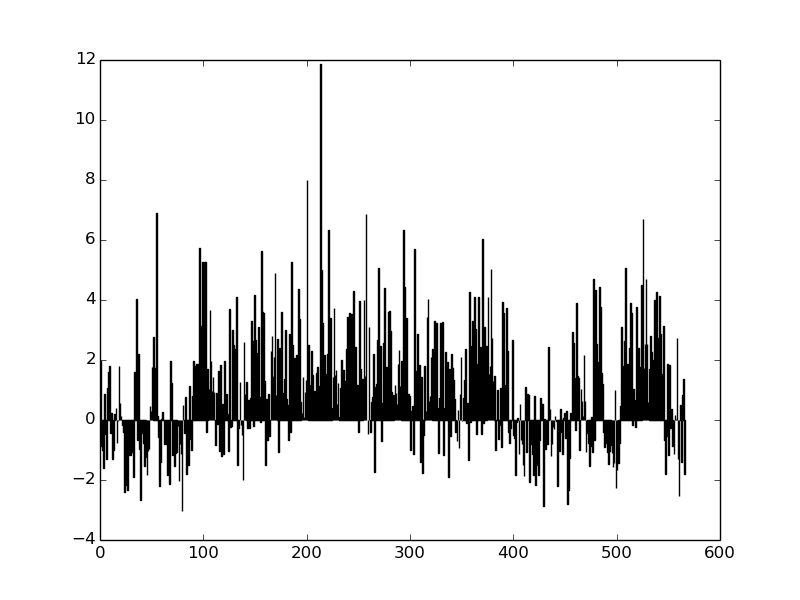}
			\includegraphics[width=0.15\textwidth]{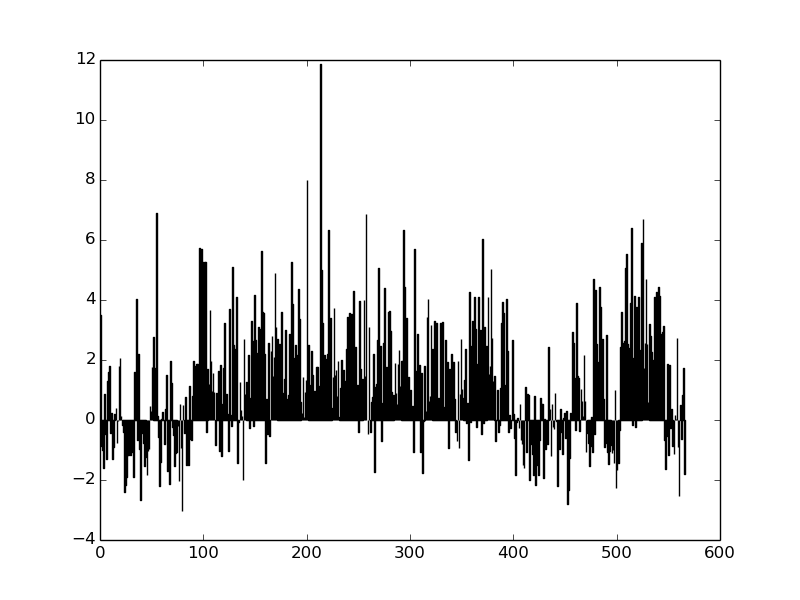}
			\end{tabular} \\			
			\begin{tabular}{cc}
			GT: \textit{A man is playing a guitar.}  &
			MM-VDN: \textit{A man is playing a guitar.}
			\end{tabular}  \\~\\
			
			\begin{tabular}{c}
			\includegraphics[width=0.15\textwidth]{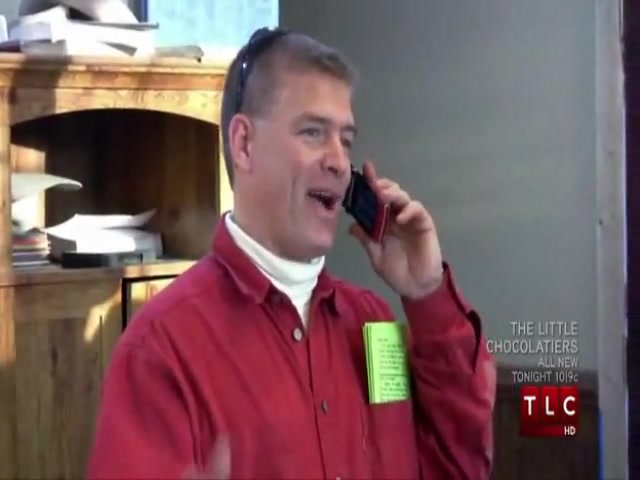}
			\includegraphics[width=0.15\textwidth]{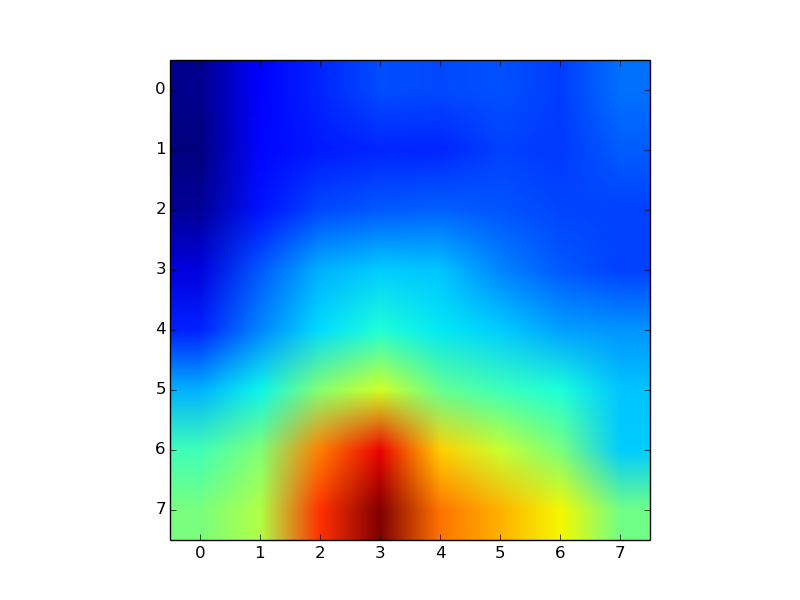}
			\includegraphics[width=0.15\textwidth]{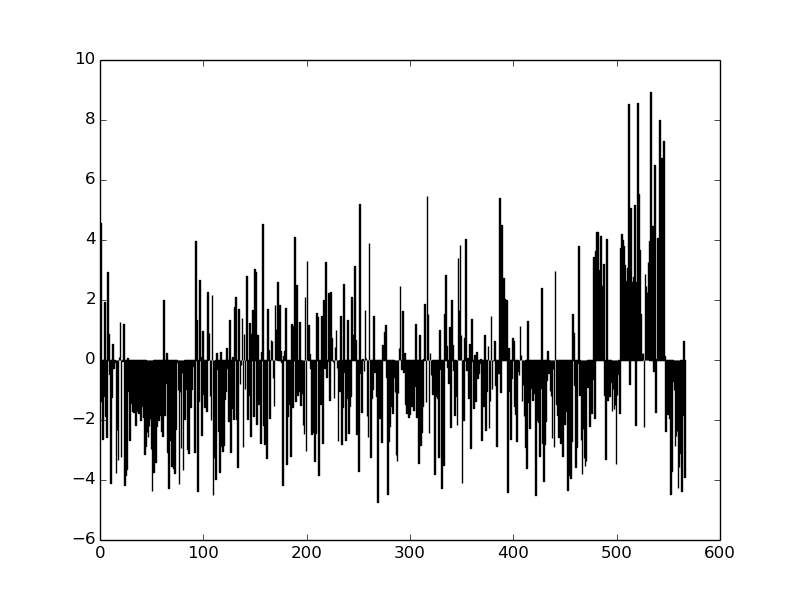}
			\includegraphics[width=0.15\textwidth]{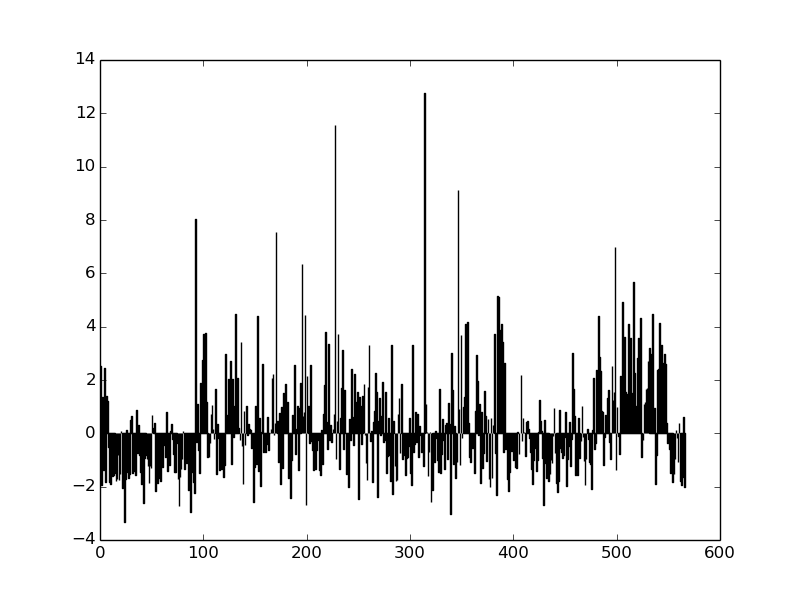}
			\includegraphics[width=0.15\textwidth]{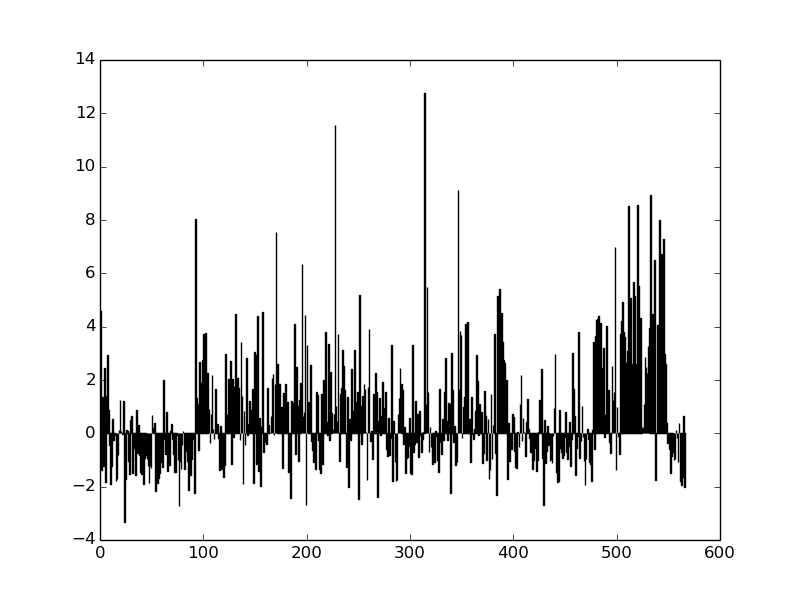}
			\end{tabular} \\			
			\begin{tabular}{cc}
			GT: \textit{A man is talking on a phone}  &
			MM-VDN: \textit{A man is talking on the phone.}
			\end{tabular}  \\~\\
			
			\begin{tabular}{c}
			\includegraphics[width=0.15\textwidth]{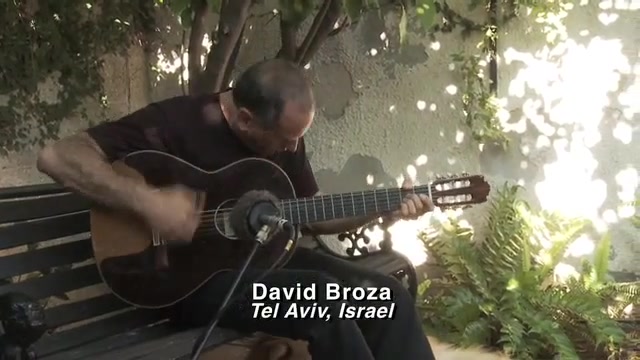}
			\includegraphics[width=0.15\textwidth]{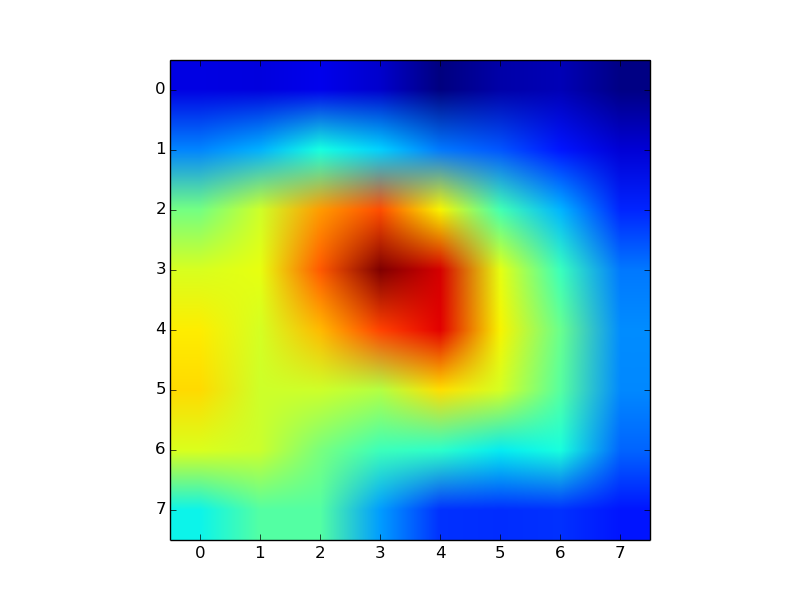}
			\includegraphics[width=0.15\textwidth]{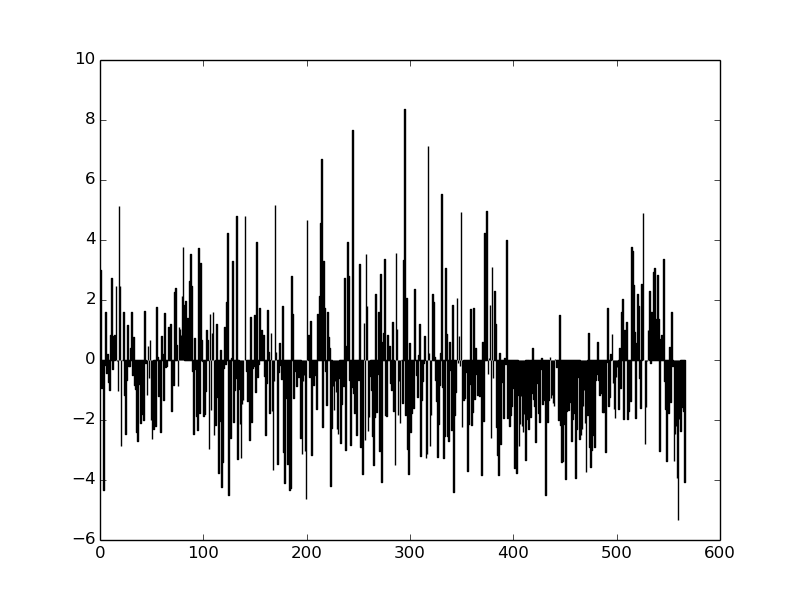}
			\includegraphics[width=0.15\textwidth]{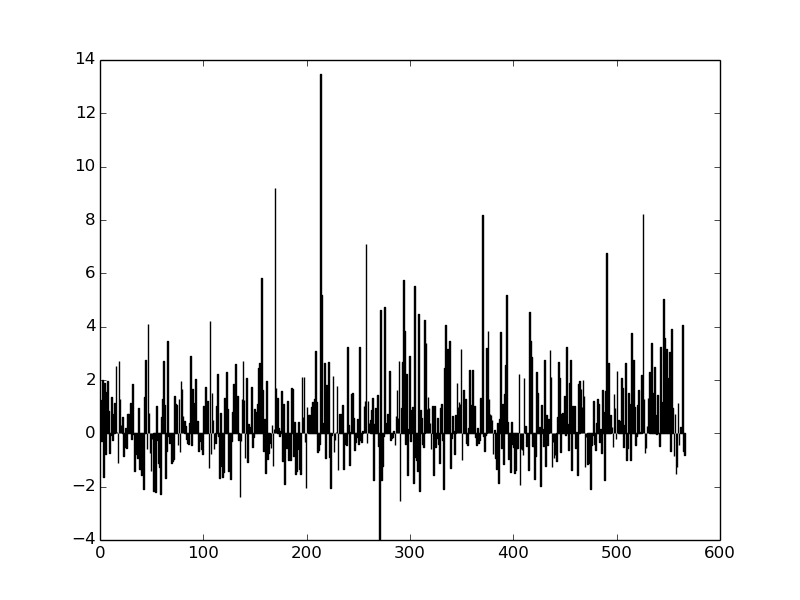}
			\includegraphics[width=0.15\textwidth]{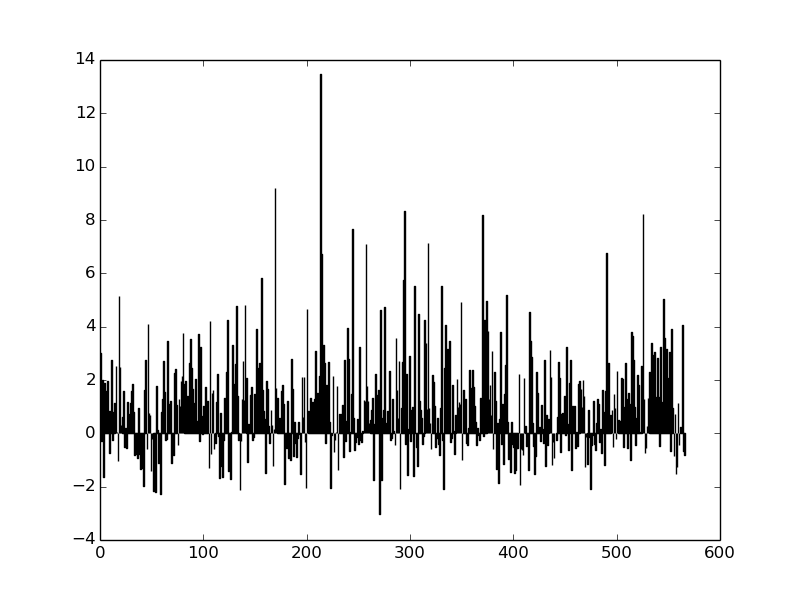}
			\end{tabular} \\			
			\begin{tabular}{cc}
			GT: \textit{A man is playing a guitar.}  &
			MM-VDN: \textit{A man is playing a guitar.}
			\end{tabular}
		\end{tabular}
	\end{center}
\end{table*}

%% file: MM-VDN.bbl
\begin{thebibliography}{10}\itemsep=-1pt

\bibitem{banerjee2005meteor}
S.~Banerjee and A.~Lavie.
\newblock Meteor: An automatic metric for mt evaluation with improved
  correlation with human judgments.
\newblock In {\em Proceedings of the acl workshop on intrinsic and extrinsic
  evaluation measures for machine translation and/or summarization}, pages
  65--72, 2005.

\bibitem{chen2011collecting}
D.~L. Chen and W.~B. Dolan.
\newblock Collecting highly parallel data for paraphrase evaluation.
\newblock In {\em Proceedings of the 49th Annual Meeting of the Association for
  Computational Linguistics: Human Language Technologies-Volume 1}, pages
  190--200. Association for Computational Linguistics, 2011.

\bibitem{elliott2014comparing}
D.~Elliott and F.~Keller.
\newblock Comparing automatic evaluation measures for image description.
\newblock In {\em Proceedings of the 52nd Annual Meeting of the Association for
  Computational Linguistics}, volume~2, pages 452--457, 2014.

\bibitem{fang2014captions}
H.~Fang, S.~Gupta, F.~Iandola, R.~Srivastava, L.~Deng, P.~Doll{\'a}r, J.~Gao,
  X.~He, M.~Mitchell, J.~Platt, et~al.
\newblock From captions to visual concepts and back.
\newblock {\em arXiv preprint arXiv:1411.4952}, 2014.

\bibitem{girshick2014rich}
R.~Girshick, J.~Donahue, T.~Darrell, and J.~Malik.
\newblock Rich feature hierarchies for accurate object detection and semantic
  segmentation.
\newblock In {\em Computer Vision and Pattern Recognition (CVPR), 2014 IEEE
  Conference on}, pages 580--587. IEEE, 2014.

\bibitem{guadarrama:iccv13}
S.~Guadarrama, N.~Krishnamoorthy, G.~Malkarnenkar, S.~Venugopalan, R.~Mooney,
  T.~Darrell, and K.~S. nko.
\newblock Youtube2text: Recognizing and describing arbitrary activities using
  semantic hierarchies and zero-shot recognition.
\newblock In {\em Proceedings of the 14th International Conference on Computer
  Vision (ICCV-2013)}, pages 2712--2719, Sydney, Australia, December 2013.

\bibitem{hochreiter1997long}
S.~Hochreiter and J.~Schmidhuber.
\newblock Long short-term memory.
\newblock {\em Neural computation}, 9(8):1735--1780, 1997.

\bibitem{jia2014caffe}
Y.~Jia, E.~Shelhamer, J.~Donahue, S.~Karayev, J.~Long, R.~Girshick,
  S.~Guadarrama, and T.~Darrell.
\newblock Caffe: Convolutional architecture for fast feature embedding.
\newblock {\em arXiv preprint arXiv:1408.5093}, 2014.

\bibitem{kang2014fully}
K.~Kang and X.~Wang.
\newblock Fully convolutional neural networks for crowd segmentation.
\newblock {\em arXiv preprint arXiv:1411.4464}, 2014.

\bibitem{karpathy14arxiv}
A.~Karpathy and L.~Fei-Fei.
\newblock Deep visual-semantic alignments for generating image descriptions.
\newblock {\em arXiv:1412.2306}, 2014.

\bibitem{krizhevsky2012imagenet}
A.~Krizhevsky, I.~Sutskever, and G.~E. Hinton.
\newblock Imagenet classification with deep convolutional neural networks.
\newblock In {\em Advances in neural information processing systems}, pages
  1097--1105, 2012.

\bibitem{Lazebnik:2006}
S.~Lazebnik, C.~Schmid, and J.~Ponce.
\newblock Beyond bags of features: Spatial pyramid matching for recognizing
  natural scene categories.
\newblock In {\em Proceedings of the 2006 IEEE Computer Society Conference on
  Computer Vision and Pattern Recognition - Volume 2}, CVPR '06, pages
  2169--2178, Washington, DC, USA, 2006. IEEE Computer Society.

\bibitem{li2010object}
L.-J. Li, H.~Su, L.~Fei-Fei, and E.~P. Xing.
\newblock Object bank: A high-level image representation for scene
  classification \& semantic feature sparsification.
\newblock In {\em Advances in neural information processing systems}, pages
  1378--1386, 2010.

\bibitem{long2014fully}
J.~Long, E.~Shelhamer, and T.~Darrell.
\newblock Fully convolutional networks for semantic segmentation.
\newblock {\em arXiv preprint arXiv:1411.4038}, 2014.

\bibitem{cnn_lstm2:archive}
J.~Mao, W.~Xu, Y.~Yang, J.~Wang, and A.~L. Yuille.
\newblock Explain images with multimodal recurrent neural networks.
\newblock {\em CoRR}, abs/1410.1090, 2014.

\bibitem{sivic}
M.~Oquab, L.~Bottou, I.~Laptev, and J.~Sivic.
\newblock Weakly supervised object recognition with convolutional neural
  networks.
\newblock Technical Report HAL-01015140, INRIA, 2014.

\bibitem{papineni2002bleu}
K.~Papineni, S.~Roukos, T.~Ward, and W.-J. Zhu.
\newblock Bleu: a method for automatic evaluation of machine translation.
\newblock In {\em Proceedings of the 40th annual meeting on association for
  computational linguistics}, pages 311--318. Association for Computational
  Linguistics, 2002.

\bibitem{fcn_MIL_segmenation:CoRR}
D.~Pathak, E.~Shelhamer, J.~Long, and T.~Darrell.
\newblock Fully convolutional multi-class multiple instance learning.
\newblock {\em CoRR}, abs/1412.7144, 2014.

\bibitem{rohrbach:iccv13}
M.~Rohrbach, W.~Qiu, I.~Titov, S.~Thater, M.~Pinkal, and B.~Schiele.
\newblock Translating video content to natural language descriptions.
\newblock pages 433--440, 2013.

\bibitem{vgg16arxiv}
K.~Simonyan and A.~Zisserman.
\newblock Very deep convolutional networks for large-scale image recognition.
\newblock {\em CoRR}, abs/1409.1556, 2014.

\bibitem{szegedy2014going}
C.~Szegedy, W.~Liu, Y.~Jia, P.~Sermanet, S.~Reed, D.~Anguelov, D.~Erhan,
  V.~Vanhoucke, and A.~Rabinovich.
\newblock Going deeper with convolutions.
\newblock {\em arXiv preprint arXiv:1409.4842}, 2014.

\bibitem{thomason:coling14}
J.~Thomason, S.~Venugopalan, S.~Guadarrama, K.~Saenko, and R.~Mooney.
\newblock Integrating language and vision to generate natural language
  descriptions of videos in the wild.
\newblock In {\em Proceedings of the 25th International Conference on
  Computational Linguistics (COLING)}, August 2014.

\bibitem{s2s:anon}
S.~Venugopalan, M.~Rohrbach, J.~Donahue, R.~Mooney, T.~Darrell, and K.~Saenko.
\newblock Sequence to sequence \textrm{-} video to text.
\newblock {\em arXiv:1505.00487v2}, 2015.

\bibitem{video_lstm:naccl}
S.~Venugopalan, H.~Xu, J.~Donahue, M.~Rohrbach, R.~J. Mooney, and K.~Saenko.
\newblock Translating videos to natural language using deep recurrent neural
  networks.
\newblock {\em CoRR}, abs/1412.4729, 2014.

\bibitem{cnn_lstm1:archive}
O.~Vinyals, A.~Toshev, S.~Bengio, and D.~Erhan.
\newblock Show and tell: {A} neural image caption generator.
\newblock {\em CoRR}, abs/1411.4555, 2014.

\bibitem{xu2015jointly}
R.~Xu, C.~Xiong, W.~Chen, and J.~J. Corso.
\newblock Jointly modeling deep video and compositional text to bridge vision
  and language in a unified framework.
\newblock In {\em Proceedings of AAAI Conference on Artificial Intelligence},
  2015.

\bibitem{yao15arxiv}
L.~Yao, A.~Torabi, K.~Cho, N.~Ballas, C.~Pal, H.~Larochelle, and A.~Courville.
\newblock Describing videos by exploiting temporal structure.
\newblock {\em arXiv:1502.08029v4}, 2015.

\end{thebibliography}
